%% file: KernelTopicModels.tex
\documentclass[twoside]{article}
\usepackage[accepted]{aistats2012_arxiv}

\usepackage{tikz,amsmath,amssymb,booktabs}
\usepackage{natbib}
\usepackage{pgfplots}

\usepackage{algorithm,algorithmic}

\usetikzlibrary{arrows,shapes,plotmarks,fadings,decorations.pathreplacing}

\tikzset{>=stealth'} 
\tikzstyle{graphnode} = 
   [circle,draw=black,minimum size=22pt,text centered,text
     width=22pt,inner sep=0pt] 
\tikzstyle{var}   =[graphnode,fill=white]
\tikzstyle{obs}   =[graphnode,fill=black,text=white]
\tikzstyle{fac}   =[rectangle,draw=black,fill=black!25,minimum size=5pt]
\tikzstyle{facprior} =[rectangle,draw=black,fill=black,text=white,minimum size=5pt]
\tikzstyle{edge}  =[draw=white,double=black,thick,-]
\tikzstyle{prior} =[rectangle, draw=black, fill=black, minimum size=
5pt, inner sep=0pt]
\tikzstyle{dirprior} = [circle, draw=black, fill=black, minimum
size=5pt, inner sep=0pt]

\newcommand{\g}{\,|\,}
\newcommand{\de}{\partial}
\renewcommand{\d}{\:\mathrm{d}}
\renewcommand{\H}{\mathcal{H}}

\renewcommand{\Re}{\mathbb{R}}
\newcommand{\N}{\mathcal{N}}

\newcommand{\D}{\mathcal{D}}
\newcommand{\Trans}{^{\intercal}}

\newcommand{\diag}{\operatorname{diag}}

\newcommand{\Exp}{\mathsf{E}}
\newcommand{\V}{\mathsf{V}}

\renewcommand{\vec}{\boldsymbol}

\renewcommand{\O}{\mathcal{O}}
\newcommand{\GP}{\mathcal{GP}}

\newcommand{\Lapl}{\mathfrak{L}}

\DeclareSymbolFont{stmry}{U}{stmry}{m}{n}
\DeclareMathSymbol\leftrightarrowtriangle\mathrel{stmry}{"5D}
\DeclareMathSymbol\leftarrowtriangle\mathrel{stmry}{"5E}
\DeclareMathSymbol\rightarrowtriangle\mathrel{stmry}{"5F}
\DeclareMathSymbol\mapstochar\mathrel{stmry}{"5A}

\renewcommand{\to}{\rightarrowtriangle}

\renewcommand{\leftrightarrow}{\leftrightarrowtriangle}

\newif\iffinal 

\pgfrealjobname{KernelTopicModels}  
\iffinal
 
\else
 
\fi

\begin{document}

\runningauthor{P. Hennig, D. Stern, R. Herbrich, T. Graepel}

\twocolumn[
\aistatstitle{Kernel Topic Models}


\aistatsauthor{ Philipp Hennig$^{1,2,3}$  \And David Stern$^3$ \And Ralf Herbrich$^3$ \And Thore Graepel$^3$ }
\aistatsaddress{ $^1$Cavendish Laboratory\\ Cambridge, UK \And $^2$Max Planck Institute for Intelligent Systems\\
T\"ubingen, Germany \And $^3$Microsoft Research\\ Cambridge, UK} 
]

\begin{abstract}
  Latent Dirichlet Allocation models discrete data as a mixture of
  discrete distributions, using Dirichlet beliefs over the mixture
  weights.  We study a variation of this concept, in which the
  documents' mixture weight beliefs are replaced with squashed
  Gaussian distributions. This allows documents to be associated with
  elements of a Hilbert space, admitting \emph{kernel topic models
    (KTM)}, modelling temporal, spatial, hierarchical, social and
  other structure between documents. The main challenge is efficient
  approximate inference on the latent Gaussian. We present an
  approximate algorithm cast around a Laplace approximation in a
  transformed basis. The KTM can also be interpreted as a type of
  Gaussian process latent variable model, or as a topic model
  conditional on document features, uncovering links between earlier
  work in these areas.
\end{abstract}

\section{Introduction}

Latent Dirichlet Allocation (LDA) \citep{blei2003latent} is a
generative model for datasets comprising collections of discrete
samples. Each collection is assumed to be generated from a mixture of
discrete distributions, such that both the belief over the discrete
distributions and over the mixture weights are Dirichlet. Text
documents constitute the most popular domain with this anatomy: Each
document in a corpus, treated as a ``bag of words'' (i.e.\ ignoring
word order), is one collection of (discrete) words, and the mixture
components are interpreted as \emph{topics}. Thus, each document
exhibits several topics to varying degree, with each word in the
document sampled from one specific topic. 

Real documents do not exist void of context. They are products of
their authors, time, and place. Electronic communication has
intensified this truism, and online corpora are now invariably
accompanied by copious amounts of meta-data. The identity of the
author may be augmented by additional knowledge about their location
in a social graph, autobiographic information, and many more. Such
features convey semantic information: Topic popularity varies between
West and East, conservatives and progressives, rich and poor,
scientists and celebrities, young and old, contemporaries and
forebears.

In its standard form, LDA can not take advantage of such metadata; but
extensions proposed by several authors have addressed certain
types of meta-structure. Dynamic development of topics over sequential
sets of documents was considered by
\citet{Blei:2006:DTM:1143844.1143859},
\citet{Wang:2006:TOT:1150402.1150450} and
\citet{wang2009continuous}. Both \citet{mimno-topic} and
\citet{zhu2010conditional} considered a more general description of 
topics in terms of a linear function in a latent real vector space, linked to
the topic dimension through the softmax function. These works differ
in their details (some assume the topics stay constant over time while
their distribution changes, others that the topics themselves
change. Words may be assumed to generate features, or the other way
round), but are linked by their common use of \emph{Gaussian} random
variables to describe dynamics or regress on document features. They
also all use maximum likelihood, or maximum a-posteriori inference
to fit regression weights where they exist.

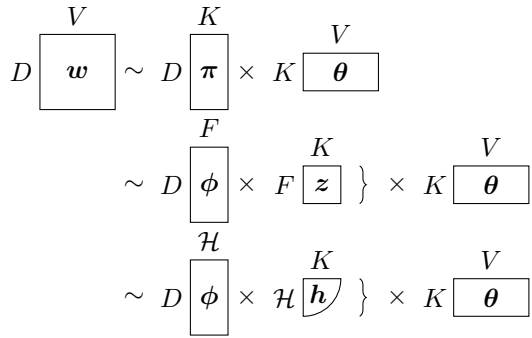
\begin{figure}%
\begin{center}
\begin{tikzpicture}

\node at (-0.25,0.5) {$D$};
\draw(0,0) rectangle (1,1);
\node at (0.5,0.5) {$\vec{w}$};
\node at (0.5,1.25) {$V$};

\node at (1.25,0.5) {$\sim$};

\node at (1.75,0.5) {$D$};
\draw (2.0,0) rectangle (2.5,1);
\node at (2.25,0.5) {$\vec{\pi}$};
\node at (2.25,1.25) {$K$};

\node at (2.75,0.5) {$\times$};

\node at (3.25,0.5) {$K$};
\draw (3.5,0.25) rectangle (4.5,0.75);
\node at (4.0,0.5) {$\vec{\theta}$};
\node at (4.0,1.0) {$V$};

\begin{scope}[yshift=-1.5cm]
\node at (1.25,0.5) {$\sim$};

\node at (1.75,0.5) {$D$};
\draw (2.0,0) rectangle (2.5,1);
\node at (2.25,0.5) {$\vec{\phi}$};
\node at (2.25,1.25) {$F$};

\node at (2.75,0.5) {$\times$};

\node at (3.25,0.5) {$F$};
\draw(3.5,0.25) rectangle (4.0,0.75);
\node at (3.75,0.5) {$\vec{z}$};
\node at (3.75,1.0) {$K$};

\draw[decoration=brace,decorate](4.25,0.75) -- (4.25,0.25);
\begin{scope}[xshift=0.5cm]
\node at (4.25,0.5) {$\times$};

\node at (4.75,0.5) {$K$};
\draw (5.0,0.25) rectangle (6.0,0.75);
\node at (5.5,0.5) {$\vec{\theta}$};
\node at (5.5,1.0) {$V$};
\end{scope}
\end{scope}

\begin{scope}[yshift=-3cm]

\node at (1.25,0.5) {$\sim$};

\node at (1.75,0.5) {$D$};
\draw (2.0,0) rectangle (2.5,1);
\node at (2.25,0.5) {$\vec{\phi}$};
\node at (2.25,1.25) {$\H$};

\node at (2.75,0.5) {$\times$};

\node at (3.25,0.5) {$\H$};
\draw (3.5,0.75) -- (3.5,0.25) arc (-90:0:0.5cm) -- (3.5,0.75);
\node at (3.7,0.55) {$\vec{h}$};
\node at (3.75,1.0) {$K$};

\draw[decoration=brace,decorate](4.25,0.75) -- (4.25,0.25);
\begin{scope}[xshift=0.5cm]
\node at (4.25,0.5) {$\times$};

\node at (4.75,0.5) {$K$};
\draw (5.0,0.25) rectangle (6.0,0.75);
\node at (5.5,0.5) {$\vec{\theta}$};
\node at (5.5,1.0) {$V$};
\end{scope}
\end{scope}

\end{tikzpicture}
\end{center}
\caption{Dimensionality-reduction view of topic models. {\bfseries
    Top:} LDA describes $D$ documents containing words from a
  vocabulary of size $V$ in terms of $K$ topics. {\bfseries Middle:}
  Dirichlet multinomial regression portrays the documents in terms of
  $F$ features, which generate the topics through a linear
  map. {\bfseries Bottom:} The kernel topic model replaces the
  features with co-ordinates of a Hilbert space $\H$, and the linear
  map with a nonlinear one. The curly brace denotes a
  softmax-projection from $\Re^K$ to the $[0,1]^K$ simplex.}%
\label{fig:dim-red}%
\end{figure}

This work generalizes these approaches by replacing real-valued
features with elements of a Hilbert space, and point estimates with
Gaussian process measures (Figure \ref{fig:dim-red}). The resulting
\emph{kernel topic model} provides an expressive framework for the
inclusion of virtually all types of metadata in the semantic
description of topical data, and allows a rich description of
nonlinear topic dynamics. The main mathematical challenge is that
inference on the latent Gaussian belief is not analytically
tractable. We address this through a numerically lightweight Laplace
approximation for Dirichlet distributions in the softmax basis,
extending on a note by \citet{mackay1998choice}. As a side effect,
this approximation also admits a particularly efficient implementation
of Bayesian inference on linear latent models, such as the one
introduced by \citet{mimno-topic}. The kernel topic model links topic
modelling and Gaussian process latent variable models, effectively
casting LDA as a likelihood for generalised Gaussian process
models. The price of the increased modelling flexibility is a
comparably high computational cost -- cubic in the number of
documents. 

\section{Methods}
\label{sec:methods}
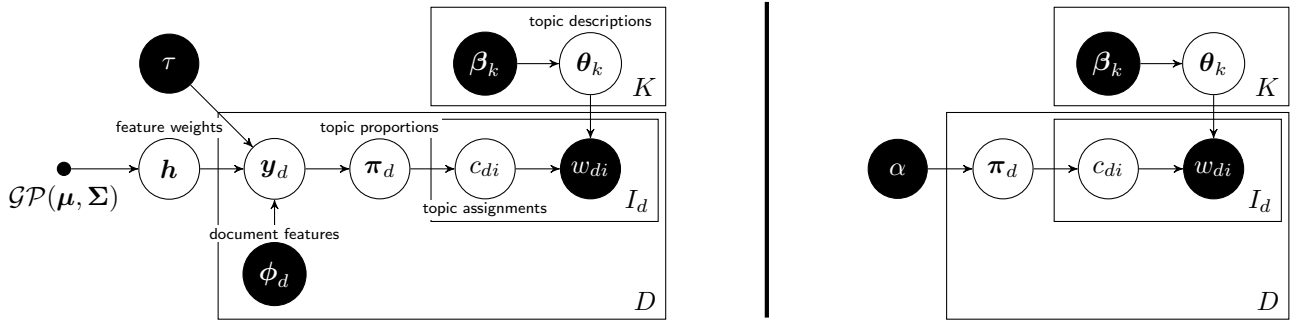
\begin{figure*}[t]%
\centering
\begin{tikzpicture}[node distance = 1.4cm]
	\node[dirprior] at (0,0) (pz) {};
        \node[anchor=north] at (pz.south) {$\GP(\vec{\mu},\vec{\Sigma})$};
	\node[var, right of=pz] (z)  {$\vec{h}$} edge [<-] (pz);
	\node[obs, above of=z] (t) {$\tau$};
	\node[var, right of=z] (y) {$\vec{y}_d$} edge [<-] (z);
	\node[obs, below of=y] (phi) {$\vec{\phi}_d$} edge [->] (y);
        \node[var, right of=y] (pi) {$\vec{\pi}_d$} edge [<-] (y);
	\node[var, right of=pi] (c)   {$c_{di}$} edge [<-] (pi);
	\node[obs, right of=c] (w)   {$w_{di}$} edge [<-] (c);
	\node[obs, above of=c] (b)  {$\vec{\beta}_k$};
        \node[var, above of=w] (theta) {$\vec{\theta}_k$} edge [<-] (b)
        edge [->] (w);

	\draw (2.05,-2) rectangle (8,0.75);
	\node[anchor=south east] at (8,-2) {$D$};

	\draw (4.88,-0.7) rectangle (7.9,0.66);
	\node[anchor=south east] at (7.9,-0.7) {$I_d$};

        \draw (4.88,0.84) rectangle (8,2.15);
        \node[anchor=south east] at (8,0.84) {$K$};

        \node[anchor=south,inner sep=1pt,fill=white] 
        at (pi.north) {\tiny \sf topic proportions};
        \node[anchor=north,inner sep=1pt,fill=white] 
        at (c.south) {\tiny \sf topic assignments};
        \node[anchor=south,inner sep=1pt,fill=white] 
        at (phi.north) {\tiny \sf document features};
        \node[anchor=south,inner sep=1pt,fill=white] 
        at (theta.north) {\tiny \sf topic descriptions};
        \node[anchor=south,inner sep=1pt,fill=white] 
        at (z.north) {\tiny \sf feature weights};
	
        \draw (t) edge[->] (y);
\end{tikzpicture}\hfill
\begin{tikzpicture}
    \draw[ultra thick] (2.0,-2) -- (2.0,2.2);
\end{tikzpicture}\hfill
\begin{tikzpicture}[node distance = 1.4cm] 
  \node[obs] at (2.8,0) (y) {$\alpha$};
  \node[var, right of=y] (pi) {$\vec{\pi}_d$} edge [<-] (y);
  \node[var, right of=pi] (c)   {$c_{di}$} edge [<-] (pi);
  \node[obs, right of=c] (w)   {$w_{di}$} edge [<-] (c);
  \node[obs, above of=c] (b)  {$\vec{\beta}_k$};
  \node[var, above of=w] (theta) {$\vec{\theta}_k$} edge [<-] (b)
  edge [->] (w);
  
  \draw (3.45,-2) rectangle (8,0.75);
  \node[anchor=south east] at (8,-2) {$D$};
  
  \draw (4.88,-0.7) rectangle (7.9,0.66);
  \node[anchor=south east] at (7.9,-0.7) {$I_d$};
  
  \draw (4.88,0.84) rectangle (8,2.15);
  \node[anchor=south east] at (8,0.84) {$K$};
\end{tikzpicture}
\caption{{\bf Left:} Directed graphical model of the kernel topic
  model.  Some variables labeled for clarity. {\bf Right:} latent
  Dirichlet allocation. The models are identical to the right of and
  including $\vec{\pi}$.}%
\label{fig:graph}%
\end{figure*}

\subsection{Model}
\label{sec:model}

We consider a corpus of $D$ documents. Document $d$ contains $I_d$
words $w_{di}\in \{1,\dots,V\}, d\in \{1,\dots,D\}, i\in
\{1,\dots,I_d\}$ from a vocabulary of size $V$. Additional aspects of
$d$ are described by features $\vec{\phi}_d\in \H$ in a Hilbert space
$\H$. In other words, the dataset consists of pairs
$(\vec{w}_d,\vec{\phi}_d) \in \{1,\dots,V\}^{I_d}\times\H$.

We construct a topic model conditional on the observable features of
the documents, using the following generative process for the vector
$\vec{w}_d$ from $K$ topics:
\begin{itemize}
\item For each topic $k\in \{1,\dots,K\}$, generate a discrete
  probability distribution with parameters $\vec{\theta}_k\in
  [0,1]^V$ over the vocabulary of size $V$ by sampling from a
  Dirichlet distribution with parameter vector $\vec{\beta}_k$ 
  ($\Gamma$ denotes the Gamma function):
  \begin{equation}
    \label{eq:t_1}
    p(\vec{\theta}_k \,|\, \vec{\beta}_k) = \D(\vec{\theta}_k ;
    \vec{\beta}_k) = \frac{\Gamma\left(\sum_v ^V \beta_{kv}
      \right)}{\prod_v ^V \Gamma(\beta_{kv})} \prod_k \theta_{kv}
    ^{{\beta_{kv}} - 1}.
  \end{equation}
\item Independently sample $K$ functions $h_k(\vec{\phi}):\H\to\Re$
  from the Hilbert space of real-valued functions over $\H$, by
  sampling from Gaussian process priors with mean functions
  $\mu_{k}(\vec{\phi}_d)$ and covariance functions
  $\Sigma_k(\vec{\phi}_d,\vec{\phi}_{d'})$, induced by (potentially
  topic-specific) kernels $\eta_k$:
  \begin{equation}
    \label{eq:t_2}
    p(h_k \g \mu_k,\Sigma_k) =
    \GP(h_{k};\mu_{k},\Sigma_{k} ^2)
  \end{equation}
\item For each document $d$ with features $\vec\phi_d
  \in \Re^{F}$,
  \begin{itemize}
  \item Draw a latent variable $\vec{y}_d$ by evaluating
    $\vec{h}(\vec{\phi}_d)$ and adding Gaussian noise of standard
    deviation $\tau$:
  \begin{equation}
    \label{eq:t_3}
    p(\vec{y}_d\g \vec{h},\tau,\vec{\phi}_d) =
    \prod_k \N(\vec{y}_{dk};h_k(\vec{\phi}_d),\tau^2)
  \end{equation}
\item Define the topic proportions $\vec{\pi}_d=\sigma(\vec{y}) \in [0,1]^K$ where
  $\sigma$ is the softmax function
  \begin{equation}
    \label{eq:t_4}
    \sigma_k(\vec{y}) = \frac{\exp(y_k)}{\sum_\ell ^K \exp(y_\ell)}
  \end{equation}
\item For each of $I_d$ words
  \begin{itemize}
  \item draw a topic $c_{di}$ from the discrete distribution defined
    by $\vec{\pi}_d$:
    \begin{equation}
      \label{eq:t_5}
      p(c_{di}=k\g \vec{\pi}_d) = \pi_{dk}
    \end{equation}
  \item draw word $w_{di}$ from the discrete distribution of topic
    $c_{di}$:
    \begin{equation}
      \label{eq:t_6}
      p(w_{di} = v \g c_{di},\vec{\Theta}) = \theta_{c_{di}v}
    \end{equation}
  \end{itemize}
  \end{itemize}
\end{itemize}
The directed graphical model in Figure \ref{fig:graph}, left, sheds
light on the dependency structure of this generative model. If we
replace everything to the left of $\vec{\pi}_d$ in that figure by a
single Dirichlet parameter vector $\vec{\alpha}$ (identical for all
$d$), then the parts shown to the right of and including the node
$\vec{\pi}$ correspond to the traditional LDA model
\citep{blei2003latent} (Figure \ref{fig:graph}, right). On the other
hand, we can identify the parts to the left of (and excluding)
$\vec{\pi}$ as a case of Gaussian process regression. It is the
connection between these two parts that makes the model challenging,
and approximate inference will in fact separate in this way.

In passing, we note a connection to the \emph{correlated topic model}
\citep{blei2007correlated}, which shares everything to the right of
and including $\vec{y}$ in Figure \ref{fig:graph}, but not the
regression element to its left. Instead, that model focusses on
estimating the correlation between topics, which here is replaced by a
simpler, diagonal covariance. Introducing correlations between topics
is possible in our model using an approach analogous to the cited work
(maximum likelihood estimation on the covariance structure), but left
out here for clarity.

\section{Inference}
\label{sec:inference}

Ample expertise has accumulated in the literature, on inference for
both LDA, and Gaussian processes given (approximately) Gaussian
likelihoods. What is missing is a connection between the two
paradigms. This link is the main contribution of this paper. To
clarify the setting, however, we give a very brief introduction to the
two sub-systems in this section, then derive the link -- the
\emph{Laplace bridge} -- in Section \ref{sec:lapl-bridge-link}.

\subsection{Semi-Collapsed Variational Inference}
\label{sec:semi-coll-vari}

Broadly speaking, there are two popular methods for inference in LDA:
variational inference \citep{blei2003latent} and collapsed Gibbs
sampling \citep{griffiths2004finding}. Gibbs samples come from the
exact posterior, but provide no analytic form for the beliefs. Since
our extension benefits from such forms, we
opt for a variational approximation. Standard inference in LDA
\citep{blei2003latent,BleiReview,NIPS2010_1291} uses a fully
factorized approximate distribution, but \citet{teh2007collapsed}
showed that this Ansatz entails an unnecessarily loose bounds and slow
convergence. To mitigate this problem, latent variables should be
integrated out wherever possible. Since we require explicit forms for
the per-document topic distributions $\vec{\pi}_d$, we can not
integrate out this variable, but we \emph{can} collapse the bound on
the per-topic distributions $\vec{\theta}$. This amounts to an
adaptation to Teh et al.'s work, which we do not dwell on here for
brevity. The bottom line is that it is possible to construct a
variational bound that, given a Dirichlet prior 
\begin{equation}
  \label{eq:1}
  p(\vec{\pi}_d\g \vec{\alpha}_d) = \D(\vec{\pi}_d ;\vec{\alpha}_d)
\end{equation}
on $\pi_d$, assigns approximate Dirichlet ``posterior'' beliefs
\begin{equation}
  \label{eq:2}
  p(\vec{\pi}_d\g \vec{\alpha_d},\vec{w}_d) =
  \D(\vec{\pi}_d;\vec{\alpha}_d + \vec{\nu}_d)
\end{equation}
with a vector $\vec{\nu}_d\in\Re^K$ of pseudo-counts. At the LDA end
of the divide between Gaussian regression and LDA, we thus require a
Dirichlet belief.

\subsection{Gaussian Process Regression}
\label{sec:appr-gauss-proc}

For the moment, assume there be some isomorphism
$\Lapl$ between $K$-dimensional Dirichlet distributions and $K$
approximately independent Gaussian ones (to be developed in Section
\ref{sec:lapl-bridge-link}).
\begin{equation}
  \label{eq:3}
  \Lapl: \quad \D(\vec{\pi}_d;\vec{\alpha}_d) \leftrightarrow \prod_{k=1} ^K
  \N(\vec{y}_d;\mu_{dk},\sigma^2 _{dk})
\end{equation}
This transform provides approximate Gaussian \emph{messages} from
$\vec{\pi}_d$ to $\vec{y}_d$ in the graph of Figure
\ref{fig:graph}. With these messages, Gaussian process inference over
the Hilbert space $\H$ becomes a known problem, and we can implement
an approximate Gaussian process latent inference algorithm: For every
topic $k$, the posterior belief over the function $h_k(\vec{\phi}_*)$
at the Hilbert location $\vec{\phi}_*$ is the product of the Gaussian
process prior and the $D$ approximately independent Gaussian messages
$p(\vec{y}_d \g h(\phi_d),\vec{W},\vec{\Theta}) =
\N(\mu_{kd};h_{k}(\phi_d),\tau^2 + \sigma_{kd} ^2)$. We subsume the
means of these messages into a vector $\vec{\mu}_k$ and their
variances into a diagonal matrix $\Sigma_k = \diag(\tau^2 + \sigma^2
_{dk})$, which allows us to write the mean and marginal variance
functions of the posterior Gaussian process as
\begin{equation}
  \label{eq:1}
  \begin{aligned}[t]
    \Exp[h_*] &=
    \eta_{k}(\vec{\phi}_*,\vec{\Phi}) \Trans H^{-1} (H^{-1} +
    \tilde\Sigma_k ^{-1})^{-1} \tilde\Sigma_k ^{-1}
    \tilde{\vec{\mu}}_k\\
    \V[h_*] &=
    \eta_{k}(\vec{\phi}_*,\vec{\phi_*}) -
    \eta_{k}(\vec{\phi}_*,\vec{\Phi}) (H + \tilde\Sigma)^{-1}
    \eta_{k}(\vec{\Phi},\vec{\phi}_*)
  \end{aligned}
\end{equation}
writing the message \emph{precisions} (inverse variances) as
$\zeta_d=(\sigma_d ^{2} + \tau^2)^{-1}$, we construct a matrix
$\tilde{S} = \diag(\vec{\zeta})$ and message precision adjusted means
$\tilde{\vec{\nu}} = \tilde{S}\tilde{\vec{\mu}}$. Using this notation,
the implementation of iterative Gaussian process inference from
approximate Gaussian messages contained in Section 3.6.3, in
particular Algorithms 3.5 and 3.6 in \citet{RasmussenWilliams} can be
used almost without changes. 

In Gaussian process generalised regression, the hyperparameters
(kernel parameters and observation noise) are usually estimated by
evidence maximisation (``type-II maximum likelihood''). In our case,
the unknown function is $h$, the data is $\vec{w}$ and let the
hyperparameters be $\xi$. Evidence maximisation would amount to
optimising $p(\vec{w}\g \xi) = \int p(\vec{w}\g f,\xi) p(f\g \xi) \d
f$. However, in our case, there is an approximate inference algorithm
separating the Gaussian process regression from the observed data, so
this kind of optimisation has exceedingly high computational cost
(each evaluation of $p(\vec{w}\g \xi)$ involves running the LDA
part of Section \ref{sec:semi-coll-vari} to convergence). Instead, a
much cheaper, if less effective, method is to maximise
$p(\vec{y}\g\xi)$, where $\vec{y}$ are the estimated per-document
topic distributions. Defining the matrix $B = \vec{I} +
\tilde{S}^{1/2}K\tilde{S}^{1/2}$, a simple algebraic argument similar
to the one in \citet{RasmussenWilliams}, Section 3.6.3, gives the log
evidence
\begin{equation}
  \label{eq:5}
  \log Z = \frac{1}{2}\left[ \log |\tilde{S}| - \log|B| - \tilde{\mu}\Trans \tilde{S}^{1/2}
    B^{-1}\tilde{S}^{1/2} \tilde{\mu} \right]
\end{equation}
which is numerically stable (because all eigenvalues of $B$ are larger
than $1$), and can be implemented efficiently. 
Derivatives of $\log Z$ with respect to the kernel parameters,
required for efficient optimisation, are straightforward to calculate
using linear algebra identities.

\subsection{The Laplace Bridge}
\label{sec:lapl-bridge-link}

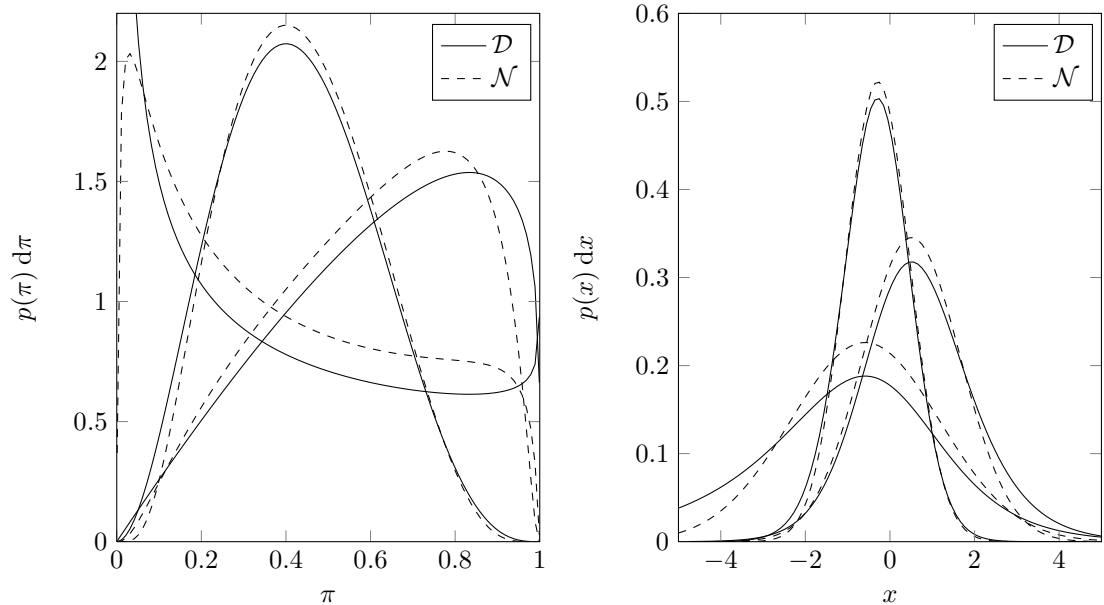
\begin{figure*}[ht]
  \centering
  \input{simplex_plots.tikz}%
  \input{real_plots.tikz}
  \caption{Laplace approximations between Gaussians and
    Dirichlets. {\bf Left:} Simplex basis. {\bf Right:} Softmax
    basis. The parameter choices for the Beta distributions (special
    1D case of the Dirichlet) are
    $(a,b)=(2,1.2);(0.5,0.9);(3,4)$. Under the Laplace approximation,
    these correspond to one-dimensional Gaussian parameters
    $(\mu,\sigma^2)=(0.5,1.3);(-0.6,3.1);(-0.3,0.6)$.  Note that the
    Laplace approximation matches modes (and means) in the softmax
    basis (right), not the simplex basis (left).}
  \label{fig:Laplace_examples}
\end{figure*}

To link these two parts of the inference, we must connect the
Dirichlet belief on $\vec{\pi}_d$ and the Gaussian domain required for
$\vec{y}_d$. Since $\sigma(\vec{y}_d) = \vec{\pi}_d$, this task
amounts to an uncertain form of logistic regression, in the sense that
discrete samples $c_{dn}$ from the distribution defined by
$\vec{\pi}_d$ are replaced by probabilistic beliefs over $c_{dn}$. Our
solution to this problem is to construct a Laplace approximation to
Dirichlet distributions \emph{in the softmax basis}, in which these
distributions can be approximated by Gaussians much better than in the
popular simplex basis.

\citet{mackay1998choice} showed that, because the softmax function has
a Jacobian proportional to $\prod_k \pi_k$, a basis change from
probabilities $\vec{\pi}$ to real numbers
$\vec{y}=\sigma^{-1}(\vec{\pi})$ gives the Dirichlet a new parametric
form
\begin{equation}
  \label{eq:lap_8_body}
  \D_y(\vec{\pi}(\vec{y});\vec{\alpha}) = \frac{\Gamma \left(\sum_k ^K
      \alpha_k \right) } {\prod_k ^K \Gamma(\alpha_k)} \prod_k ^K \pi_k
  ^{\alpha_k} g(\vec{1}\Trans\vec{y})
\end{equation}
$g(\vec{1}\Trans\vec{y})$ is an arbitrary normalisable measure,
required to ensure integrability by restricting the sum of the
elements of $\vec{y}$ ($\vec{1}$ is the vector $[1,1,1,\dots]$). In
this basis, the Dirichlet lacks the $-1$ terms in the exponents
present in the standard representation, and thus does not diverge for
$|x|\to\infty$ and $\alpha_i<1$. It is also a unimodal distribution
whose mode at $\vec{\pi}(\vec{y}) = \vec{\alpha} / \| \vec{\alpha} \|$
now falls together with its mean. These aspects allow a good quality
Laplace approximation.

For numerical convenience, we choose (like MacKay)
\begin{equation}
  \label{eq:lap_2}
  g = \exp\left(-\frac{\epsilon}{2} (\vec{1}\Trans\vec{y})^2 \right).
\end{equation}
MacKay shows the Hessian of the logarithm of this distribution has
elements
\begin{equation}
  \label{eq:lap_3}
  \begin{aligned}
    L_{k\ell} (\vec{y})&=  \frac{\de^2 \D(\vec{y})}{\de y_k \de y_l}
    = \hat{\alpha} \left( \delta_{k\ell} \pi_k - \pi_k
      \pi_\ell \right) + \epsilon (\vec{1}\vec{1}\Trans)_{k\ell}
  \end{aligned}
\end{equation}
(using Kronecker's $\delta$, and $\hat\alpha=\sum_k \alpha_k$. The
$\epsilon$ stems from Eq.\ \eqref{eq:lap_2}). To construct a Laplace
approximation of the Dirichlet in the form of a multivariate Gaussian
$\N(\vec{y};\vec{\mu},\vec{\Sigma})$ (deviating from MacKay's
derivations from here on), we identify the mean $\vec{\mu}$ with the
mode of the distribution,
\begin{equation}
  \label{eq:4}
  \mu_k = \log \alpha_k - \frac{1}{K} \sum_{\ell = 1} ^K \log
  \alpha_\ell,
\end{equation}
and the negative logarithm of its Hessian with $\vec{\Sigma}$. To gain
a sparse approximation, we analytically invert the Hessian. To do so,
we introduce the rectangular matrix $\vec{X} \in \Re^{K\times 2}$ with
elements $X_{ku} = \hat\pi_k \delta_{1u} + \vec{1}_k \delta_{2u}$ and
the square matrices $\vec{A}\in\Re^{K\times K}$ and
$\vec{B}\in\Re^{2\times 2}$
\begin{equation}
  \label{eq:lap11}
  \vec{A} = \diag (\vec{\alpha}) \qquad \text{and} \qquad \vec{B} =
  \begin{pmatrix}
    -\hat \alpha & 0 \\ 0 & \epsilon
  \end{pmatrix}
\end{equation}
which allows us to write $\vec{L} = \vec{A} + \vec{XBX}\Trans$.  Both
$\vec{A}$ and $\vec{B}$ are diagonal with strictly positive diagonal
elements, and thus invertible. Hence we can use the matrix inversion
lemma, which exposes an analytically invertible $2\times 2$ Schur
complement and thus easily yields the inverse of the Hessian
\begin{equation}
  \label{eq:lap_17}
  \begin{aligned}
L^{-1} _{k\ell} &= \delta_{k\ell} \frac{1}{\alpha_k} - \frac{1}{K}
\left[ \frac{1}{\alpha_k} + \frac{1}{\alpha_\ell} - \frac{1}{K}
  \left(\frac{1}{\epsilon} + \sum_u ^K \frac{1}{\alpha_u}\right) \right]
  \end{aligned}
\end{equation}
because this inverse is defined for all positive values of $\epsilon$,
we can safely take the limit of $\epsilon\to\infty$, i.e.\
$g(x)\to\delta(x)$, to the Dirac point distribution. Note that the
off-diagonal elements of this matrix are suppressed with $\O(1/K)$, so
for large $K$, the belief is approximately independent, with
element-wise variances
\begin{equation}
  \label{eq:lap18}
  \Sigma_{kk} = \frac{1}{\alpha_k} \left( 1 - \frac{2}{K}\right) +
  \frac{1}{K^2} \sum_\ell ^K \frac{1}{\alpha_\ell}.
\end{equation}
(This map is only valid for $K>2$. In the 2D-case, a special, much
simpler solution can be derived by mapping directly to the real
line. See also Figure \ref{fig:Laplace_examples}). It is not hard to
invert this $\vec{\alpha}\to(\vec{\mu},\vec{\Sigma})$ map from
Dirichlet to Gaussian parameters, giving
\begin{equation}
  \label{eq:lap21}
  \alpha_k = \frac{1}{\Sigma_{kk}} \left(1 - \frac{2}{K} +
    \frac{e^{-\mu_k}}{K^2}\sum_\ell ^K e^{-\mu_\ell} \right)\;\forall\;k = 1,\dots,K
\end{equation}
Figure \ref{fig:Laplace_examples} gives an intuition for the quality
and defects of this approximation in the 2D case. The approximation is
very good for large entries in $\vec{\alpha}$, but retains good
quality even for $\alpha < 1$, which is important for topic models,
where the prior is often sparse.

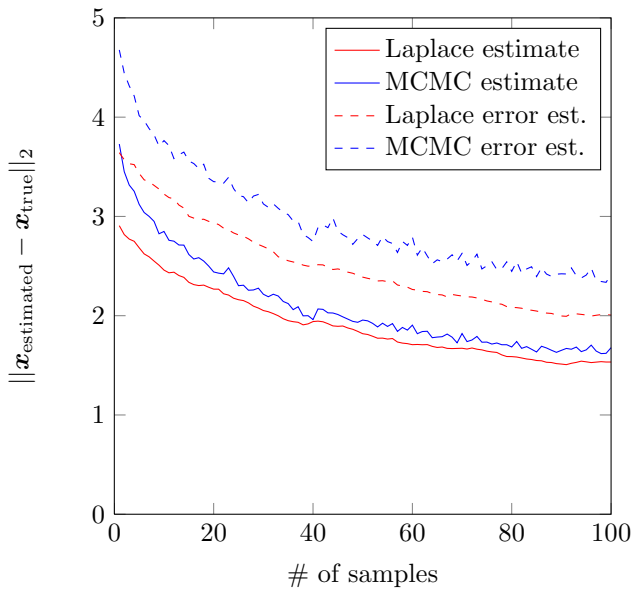
\begin{figure}[ht]
  \centering
   \beginpgfgraphicnamed{mean_error-external}%
   \input{mean_error.tikz}%
   \endpgfgraphicnamed%
 
\caption[Convergence behaviour of the Laplace bridge]{Convergence
  behaviour of approximate inference using the Laplace bridge compared
  to MCMC inference. Solid lines represent deviation of mean estimate
  (sample mean for MCMC) from ground truth, dashed lines the error
  estimate of the inference algorithm (one standard deviation). Both
  methods were initialised with a prior of $\mu=0,\sigma=1$. Plots are
  averages over 12 independent experiments.}
  \label{fig:Laplaceconvergence}
\end{figure}

While it has previously been investigated in \cite{mackay1998choice},
the use of this approximation here differs considerably from the
setting studied in the cited paper (which dealt with evidence
estimation in neural networks). Its use here amounts to
the following:
\begin{itemize}
\item Some unobserved process with known parameters
  $\vec{\mu},\vec{\sigma}$ generates data as follows:
  \begin{itemize}
  \item Sample $\vec{x}\in\Re^{K} \sim
    \N(\vec{x};\vec{\mu},\vec{\Sigma})\N(0;\vec{1}\Trans\vec{x},\epsilon^{2})$
  \item Map $\vec{\pi} = \sigma(\vec{x})$
  \item Sample data $c$ from $p(c=k\g \vec{\pi}) = \vec{\pi}$
  \end{itemize}
\item The inference method tries to infer $\vec{x}$ thus:
  \begin{itemize}
  \item Use the Laplace map to gain a Dirichlet belief on $\vec{\pi}$
    from the Gaussian prior \eqref{eq:4}
  \item Update this belief using the data (which is trivial, due to
    the Dirichlet's conjugacy to the Multinomial distribution)
  \item Use the Laplace map in the opposite direction, to get a
    Gaussian belief on $\Re^k$, claim the resulting belief to be
    an approximate posterior on $\vec{x}$
  \end{itemize}
\end{itemize}
Figure \ref{fig:Laplaceconvergence} compares this approximate scheme
to an asymptotically exact Markov Chain Monte Carlo scheme (the
particular MCMC method chosen for this task is elliptical slice
sampling \citep{murray-adams-mackay-2010a}, which has the advantage of
having no free parameters). The figure shows the $2$-norm error of a
point estimate for $\vec{x}$ returned by the two methods (solid lines)
and error estimates constructed from the algorithms' results. For the
MCMC sampler, these two estimates are the sample mean and (unbiased)
sample covariance. For the Laplace approximations, the two estimates
are the mean and standard deviation of the approximate Gaussian
belief. The prior mean and covariance were sampled, for each
experiment separately, from the standard Gaussian and the standard
inverse Wishart distribution, respectively. The number of dimensions
was set to $K=10$. Note that the Laplace bridge does not show any
discernible bias or over-convergence. Its only two apparent drawbacks
are its relatively bad fit for $\alpha\to 0$ and that covariance can
not be captured by the Dirichlet.

\subsection{The Wider View}
\label{sec:wider-view}

Within the wider context of unsupervised learning methods, the kernel
topic model establishes a connection between conditional topic models
and Gaussian process latent variable models (GPLVM)
\citep{lawrence2004gaussian}. GPLVMs learn mappings from data-space to
a lower-dimensional space, assuming the generative model for the data
in the latent space is a Gaussian process. In the case of the kernel
topic model, the variational part of the inference learns a mapping
from the $V$-dimensional space of documents defined by their words to
the $K$-dimensional space of topics defined by their topics (Figure
\ref{fig:dim-red}), where the documents' topics are assumed to be
generated by a Gaussian process.  However, in GPLVMs the map between
data and their low-dimensional representation is usually assumed to be
generated by another Gaussian process. In the kernel topic model, the
lower dimensional distributions are discrete, and sampled from
Dirichlet distributions. The kernel topic model thus performs Gaussian
process regression, under a ``latent Dirichlet likelihood''.

\section{Experiments}
\label{sec:experiments}

\subsection{Euclidean and Discrete Spaces}
\label{sec:comparative}
We compare the kernel topic model to its conceptually closest
competitor, the Dirichlet-Multinomial Regression (DMR) model by
\citet{mimno-topic}, which was, in the cited work, shown to give
superior results to a number of other models, such as topics through
time \citep{Wang:2006:TOT:1150402.1150450} and the author topic model
\citep{rosen2004author}. The dataset consists of the annual State Of The
Union addresses by US presidents to the joint chambers of Congress,
annotated with both the speaker's identity and the year of
delivery. This dataset is interesting because it combines continuous
features (time) with 44 discrete ones (author identity) and thus falls
outside of the descriptive power of time drift models like the one by
\citet{wang2009continuous}. All models used $K=10$ topics.

For the linear model of DMR, we represented time using 100 radial
basis functions spaced evenly through the time period from years 1790
to 2011, each with a width of 5 years, and used 44 binary author
indicator features. For the kernel topic model, we used a rational
quadratic kernel \citep{matern1960spatial,RasmussenWilliams} on the
Hilbert space of time and author identity, assigning a distance
between documents linear in time (initially using the same scale of 5
years), with an additional constant term if the authors of two
documents are not the same.  The rational quadratic kernel is
equivalent to an infinite scale mixture of square exponential kernels:
It assigns nonzero mass to functions with a range of length scales,
while the the square exponential (for which the radial basis functions
of the linear model are a finite-dimensional approximation) can only
construct functions of a single length scale. So the kernel model is
strictly more expressive than the linear model in this case. In
addition, the evidence maximisation description as introduced in
Section \ref{sec:appr-gauss-proc} allows an optimisation of the kernel
parameters during training. For DMR, this would amount to optimising
the feature set, rather than the feature parameters, which is more
difficult to do efficiently.

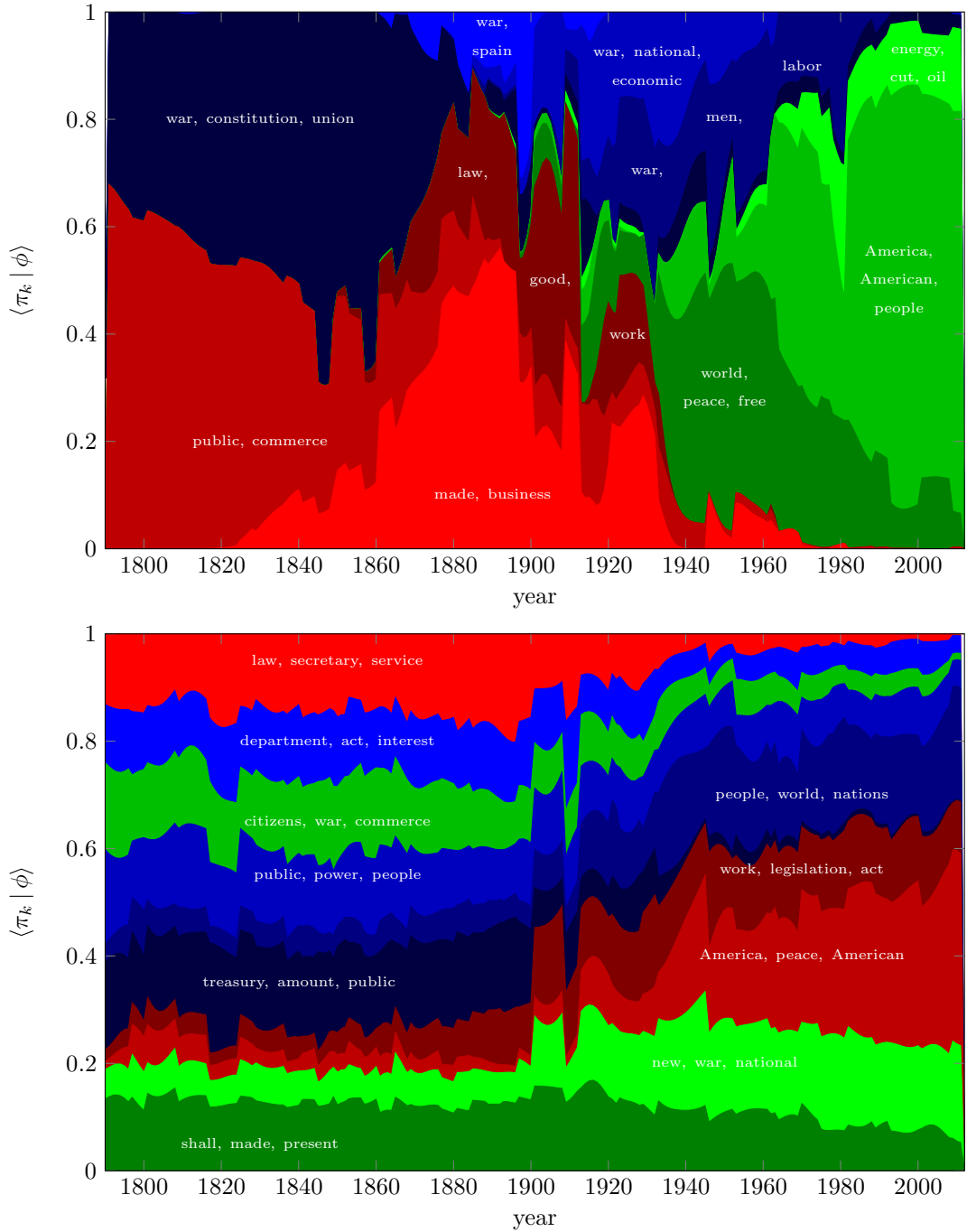
\begin{figure*}[t!]
  \centering
   \beginpgfgraphicnamed{StateOfTheUnion_stacked-external}%
   \input{StateOfTheUnion_stacked.tikz}%
   \endpgfgraphicnamed%
 \\
   \beginpgfgraphicnamed{MimnoStateOfTheUnion_stacked-external}%
   \input{MimnoStateOfTheUnion_stacked.tikz}%
   \endpgfgraphicnamed%
 
\caption{Inferred topic distribution of State Of The Union addresses
  by US American presidents. {\bfseries Top:} kernel topic model using
  a rational quadratic kernel on the 45 dimensional space of authors
  and time; {\bfseries Bottom:} Linear model using 100 radial basis
  functions in time and 44 binary author features. To generate this
  plot, either model was used to predict the topic distributions at
  the given date, conditioned on the author being the president in
  office at that time.}
  \label{fig:stateoftheunion}
\end{figure*}

Figure \ref{fig:stateoftheunion} shows the
consequences of this additional expressive power: The kernel model
captures interesting detail in the development of American interior
and foreign policy, including long-term developments like the
industrial revolution (bright red topic at bottom of plot) and
faster developments like the Spanish-American war (light blue, top). 

\begin{figure}[t!]
  \centering
   \beginpgfgraphicnamed{perplexity-external}%
   \input{perplexity.tikz}%
   \endpgfgraphicnamed%

   \beginpgfgraphicnamed{wiki_perplexity-external}%
   \input{wiki_perplexity.tikz}%
   \endpgfgraphicnamed%

   \beginpgfgraphicnamed{NIPS_perplexity-external}%
   \input{NIPS_perplexity.tikz}%
   \endpgfgraphicnamed%
 
  \caption{Perplexity of kernel topic model (blue) and linear
    maximum-likelihood Gaussian model or a constant LDA model
    (red). KTM \emph{hyper}-parameters were optimised after every 10
    iterations (the kernel regression itself is updated after every
    document inference). {\bfseries Top:} State Of The Union
    dataset. Here, the hyperparameters happened to be chosen well,
    optimising them had negligible effect on perplexity. {\bfseries
      Middle:} Wiki documents (Section \ref{sec:topics-graphs}). Note
    the spike in the perplexity of the kernel model in the latter
    plot, caused by the optimisation of hyperparameters -- since the
    optimisation is not performed directly on the word level, the
    topic model crosses over into a more perplexed state at this
    point, but this subsequently allows a better
    representation. {\bfseries Bottom:} NIPS dataset
    \citep{chechik2007eec}, again showing considerable improvement
    after hyperparameter optimisation.}
  \label{fig:perplexity}
\end{figure}
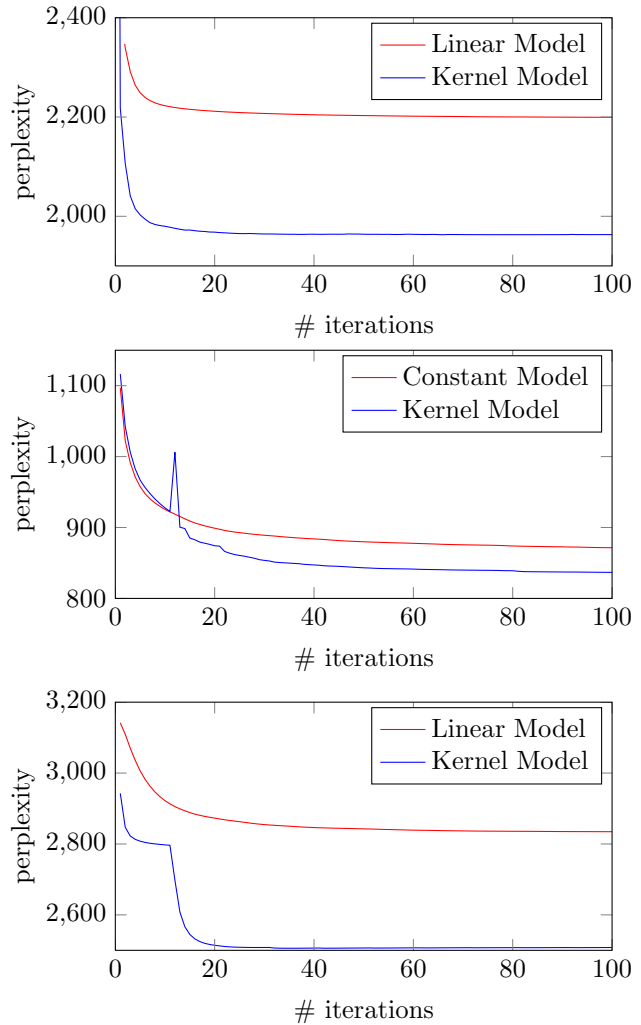

Figure \ref{fig:perplexity} shows the development of the perplexity
score \citep{rosen2004author} of the two models, on the training set,
during training on three different datasets (see caption for
details on datasets). (The vocabulary size for this dataset is $V=5000$, so the
initial perplexity is 5000.)  Optimisation of kernel hyperparameters
was performed every tenth variational loop, and is visible as a
discrete steps in the plots when it has non-negligible effect, thus
also giving an intuition for the model performance without
hyper-optimisation.

The kernel topic model converges about as fast as the DMR, but
achieves a final score about $12\%$ below that of DMR. The two
methods' runtimes are roughly comparable on our datasets: Both
models share the LDA part. In the regression part, DMR requires
numerical optimisation of the feature weights, while the kernel topic
model requires inverting a large matrix.

\subsection{Topics on Graphs}
\label{sec:topics-graphs}
The kernel view on topic models also allows a relatively elegant
treatment of non-Euclidean feature spaces. As an example, we construct
a topic model on a graph. For our experiment, $D=318$ documents were
taken from Wikipedia's ``list of probability topics\footnote{{\tiny\tt
    http://en.wikipedia.org/wiki/List\_of\_probability\_topics}}''. We
construct a positive definite kernel by embedding the documents in the
$\Re^D$ Euclidean vector space, setting
\begin{equation}
  k(d_1,d_2) = s \exp(-\frac{1}{2}(\vec{x}_1-\vec{x}_2)\Trans \vec{S} (\vec{x}_1-\vec{x}_2))
\label{eq:21}
\end{equation}
where the vector elements $x_{d,j}$ are the shortest distances, on the
graph of links between documents, from document $d$ to document $i$
(links are interpreted as undirected edges, documents not linked by
any path are assigned infinite distance), $s$ and
$\vec{S}=\operatorname{diag}_i(S_i)$ are parameters. Of course it is
possible to define corresponding linear features, but the kernel view
arguably allows a more natural way of deriving such measures.

\section{Conclusion}
\label{sec:conclusion}

We have presented the kernel topic model, allowing nonparametric
regression of topics on document metadata of various kinds. The model
is a combination of Gaussian process regression and latent Dirichlet
allocation; these two conditionally independent parts are linked
efficiently through a lightweight Laplace approximation. Inference in
the kernel topic model is cubic in the number of documents. In large
corpora, this can compare unfavourably to other feature-based topic
models, but it offers superior power of expression for small and
medium-sized corpora, where (approximate) analytic Gaussian process
inference can even be faster than EM optimization of point
estimates. An elegant side-effect of the Laplace approximation, which
we have only touched upon marginally in this paper, is that it
replaces the point estimates of earlier approaches with a full
Bayesian belief. This means that topics can be predicted with
uncertainty, and that hyperparameters of the model can be inferred
consistently, using higher order maximum likelihood (maximum evidence)
optimisation.
	
\subsubsection*{Acknowledgements}

We would like to thank David MacKay and David Knowles for helpful
discussions and comments.

\newpage
\bibliographystyle{plainnat}
\bibliography{../bibfile}

\end{document}

%% file: simplex_plots.tikz
%
%
\begin{tikzpicture}

\begin{axis}[%
scale only axis,
width=160pt, 
height=200pt, 
xmin=0, xmax=1,
ymin=0, ymax=2.2,
xlabel={$\pi$},
ylabel={$p(\pi)\d\pi$},
axis on top,
legend entries={$\D$,$\N$},
legend style={nodes=right}]
\addplot [
color=black, 
solid
]
coordinates{
 (0.001,0.00263947)(0.0110808,0.0291882)(0.0211616,0.0556282)(0.0312424,0.0819581)(0.0413232,0.108176)(0.051404,0.134282)(0.0614848,0.160273)(0.0715657,0.186148)(0.0816465,0.211906)(0.0917273,0.237545)(0.101808,0.263063)(0.111889,0.288459)(0.12197,0.313731)(0.132051,0.338878)(0.142131,0.363896)(0.152212,0.388786)(0.162293,0.413544)(0.172374,0.438169)(0.182455,0.462659)(0.192535,0.487011)(0.202616,0.511224)(0.212697,0.535296)(0.222778,0.559223)(0.232859,0.583004)(0.242939,0.606636)(0.25302,0.630117)(0.263101,0.653443)(0.273182,0.676614)(0.283263,0.699625)(0.293343,0.722474)(0.303424,0.745157)(0.313505,0.767672)(0.323586,0.790016)(0.333667,0.812185)(0.343747,0.834176)(0.353828,0.855985)(0.363909,0.877608)(0.37399,0.899042)(0.384071,0.920283)(0.394152,0.941326)(0.404232,0.962167)(0.414313,0.982801)(0.424394,1.00322)(0.434475,1.02343)(0.444556,1.04342)(0.454636,1.06318)(0.464717,1.0827)(0.474798,1.10199)(0.484879,1.12103)(0.49496,1.13983)(0.50504,1.15836)(0.515121,1.17663)(0.525202,1.19463)(0.535283,1.21234)(0.545364,1.22977)(0.555444,1.24689)(0.565525,1.26371)(0.575606,1.28022)(0.585687,1.29639)(0.595768,1.31222)(0.605848,1.3277)(0.615929,1.34282)(0.62601,1.35755)(0.636091,1.3719)(0.646172,1.38583)(0.656253,1.39934)(0.666333,1.4124)(0.676414,1.425)(0.686495,1.43711)(0.696576,1.44871)(0.706657,1.45978)(0.716737,1.47029)(0.726818,1.4802)(0.736899,1.48949)(0.74698,1.49811)(0.757061,1.50603)(0.767141,1.51321)(0.777222,1.51958)(0.787303,1.5251)(0.797384,1.5297)(0.807465,1.53331)(0.817545,1.53584)(0.827626,1.53721)(0.837707,1.53729)(0.847788,1.53596)(0.857869,1.53307)(0.867949,1.52843)(0.87803,1.52182)(0.888111,1.51296)(0.898192,1.50151)(0.908273,1.48703)(0.918354,1.46893)(0.928434,1.44643)(0.938515,1.4184)(0.948596,1.3832)(0.958677,1.33818)(0.968758,1.2787)(0.978838,1.19516)(0.988919,1.06091)(0.999,0.662475) 
};

\addplot [
color=black, 
dashed
]
coordinates{
 (0.001,3.78628e-07)(0.0110808,0.00265201)(0.0211616,0.0140453)(0.0312424,0.0333272)(0.0413232,0.0582065)(0.051404,0.0868339)(0.0614848,0.117884)(0.0715657,0.150428)(0.0816465,0.183819)(0.0917273,0.217601)(0.101808,0.251456)(0.111889,0.285161)(0.12197,0.318563)(0.132051,0.351553)(0.142131,0.384061)(0.152212,0.416041)(0.162293,0.447467)(0.172374,0.478326)(0.182455,0.508614)(0.192535,0.538336)(0.202616,0.5675)(0.212697,0.596119)(0.222778,0.62421)(0.232859,0.65179)(0.242939,0.678876)(0.25302,0.705488)(0.263101,0.731644)(0.273182,0.757365)(0.283263,0.782668)(0.293343,0.807572)(0.303424,0.832095)(0.313505,0.856254)(0.323586,0.880065)(0.333667,0.903545)(0.343747,0.926707)(0.353828,0.949566)(0.363909,0.972135)(0.37399,0.994426)(0.384071,1.01645)(0.394152,1.03822)(0.404232,1.05974)(0.414313,1.08103)(0.424394,1.10208)(0.434475,1.12291)(0.444556,1.14353)(0.454636,1.16393)(0.464717,1.18412)(0.474798,1.2041)(0.484879,1.22388)(0.49496,1.24345)(0.50504,1.2628)(0.515121,1.28195)(0.525202,1.30088)(0.535283,1.31959)(0.545364,1.33806)(0.555444,1.35629)(0.565525,1.37427)(0.575606,1.39197)(0.585687,1.40939)(0.595768,1.42649)(0.605848,1.44327)(0.615929,1.45968)(0.62601,1.47569)(0.636091,1.49128)(0.646172,1.50639)(0.656253,1.52099)(0.666333,1.53501)(0.676414,1.5484)(0.686495,1.56108)(0.696576,1.57298)(0.706657,1.584)(0.716737,1.59405)(0.726818,1.60301)(0.736899,1.61075)(0.74698,1.61711)(0.757061,1.62192)(0.767141,1.625)(0.777222,1.6261)(0.787303,1.62498)(0.797384,1.62133)(0.807465,1.61482)(0.817545,1.60503)(0.827626,1.59152)(0.837707,1.57373)(0.847788,1.55105)(0.857869,1.52274)(0.867949,1.48794)(0.87803,1.44563)(0.888111,1.39463)(0.898192,1.33354)(0.908273,1.26072)(0.918354,1.17425)(0.928434,1.07198)(0.938515,0.951547)(0.948596,0.810609)(0.958677,0.64752)(0.968758,0.463042)(0.978838,0.265119)(0.988919,0.0828308)(0.999,7.52654e-05) 
};

\addplot [
color=black, 
solid
]
coordinates{
 (0.0413232,2.31412)(0.051404,2.07704)(0.0614848,1.90118)(0.0715657,1.7641)(0.0816465,1.65341)(0.0917273,1.56164)(0.101808,1.48396)(0.111889,1.41713)(0.12197,1.35886)(0.132051,1.30747)(0.142131,1.26172)(0.152212,1.22067)(0.162293,1.18356)(0.172374,1.14983)(0.182455,1.11898)(0.192535,1.09064)(0.202616,1.0645)(0.212697,1.04029)(0.222778,1.0178)(0.232859,0.996821)(0.242939,0.977212)(0.25302,0.958832)(0.263101,0.941562)(0.273182,0.9253)(0.283263,0.909956)(0.293343,0.895451)(0.303424,0.881717)(0.313505,0.86869)(0.323586,0.856318)(0.333667,0.84455)(0.343747,0.833344)(0.353828,0.822659)(0.363909,0.812461)(0.37399,0.802718)(0.384071,0.7934)(0.394152,0.784482)(0.404232,0.775939)(0.414313,0.76775)(0.424394,0.759896)(0.434475,0.752356)(0.444556,0.745116)(0.454636,0.73816)(0.464717,0.731473)(0.474798,0.725043)(0.484879,0.718859)(0.49496,0.712908)(0.50504,0.707182)(0.515121,0.70167)(0.525202,0.696365)(0.535283,0.691258)(0.545364,0.686343)(0.555444,0.681613)(0.565525,0.677062)(0.575606,0.672684)(0.585687,0.668475)(0.595768,0.66443)(0.605848,0.660545)(0.615929,0.656817)(0.62601,0.653242)(0.636091,0.649819)(0.646172,0.646544)(0.656253,0.643416)(0.666333,0.640433)(0.676414,0.637596)(0.686495,0.634904)(0.696576,0.632356)(0.706657,0.629954)(0.716737,0.6277)(0.726818,0.625595)(0.736899,0.623641)(0.74698,0.621843)(0.757061,0.620206)(0.767141,0.618734)(0.777222,0.617435)(0.787303,0.616317)(0.797384,0.615389)(0.807465,0.614665)(0.817545,0.614157)(0.827626,0.613885)(0.837707,0.613868)(0.847788,0.614133)(0.857869,0.614712)(0.867949,0.615644)(0.87803,0.61698)(0.888111,0.618783)(0.898192,0.621138)(0.908273,0.624156)(0.918354,0.627989)(0.928434,0.632856)(0.938515,0.639078)(0.948596,0.647158)(0.958677,0.657954)(0.968758,0.673084)(0.978838,0.696211)(0.988919,0.738947)(0.999,0.935122) 
};

\addplot [
color=black, 
dashed
]
coordinates{
 (0.001,0.369764)(0.0110808,1.78294)(0.0211616,2.00722)(0.0312424,2.03224)(0.0413232,1.99741)(0.051404,1.94209)(0.0614848,1.88053)(0.0715657,1.81851)(0.0816465,1.75852)(0.0917273,1.70157)(0.101808,1.64799)(0.111889,1.59782)(0.12197,1.55091)(0.132051,1.50708)(0.142131,1.46611)(0.152212,1.42777)(0.162293,1.39187)(0.172374,1.35819)(0.182455,1.32657)(0.192535,1.29683)(0.202616,1.26883)(0.212697,1.24242)(0.222778,1.2175)(0.232859,1.19394)(0.242939,1.17165)(0.25302,1.15054)(0.263101,1.13052)(0.273182,1.11152)(0.283263,1.09347)(0.293343,1.07632)(0.303424,1.05999)(0.313505,1.04445)(0.323586,1.02965)(0.333667,1.01554)(0.343747,1.00207)(0.353828,0.98923)(0.363909,0.976967)(0.37399,0.965256)(0.384071,0.954068)(0.394152,0.943376)(0.404232,0.933157)(0.414313,0.923386)(0.424394,0.914043)(0.434475,0.905109)(0.444556,0.896566)(0.454636,0.888396)(0.464717,0.880583)(0.474798,0.873113)(0.484879,0.865972)(0.49496,0.859148)(0.50504,0.852627)(0.515121,0.846399)(0.525202,0.840453)(0.535283,0.834779)(0.545364,0.829367)(0.555444,0.824209)(0.565525,0.819295)(0.575606,0.814619)(0.585687,0.810171)(0.595768,0.805944)(0.605848,0.801932)(0.615929,0.798128)(0.62601,0.794523)(0.636091,0.791113)(0.646172,0.787889)(0.656253,0.784846)(0.666333,0.781975)(0.676414,0.779271)(0.686495,0.776725)(0.696576,0.77433)(0.706657,0.772076)(0.716737,0.769955)(0.726818,0.767955)(0.736899,0.766065)(0.74698,0.764271)(0.757061,0.762559)(0.767141,0.760908)(0.777222,0.759299)(0.787303,0.757704)(0.797384,0.756095)(0.807465,0.754433)(0.817545,0.752673)(0.827626,0.750758)(0.837707,0.748618)(0.847788,0.746166)(0.857869,0.743288)(0.867949,0.73984)(0.87803,0.735633)(0.888111,0.730418)(0.898192,0.723858)(0.908273,0.715492)(0.918354,0.704671)(0.928434,0.690458)(0.938515,0.671449)(0.948596,0.645453)(0.958677,0.608842)(0.968758,0.555141)(0.978838,0.471402)(0.988919,0.326648)(0.999,0.0271957) 
};

\addplot [
color=black, 
solid
]
coordinates{
 (0.001,5.98202e-05)(0.0110808,0.00712486)(0.0211616,0.0251989)(0.0312424,0.0532459)(0.0413232,0.0902727)(0.051404,0.135329)(0.0614848,0.187504)(0.0715657,0.245932)(0.0816465,0.309782)(0.0917273,0.378266)(0.101808,0.450633)(0.111889,0.526171)(0.12197,0.604204)(0.132051,0.684092)(0.142131,0.765232)(0.152212,0.847054)(0.162293,0.929024)(0.172374,1.01064)(0.182455,1.09143)(0.192535,1.17096)(0.202616,1.24883)(0.212697,1.32465)(0.222778,1.39808)(0.232859,1.4688)(0.242939,1.53652)(0.25302,1.60099)(0.263101,1.66196)(0.273182,1.71922)(0.283263,1.7726)(0.293343,1.82192)(0.303424,1.86705)(0.313505,1.90789)(0.323586,1.94432)(0.333667,1.97629)(0.343747,2.00375)(0.353828,2.02666)(0.363909,2.04501)(0.37399,2.0588)(0.384071,2.06807)(0.394152,2.07286)(0.404232,2.07321)(0.414313,2.06921)(0.424394,2.06094)(0.434475,2.0485)(0.444556,2.03201)(0.454636,2.01158)(0.464717,1.98737)(0.474798,1.95951)(0.484879,1.92817)(0.49496,1.89352)(0.50504,1.85572)(0.515121,1.81497)(0.525202,1.77146)(0.535283,1.72538)(0.545364,1.67693)(0.555444,1.62634)(0.565525,1.5738)(0.575606,1.51953)(0.585687,1.46375)(0.595768,1.40669)(0.605848,1.34856)(0.615929,1.28957)(0.62601,1.22997)(0.636091,1.16995)(0.646172,1.10975)(0.656253,1.04957)(0.666333,0.98963)(0.676414,0.930133)(0.686495,0.871284)(0.696576,0.813278)(0.706657,0.756305)(0.716737,0.70055)(0.726818,0.646187)(0.736899,0.593382)(0.74698,0.542294)(0.757061,0.493068)(0.767141,0.44584)(0.777222,0.400735)(0.787303,0.357865)(0.797384,0.317328)(0.807465,0.279209)(0.817545,0.243579)(0.827626,0.210491)(0.837707,0.179984)(0.847788,0.15208)(0.857869,0.126783)(0.867949,0.104078)(0.87803,0.0839317)(0.888111,0.0662898)(0.898192,0.0510783)(0.908273,0.0382014)(0.918354,0.0275412)(0.928434,0.0189569)(0.938515,0.012284)(0.948596,0.0073334)(0.958677,0.00389115)(0.968758,0.00171718)(0.978838,0.000544778)(0.988919,7.98339e-05)(0.999,5.98801e-08) 
};

\addplot [
color=black, 
dashed
]
coordinates{
 (0.001,2.56577e-14)(0.0110808,1.25911e-05)(0.0211616,0.000524371)(0.0312424,0.00355952)(0.0413232,0.0120996)(0.051404,0.028825)(0.0614848,0.0555198)(0.0715657,0.0930067)(0.0816465,0.141307)(0.0917273,0.199852)(0.101808,0.267684)(0.111889,0.343611)(0.12197,0.426327)(0.132051,0.514494)(0.142131,0.606804)(0.152212,0.702011)(0.162293,0.798957)(0.172374,0.896582)(0.182455,0.99393)(0.192535,1.09015)(0.202616,1.18449)(0.212697,1.27629)(0.222778,1.36499)(0.232859,1.4501)(0.242939,1.53121)(0.25302,1.60798)(0.263101,1.68014)(0.273182,1.74745)(0.283263,1.80974)(0.293343,1.86689)(0.303424,1.91879)(0.313505,1.96541)(0.323586,2.0067)(0.333667,2.04268)(0.343747,2.07337)(0.353828,2.09882)(0.363909,2.11909)(0.37399,2.13428)(0.384071,2.14448)(0.394152,2.14979)(0.404232,2.15034)(0.414313,2.14627)(0.424394,2.13771)(0.434475,2.12481)(0.444556,2.10772)(0.454636,2.0866)(0.464717,2.06163)(0.474798,2.03296)(0.484879,2.00077)(0.49496,1.96524)(0.50504,1.92654)(0.515121,1.88486)(0.525202,1.84039)(0.535283,1.79331)(0.545364,1.7438)(0.555444,1.69207)(0.565525,1.63831)(0.575606,1.58271)(0.585687,1.52546)(0.595768,1.46677)(0.605848,1.40684)(0.615929,1.34587)(0.62601,1.28406)(0.636091,1.22162)(0.646172,1.15876)(0.656253,1.09567)(0.666333,1.03258)(0.676414,0.969693)(0.686495,0.907214)(0.696576,0.845358)(0.706657,0.784336)(0.716737,0.724359)(0.726818,0.665635)(0.736899,0.608373)(0.74698,0.552779)(0.757061,0.499053)(0.767141,0.447392)(0.777222,0.397988)(0.787303,0.351022)(0.797384,0.306666)(0.807465,0.265081)(0.817545,0.226411)(0.827626,0.190781)(0.837707,0.158296)(0.847788,0.129031)(0.857869,0.103035)(0.867949,0.0803152)(0.87803,0.060841)(0.888111,0.0445326)(0.898192,0.0312575)(0.908273,0.0208254)(0.918354,0.0129852)(0.928434,0.00742461)(0.938515,0.00377524)(0.948596,0.00162548)(0.958677,0.000544436)(0.968758,0.000120309)(0.978838,1.19461e-05)(0.988919,1.5001e-07)(0.999,2.82248e-17) 
};

\end{axis}
\end{tikzpicture}

%% file: real_plots.tikz
%
%
\begin{tikzpicture}

\begin{axis}[%
scale only axis,
width=160pt, 
height=200pt, 
xmin=-5, xmax=5,
ymin=0, ymax=0.6,
xlabel={$x$},
ylabel={$p(x)\d x$},
axis on top,
legend entries={$\D$,$\N$},
legend style={nodes=right}]
\addplot [
color=black, 
solid
]
coordinates{
 (-5,0.000117308)(-4.89899,0.000143243)(-4.79798,0.000174871)(-4.69697,0.000213426)(-4.59596,0.000260404)(-4.49495,0.000317621)(-4.39394,0.000387272)(-4.29293,0.00047201)(-4.19192,0.000575039)(-4.09091,0.000700222)(-3.9899,0.000852205)(-3.88889,0.00103657)(-3.78788,0.00126002)(-3.68687,0.00153054)(-3.58586,0.00185771)(-3.48485,0.00225287)(-3.38384,0.00272951)(-3.28283,0.00330357)(-3.18182,0.00399382)(-3.08081,0.00482225)(-2.9798,0.00581455)(-2.87879,0.0070005)(-2.77778,0.00841448)(-2.67677,0.0100959)(-2.57576,0.0120894)(-2.47475,0.0144456)(-2.37374,0.0172206)(-2.27273,0.0204766)(-2.17172,0.0242809)(-2.07071,0.0287057)(-1.9697,0.0338269)(-1.86869,0.0397221)(-1.76768,0.0464685)(-1.66667,0.0541401)(-1.56566,0.0628036)(-1.46465,0.072514)(-1.36364,0.08331)(-1.26263,0.0952079)(-1.16162,0.108196)(-1.06061,0.122229)(-0.959596,0.137224)(-0.858586,0.153055)(-0.757576,0.169551)(-0.656566,0.186497)(-0.555556,0.203637)(-0.454545,0.220677)(-0.353535,0.237297)(-0.252525,0.253158)(-0.151515,0.267919)(-0.0505051,0.281249)(0.0505051,0.292845)(0.151515,0.302444)(0.252525,0.309833)(0.353535,0.314864)(0.454545,0.317456)(0.555556,0.317597)(0.656566,0.315346)(0.757576,0.31082)(0.858586,0.304194)(0.959596,0.295684)(1.06061,0.28554)(1.16162,0.27403)(1.26263,0.261429)(1.36364,0.248012)(1.46465,0.234041)(1.56566,0.21976)(1.66667,0.20539)(1.76768,0.191123)(1.86869,0.177125)(1.9697,0.163533)(2.07071,0.150455)(2.17172,0.137974)(2.27273,0.126149)(2.37374,0.115019)(2.47475,0.104605)(2.57576,0.0949108)(2.67677,0.0859307)(2.77778,0.0776473)(2.87879,0.0700362)(2.9798,0.0630672)(3.08081,0.0567064)(3.18182,0.0509173)(3.28283,0.0456621)(3.38384,0.0409026)(3.48485,0.0366013)(3.58586,0.0327214)(3.68687,0.0292277)(3.78788,0.0260868)(3.88889,0.0232669)(3.9899,0.0207385)(4.09091,0.0184741)(4.19192,0.0164482)(4.29293,0.0146375)(4.39394,0.0130205)(4.49495,0.0115775)(4.59596,0.0102908)(4.69697,0.00914411)(4.79798,0.00812283)(4.89899,0.0072137)(5,0.00640478) 
};

\addplot [
color=black, 
dashed
]
coordinates{
 (-5,3.91291e-06)(-4.89899,5.91767e-06)(-4.79798,8.88132e-06)(-4.69697,1.32276e-05)(-4.59596,1.95507e-05)(-4.49495,2.8676e-05)(-4.39394,4.174e-05)(-4.29293,6.02925e-05)(-4.19192,8.64272e-05)(-4.09091,0.000122946)(-3.9899,0.000173562)(-3.88889,0.000243149)(-3.78788,0.000338038)(-3.68687,0.000466376)(-3.58586,0.000638533)(-3.48485,0.000867575)(-3.38384,0.00116979)(-3.28283,0.00156525)(-3.18182,0.00207844)(-3.08081,0.00273885)(-2.9798,0.00358159)(-2.87879,0.00464792)(-2.77778,0.00598576)(-2.67677,0.00764991)(-2.57576,0.0097022)(-2.47475,0.0122113)(-2.37374,0.015252)(-2.27273,0.0189047)(-2.17172,0.0232537)(-2.07071,0.0283849)(-1.9697,0.0343844)(-1.86869,0.0413344)(-1.76768,0.0493104)(-1.66667,0.0583771)(-1.56566,0.068584)(-1.46465,0.0799612)(-1.36364,0.0925152)(-1.26263,0.106224)(-1.16162,0.121035)(-1.06061,0.136859)(-0.959596,0.153573)(-0.858586,0.171014)(-0.757576,0.188984)(-0.656566,0.20725)(-0.555556,0.225549)(-0.454545,0.243593)(-0.353535,0.261074)(-0.252525,0.277678)(-0.151515,0.293086)(-0.0505051,0.30699)(0.0505051,0.319103)(0.151515,0.329166)(0.252525,0.336957)(0.353535,0.342304)(0.454545,0.345084)(0.555556,0.345235)(0.656566,0.342753)(0.757576,0.337695)(0.858586,0.330175)(0.959596,0.320362)(1.06061,0.308471)(1.16162,0.294757)(1.26263,0.279506)(1.36364,0.263024)(1.46465,0.245626)(1.56566,0.227631)(1.66667,0.209346)(1.76768,0.191062)(1.86869,0.173046)(1.9697,0.155533)(2.07071,0.138728)(2.17172,0.122795)(2.27273,0.107863)(2.37374,0.094025)(2.47475,0.0813373)(2.57576,0.0698253)(2.67677,0.0594857)(2.77778,0.0502909)(2.87879,0.0421932)(2.9798,0.0351295)(3.08081,0.0290254)(3.18182,0.0237992)(3.28283,0.0193652)(3.38384,0.0156372)(3.48485,0.0125306)(3.58586,0.00996461)(3.68687,0.0078637)(3.78788,0.00615843)(3.88889,0.00478619)(3.9899,0.00369136)(4.09091,0.00282526)(4.19192,0.0021459)(4.29293,0.00161747)(4.39394,0.00120987)(4.49495,0.000898085)(4.59596,0.000661567)(4.69697,0.000483623)(4.79798,0.000350846)(4.89899,0.000252582)(5,0.000180454) 
};

\addplot [
color=black, 
solid
]
coordinates{
 (-5,0.0380918)(-4.89899,0.0400252)(-4.79798,0.0420523)(-4.69697,0.0441769)(-4.59596,0.0464029)(-4.49495,0.0487342)(-4.39394,0.0511747)(-4.29293,0.0537281)(-4.19192,0.0563981)(-4.09091,0.0591885)(-3.9899,0.0621025)(-3.88889,0.0651434)(-3.78788,0.068314)(-3.68687,0.0716167)(-3.58586,0.0750535)(-3.48485,0.0786258)(-3.38384,0.0823341)(-3.28283,0.0861782)(-3.18182,0.0901569)(-3.08081,0.0942677)(-2.9798,0.0985068)(-2.87879,0.102869)(-2.77778,0.107347)(-2.67677,0.111931)(-2.57576,0.116611)(-2.47475,0.121373)(-2.37374,0.1262)(-2.27273,0.131073)(-2.17172,0.135969)(-2.07071,0.140863)(-1.9697,0.145726)(-1.86869,0.150526)(-1.76768,0.155226)(-1.66667,0.159789)(-1.56566,0.164173)(-1.46465,0.168332)(-1.36364,0.172221)(-1.26263,0.175793)(-1.16162,0.178998)(-1.06061,0.18179)(-0.959596,0.184122)(-0.858586,0.185951)(-0.757576,0.187239)(-0.656566,0.187952)(-0.555556,0.188063)(-0.454545,0.187552)(-0.353535,0.186409)(-0.252525,0.184633)(-0.151515,0.182233)(-0.0505051,0.179226)(0.0505051,0.175642)(0.151515,0.171516)(0.252525,0.166894)(0.353535,0.161827)(0.454545,0.156372)(0.555556,0.150589)(0.656566,0.144541)(0.757576,0.138291)(0.858586,0.131901)(0.959596,0.125431)(1.06061,0.118939)(1.16162,0.112475)(1.26263,0.106086)(1.36364,0.0998157)(1.46465,0.0936983)(1.56566,0.0877644)(1.66667,0.0820386)(1.76768,0.0765401)(1.86869,0.0712831)(1.9697,0.0662774)(2.07071,0.0615288)(2.17172,0.0570392)(2.27273,0.0528079)(2.37374,0.0488313)(2.47475,0.0451039)(2.57576,0.0416185)(2.67677,0.0383663)(2.77778,0.0353378)(2.87879,0.0325227)(2.9798,0.0299104)(3.08081,0.0274898)(3.18182,0.02525)(3.28283,0.0231799)(3.38384,0.021269)(3.48485,0.0195068)(3.58586,0.0178831)(3.68687,0.0163885)(3.78788,0.0150137)(3.88889,0.01375)(3.9899,0.012589)(4.09091,0.0115232)(4.19192,0.0105452)(4.29293,0.00964812)(4.39394,0.00882571)(4.49495,0.00807201)(4.59596,0.00738151)(4.69697,0.00674913)(4.79798,0.00617014)(4.89899,0.00564016)(5,0.00515517) 
};

\addplot [
color=black, 
dashed
]
coordinates{
 (-5,0.00990066)(-4.89899,0.0114069)(-4.79798,0.0130992)(-4.69697,0.0149933)(-4.59596,0.0171051)(-4.49495,0.0194505)(-4.39394,0.0220451)(-4.29293,0.024904)(-4.19192,0.0280415)(-4.09091,0.0314709)(-3.9899,0.035204)(-3.88889,0.0392511)(-3.78788,0.0436201)(-3.68687,0.0483167)(-3.58586,0.0533438)(-3.48485,0.058701)(-3.38384,0.0643848)(-3.28283,0.0703877)(-3.18182,0.0766984)(-3.08081,0.0833012)(-2.9798,0.0901761)(-2.87879,0.0972989)(-2.77778,0.104641)(-2.67677,0.112168)(-2.57576,0.119843)(-2.47475,0.127623)(-2.37374,0.135464)(-2.27273,0.143316)(-2.17172,0.151127)(-2.07071,0.158842)(-1.9697,0.166403)(-1.86869,0.173754)(-1.76768,0.180836)(-1.66667,0.18759)(-1.56566,0.193959)(-1.46465,0.199888)(-1.36364,0.205323)(-1.26263,0.210216)(-1.16162,0.214521)(-1.06061,0.218197)(-0.959596,0.221209)(-0.858586,0.223529)(-0.757576,0.225134)(-0.656566,0.226007)(-0.555556,0.226141)(-0.454545,0.225535)(-0.353535,0.224193)(-0.252525,0.22213)(-0.151515,0.219365)(-0.0505051,0.215925)(0.0505051,0.211844)(0.151515,0.207159)(0.252525,0.201914)(0.353535,0.196158)(0.454545,0.189942)(0.555556,0.183321)(0.656566,0.176351)(0.757576,0.16909)(0.858586,0.161598)(0.959596,0.153932)(1.06061,0.14615)(1.16162,0.138307)(1.26263,0.130456)(1.36364,0.122648)(1.46465,0.11493)(1.56566,0.107345)(1.66667,0.0999317)(1.76768,0.0927261)(1.86869,0.0857585)(1.9697,0.0790546)(2.07071,0.0726363)(2.17172,0.0665205)(2.27273,0.0607202)(2.37374,0.0552441)(2.47475,0.0500974)(2.57576,0.0452814)(2.67677,0.0407944)(2.77778,0.0366316)(2.87879,0.032786)(2.9798,0.029248)(3.08081,0.0260063)(3.18182,0.0230482)(3.28283,0.0203598)(3.38384,0.017926)(3.48485,0.0157315)(3.58586,0.0137604)(3.68687,0.0119969)(3.78788,0.0104251)(3.88889,0.00902964)(3.9899,0.00779534)(4.09091,0.00670773)(4.19192,0.00575296)(4.29293,0.00491794)(4.39394,0.00419035)(4.49495,0.00355872)(4.59596,0.0030124)(4.69697,0.0025416)(4.79798,0.00213736)(4.89899,0.00179153)(5,0.00149674) 
};

\addplot [
color=black, 
solid
]
coordinates{
 (-5,1.75113e-05)(-4.89899,2.35918e-05)(-4.79798,3.1767e-05)(-4.69697,4.27503e-05)(-4.59596,5.74941e-05)(-4.49495,7.72682e-05)(-4.39394,0.000103762)(-4.29293,0.000139221)(-4.19192,0.000186619)(-4.09091,0.000249892)(-3.9899,0.00033423)(-3.88889,0.000446464)(-3.78788,0.000595548)(-3.68687,0.000793186)(-3.58586,0.00105461)(-3.48485,0.00139958)(-3.38384,0.00185355)(-3.28283,0.00244922)(-3.18182,0.00322826)(-3.08081,0.00424348)(-2.9798,0.00556125)(-2.87879,0.00726432)(-2.77778,0.00945483)(-2.67677,0.0122575)(-2.57576,0.0158227)(-2.47475,0.0203293)(-2.37374,0.025986)(-2.27273,0.0330323)(-2.17172,0.041736)(-2.07071,0.052388)(-1.9697,0.0652932)(-1.86869,0.0807552)(-1.76768,0.0990557)(-1.66667,0.120427)(-1.56566,0.145018)(-1.46465,0.172857)(-1.36364,0.203807)(-1.26263,0.237532)(-1.16162,0.27346)(-1.06061,0.310767)(-0.959596,0.348377)(-0.858586,0.384994)(-0.757576,0.419157)(-0.656566,0.449324)(-0.555556,0.473991)(-0.454545,0.49181)(-0.353535,0.501718)(-0.252525,0.503046)(-0.151515,0.495595)(-0.0505051,0.479666)(0.0505051,0.456042)(0.151515,0.425917)(0.252525,0.390785)(0.353535,0.352307)(0.454545,0.31217)(0.555556,0.271954)(0.656566,0.233033)(0.757576,0.196501)(0.858586,0.163146)(0.959596,0.133445)(1.06061,0.107602)(1.16162,0.0855876)(1.26263,0.0672003)(1.36364,0.0521195)(1.46465,0.0399576)(1.56566,0.0303017)(1.66667,0.0227457)(1.76768,0.0169117)(1.86869,0.0124626)(1.9697,0.00910834)(2.07071,0.00660595)(2.17172,0.00475714)(2.27273,0.00340335)(2.37374,0.00242013)(2.47475,0.00171141)(2.57576,0.00120405)(2.67677,0.000843134)(2.77778,0.000587868)(2.87879,0.000408275)(2.9798,0.000282529)(3.08081,0.00019487)(3.18182,0.000134005)(3.28283,9.18995e-05)(3.38384,6.28669e-05)(3.48485,4.29088e-05)(3.58586,2.92263e-05)(3.68687,1.98696e-05)(3.78788,1.34853e-05)(3.88889,9.13826e-06)(3.9899,6.18379e-06)(4.09091,4.1792e-06)(4.19192,2.82116e-06)(4.29293,1.90243e-06)(4.39394,1.28167e-06)(4.49495,8.62719e-07)(4.59596,5.80261e-07)(4.69697,3.90006e-07)(4.79798,2.61963e-07)(4.89899,1.75856e-07)(5,1.1799e-07) 
};

\addplot [
color=black, 
dashed
]
coordinates{
 (-5,2.82968e-09)(-4.89899,6.34332e-09)(-4.79798,1.39734e-08)(-4.69697,3.02474e-08)(-4.59596,6.43398e-08)(-4.49495,1.34485e-07)(-4.39394,2.76232e-07)(-4.29293,5.57541e-07)(-4.19192,1.10582e-06)(-4.09091,2.15523e-06)(-3.9899,4.12769e-06)(-3.88889,7.76829e-06)(-3.78788,1.43664e-05)(-3.68687,2.61079e-05)(-3.58586,4.66232e-05)(-3.48485,8.18153e-05)(-3.38384,0.000141082)(-3.28283,0.000239063)(-3.18182,0.000398067)(-3.08081,0.000651336)(-2.9798,0.00104727)(-2.87879,0.00165467)(-2.77778,0.00256905)(-2.67677,0.00391954)(-2.57576,0.00587628)(-2.47475,0.00865712)(-2.37374,0.0125328)(-2.27273,0.017829)(-2.17172,0.0249235)(-2.07071,0.0342369)(-1.9697,0.0462152)(-1.86869,0.0613025)(-1.76768,0.0799053)(-1.66667,0.102347)(-1.56566,0.12882)(-1.46465,0.159328)(-1.36364,0.193644)(-1.26263,0.23127)(-1.16162,0.271419)(-1.06061,0.313014)(-0.959596,0.354725)(-0.858586,0.395024)(-0.757576,0.432273)(-0.656566,0.464833)(-0.555556,0.49118)(-0.454545,0.51002)(-0.353535,0.5204)(-0.252525,0.521785)(-0.151515,0.514102)(-0.0505051,0.49775)(0.0505051,0.473562)(0.151515,0.442738)(0.252525,0.406742)(0.353535,0.367195)(0.454545,0.325745)(0.555556,0.283963)(0.656566,0.243248)(0.757576,0.204759)(0.858586,0.169371)(0.959596,0.137669)(1.06061,0.109962)(1.16162,0.0863075)(1.26263,0.0665671)(1.36364,0.0504515)(1.46465,0.0375745)(1.56566,0.0274989)(1.66667,0.0197761)(1.76768,0.0139756)(1.86869,0.00970521)(1.9697,0.00662282)(2.07071,0.00444103)(2.17172,0.00292637)(2.27273,0.00189486)(2.37374,0.00120568)(2.47475,0.000753854)(2.57576,0.000463177)(2.67677,0.000279648)(2.77778,0.000165913)(2.87879,9.67277e-05)(2.9798,5.54149e-05)(3.08081,3.11965e-05)(3.18182,1.72579e-05)(3.28283,9.38157e-06)(3.38384,5.01148e-06)(3.48485,2.63063e-06)(3.58586,1.35693e-06)(3.68687,6.87797e-07)(3.78788,3.42583e-07)(3.88889,1.67678e-07)(3.9899,8.06471e-08)(4.09091,3.81159e-08)(4.19192,1.77022e-08)(4.29293,8.07888e-09)(4.39394,3.6231e-09)(4.49495,1.59666e-09)(4.59596,6.9143e-10)(4.69697,2.94231e-10)(4.79798,1.23036e-10)(4.89899,5.05567e-11)(5,2.04141e-11) 
};

\end{axis}
\end{tikzpicture}

%% file: mean_error.tikz
%
%
\begin{tikzpicture}

\begin{axis}[%
scale only axis,
width=0.8\columnwidth,
height=0.8\columnwidth,
xmin=0, xmax=100,
ymin=0, ymax=5,
xlabel={$$\# of samples$$},
ylabel={$||\vec{x}_\text{estimated}-\vec{x}_\text{true}||_2$},
axis on top,
legend entries={$$Laplace estimate$$,$$MCMC estimate$$,$$Laplace error est.$$,$$MCMC error est.$$},
legend style={nodes=right}]
\addplot [
color=red,
solid
]
coordinates{
 (1,2.90692)(2,2.81694)(3,2.77168)(4,2.74945)(5,2.68446)(6,2.62274)(7,2.59335)(8,2.55024)(9,2.50868)(10,2.45939)(11,2.43211)(12,2.43826)(13,2.40465)(14,2.38612)(15,2.33624)(16,2.31307)(17,2.30352)(18,2.30809)(19,2.28655)(20,2.2676)(21,2.26705)(22,2.2207)(23,2.20796)(24,2.17628)(25,2.16091)(26,2.15216)(27,2.12925)(28,2.09625)(29,2.07699)(30,2.052)(31,2.04013)(32,2.02116)(33,1.99915)(34,1.96941)(35,1.94989)(36,1.93967)(37,1.93347)(38,1.90781)(39,1.91746)(40,1.94084)(41,1.94525)(42,1.93873)(43,1.91963)(44,1.89719)(45,1.89314)(46,1.89615)(47,1.87642)(48,1.8631)(49,1.84218)(50,1.81846)(51,1.81086)(52,1.79282)(53,1.77311)(54,1.77416)(55,1.76459)(56,1.76994)(57,1.73484)(58,1.72391)(59,1.71864)(60,1.70804)(61,1.71028)(62,1.70737)(63,1.70893)(64,1.691)(65,1.6801)(66,1.68452)(67,1.67086)(68,1.66887)(69,1.67036)(70,1.67193)(71,1.66422)(72,1.67464)(73,1.66257)(74,1.65571)(75,1.64279)(76,1.63522)(77,1.62975)(78,1.60505)(79,1.58873)(80,1.58791)(81,1.58371)(82,1.57413)(83,1.56385)(84,1.55855)(85,1.54866)(86,1.54565)(87,1.53085)(88,1.52811)(89,1.51732)(90,1.51271)(91,1.50696)(92,1.5205)(93,1.52968)(94,1.54205)(95,1.53533)(96,1.52767)(97,1.53234)(98,1.5371)(99,1.53335)(100,1.53338) 
};

\addplot [
color=blue,
solid
]
coordinates{
 (1,3.72948)(2,3.45044)(3,3.31621)(4,3.25313)(5,3.12175)(6,3.03973)(7,3.00055)(8,2.95012)(9,2.82532)(10,2.84857)(11,2.75955)(12,2.75065)(13,2.71219)(14,2.71098)(15,2.61663)(16,2.57107)(17,2.58446)(18,2.5504)(19,2.49991)(20,2.44064)(21,2.42858)(22,2.42157)(23,2.48233)(24,2.39669)(25,2.30236)(26,2.30702)(27,2.25644)(28,2.25918)(29,2.27737)(30,2.21618)(31,2.19109)(32,2.21631)(33,2.19819)(34,2.14333)(35,2.12215)(36,2.06105)(37,2.09033)(38,1.99843)(39,2.00001)(40,1.96098)(41,2.068)(42,2.06384)(43,2.02874)(44,2.02721)(45,2.01214)(46,1.98832)(47,1.96584)(48,1.94288)(49,1.93352)(50,1.9561)(51,1.94825)(52,1.92268)(53,1.88689)(54,1.92418)(55,1.89177)(56,1.92058)(57,1.84177)(58,1.88173)(59,1.85481)(60,1.90509)(61,1.81876)(62,1.84039)(63,1.84327)(64,1.77846)(65,1.78472)(66,1.7885)(67,1.81509)(68,1.78373)(69,1.72684)(70,1.82018)(71,1.73912)(72,1.75532)(73,1.79106)(74,1.73299)(75,1.73422)(76,1.71429)(77,1.725)(78,1.70748)(79,1.70756)(80,1.68484)(81,1.73993)(82,1.67115)(83,1.68852)(84,1.6618)(85,1.63449)(86,1.69215)(87,1.67496)(88,1.65194)(89,1.63085)(90,1.65111)(91,1.67022)(92,1.66215)(93,1.68408)(94,1.63807)(95,1.65571)(96,1.70345)(97,1.64366)(98,1.61818)(99,1.62091)(100,1.67779) 
};

\addplot [
color=red,
dashed
]
coordinates{
 (1,3.64197)(2,3.57091)(3,3.5311)(4,3.52172)(5,3.43327)(6,3.36883)(7,3.34906)(8,3.31269)(9,3.27545)(10,3.22523)(11,3.19007)(12,3.17658)(13,3.11482)(14,3.0841)(15,3.00138)(16,2.98295)(17,2.96894)(18,2.97178)(19,2.95036)(20,2.93759)(21,2.91778)(22,2.88173)(23,2.86372)(24,2.83382)(25,2.81388)(26,2.80026)(27,2.77419)(28,2.74076)(29,2.71743)(30,2.69645)(31,2.67381)(32,2.6443)(33,2.59371)(34,2.57218)(35,2.55351)(36,2.54119)(37,2.53063)(38,2.50974)(39,2.49891)(40,2.50089)(41,2.51371)(42,2.51185)(43,2.48601)(44,2.46546)(45,2.47114)(46,2.45894)(47,2.43884)(48,2.42752)(49,2.40668)(50,2.38858)(51,2.37647)(52,2.3655)(53,2.34995)(54,2.35233)(55,2.343)(56,2.3503)(57,2.30922)(58,2.29894)(59,2.28728)(60,2.26411)(61,2.25755)(62,2.2451)(63,2.24193)(64,2.22282)(65,2.21346)(66,2.20915)(67,2.18883)(68,2.21394)(69,2.20531)(70,2.19819)(71,2.19025)(72,2.18832)(73,2.1787)(74,2.16541)(75,2.14859)(76,2.13868)(77,2.12253)(78,2.11546)(79,2.09154)(80,2.08423)(81,2.08122)(82,2.0751)(83,2.06118)(84,2.05499)(85,2.0468)(86,2.03602)(87,2.02654)(88,2.02277)(89,2.00757)(90,1.99713)(91,1.99384)(92,2.02014)(93,2.01882)(94,2.01326)(95,2.00709)(96,1.99623)(97,2.0044)(98,2.01123)(99,2.01149)(100,2.00574) 
};

\addplot [
color=blue,
dashed
]
coordinates{
 (1,4.67731)(2,4.44317)(3,4.31529)(4,4.21351)(5,4.01504)(6,3.96525)(7,3.88504)(8,3.82265)(9,3.7078)(10,3.76398)(11,3.71385)(12,3.58272)(13,3.62156)(14,3.64809)(15,3.55832)(16,3.53271)(17,3.48263)(18,3.5255)(19,3.38446)(20,3.35015)(21,3.34865)(22,3.33514)(23,3.39398)(24,3.28143)(25,3.21121)(26,3.19864)(27,3.13781)(28,3.20488)(29,3.22146)(30,3.12613)(31,3.09153)(32,3.12719)(33,3.08143)(34,3.04261)(35,3.01793)(36,2.95452)(37,2.94915)(38,2.81777)(39,2.78783)(40,2.74707)(41,2.8879)(42,2.91415)(43,2.88687)(44,2.98486)(45,2.8524)(46,2.8154)(47,2.78624)(48,2.71849)(49,2.77613)(50,2.81725)(51,2.78373)(52,2.75172)(53,2.70797)(54,2.79746)(55,2.74683)(56,2.72734)(57,2.63691)(58,2.74789)(59,2.70793)(60,2.79194)(61,2.60979)(62,2.68593)(63,2.60176)(64,2.52365)(65,2.56277)(66,2.54503)(67,2.56991)(68,2.61385)(69,2.55146)(70,2.62743)(71,2.50636)(72,2.55104)(73,2.6197)(74,2.47059)(75,2.50358)(76,2.54336)(77,2.46455)(78,2.4733)(79,2.5427)(80,2.4468)(81,2.54562)(82,2.42816)(83,2.47068)(84,2.49312)(85,2.39403)(86,2.48338)(87,2.41951)(88,2.39635)(89,2.41822)(90,2.42199)(91,2.42024)(92,2.48172)(93,2.40306)(94,2.38014)(95,2.40917)(96,2.48027)(97,2.35576)(98,2.34451)(99,2.33603)(100,2.41876) 
};

\end{axis}
\end{tikzpicture}

%% file: StateOfTheUnion_stacked.tikz
%
%
\begin{tikzpicture}

\definecolor{mycolor1}{rgb}{0.5,0.5,0}
\definecolor{mycolor2}{rgb}{0,1,1}
\definecolor{mycolor3}{rgb}{1,0,1}
\definecolor{mycolor4}{rgb}{1,1,0}

\definecolor{tcolor10}{rgb}{1.0,0.0,0.0}
\definecolor{tcolor3}{rgb}{0.75,0.0,0.0}
\definecolor{tcolor4}{rgb}{0.5,0.0,0.0}
\definecolor{tcolor1}{rgb}{0.0,0.5,0.0}
\definecolor{tcolor8}{rgb}{0.0,0.75,0.0}
\definecolor{tcolor2}{rgb}{0.0,1.0,0.0}
\definecolor{tcolor5}{rgb}{0.0,0.0,0.25}
\definecolor{tcolor6}{rgb}{0.0,0.0,0.5}
\definecolor{tcolor7}{rgb}{0.0,0.0,0.75}
\definecolor{tcolor9}{rgb}{0.0,0.0,1.0}

\pgfkeys{/pgf/number format/.cd, set thousands separator={}}

\begin{axis}[%
scale only axis,
stack plots=y,
width=360pt,
height=225pt,
xmin=1790, xmax=2012,
ymin=0, ymax=1.0,
xlabel={year},
ylabel={$\langle\pi_k\g\phi\rangle$},
axis on top]

\node[text=white,text centered,text width=3.0cm] at (axis cs:1830,0.8) 
{\tiny{war, constitution, union}};
\node[text=white,text centered,text width=2.5cm] at (axis cs:1830,0.2) 
{\tiny{public, commerce}};
\node[text=white,text centered,text width=2.5cm] at (axis cs:1890,0.1) 
{\tiny{made, business}};
\node[text=white,text centered,text width=1.5cm] at (axis cs:2000,0.9) 
{\tiny{energy, cut, oil}};

\node[text=white,text centered,text width=1.0cm] at (axis cs:1885,0.7) 
{\tiny{law,}};
\node[text=white,text centered,text width=1.0cm] at (axis cs:1905,0.5) 
{\tiny{good,}};
\node[text=white,text centered,text width=1.0cm] at (axis cs:1925,0.4) 
{\tiny{work}};

\node[text=white,text centered,text width=1cm] at (axis cs:1890,0.95) 
{\tiny{war, spain}};

\node[text=white,text centered,text width=2.5cm] at (axis cs:1930,0.9) 
{\tiny{war, national, economic}};

\node[text=white,text centered,text width=1cm] at (axis cs:1930,0.7) 
{\tiny{war,}};
\node[text=white,text centered,text width=1cm] at (axis cs:1950,0.8) 
{\tiny{men,}};
\node[text=white,text centered,text width=1cm] at (axis cs:1970,0.9) 
{\tiny{labor}};

\node[text=white,text centered,text width=1.5cm] at (axis cs:1950,0.3) 
{\tiny{world, peace, free}};

\node[text=white,text centered,text width=1.5cm] at (axis cs:1995,0.5) 
{\tiny{America, American, people}};

\addplot [
color=tcolor10,
solid,
fill=tcolor10,
]
coordinates{
 (1790,1.29118e-005)(1791,9.00606e-006)(1792,7.36706e-006)(1793,6.66477e-006)(1794,6.40283e-006)(1795,6.69335e-006)(1796,7.99498e-006)(1797,7.44053e-006)(1798,5.8115e-006)(1799,5.68577e-006)(1800,6.92845e-006)(1801,6.23661e-006)(1802,4.90626e-006)(1803,4.48809e-006)(1804,4.53989e-006)(1805,4.82809e-006)(1806,5.36664e-006)(1807,6.48934e-006)(1808,8.91037e-006)(1809,1.46048e-005)(1810,1.27985e-005)(1811,1.29393e-005)(1812,1.44046e-005)(1813,1.68711e-005)(1814,2.07018e-005)(1815,2.74405e-005)(1816,4.03194e-005)(1817,0.00040989)(1818,0.000606241)(1819,0.000921313)(1820,0.00140537)(1821,0.00211263)(1822,0.00309527)(1823,0.00440135)(1824,0.00607872)(1825,0.0164985)(1826,0.0227655)(1827,0.0298255)(1828,0.0371249)(1829,0.0330644)(1830,0.041232)(1831,0.0495035)(1832,0.0574208)(1833,0.0646887)(1834,0.0711738)(1835,0.0768474)(1836,0.0817325)(1837,0.0906224)(1838,0.0989018)(1839,0.105456)(1840,0.110127)(1841,0.0851822)(1842,0.089285)(1843,0.0929642)(1844,0.0963981)(1845,0.0648223)(1846,0.0667361)(1847,0.0697419)(1848,0.0739715)(1849,0.109643)(1850,0.145635)(1851,0.152548)(1852,0.158405)(1853,0.144503)(1854,0.148593)(1855,0.152428)(1856,0.156623)(1857,0.109073)(1858,0.111884)(1859,0.116506)(1860,0.123547)(1861,0.255827)(1862,0.265356)(1863,0.272998)(1864,0.279806)(1865,0.219828)(1866,0.240265)(1867,0.265057)(1868,0.293502)(1869,0.305099)(1870,0.313991)(1871,0.321517)(1872,0.328911)(1873,0.337369)(1874,0.347832)(1875,0.360851)(1876,0.376294)(1877,0.455451)(1878,0.46884)(1879,0.479919)(1880,0.488666)(1881,0.478172)(1882,0.474979)(1883,0.471469)(1884,0.469678)(1885,0.487031)(1886,0.501533)(1887,0.521693)(1888,0.539287)(1889,0.529398)(1890,0.539006)(1891,0.549484)(1892,0.558914)(1893,0.534958)(1894,0.527394)(1895,0.519935)(1896,0.516256)(1897,0.310323)(1898,0.308445)(1899,0.317939)(1900,0.337379)(1901,0.205632)(1902,0.214985)(1903,0.225552)(1904,0.228521)(1905,0.221102)(1906,0.206898)(1907,0.191953)(1908,0.180657)(1909,0.385392)(1910,0.363403)(1911,0.343562)(1912,0.324879)(1913,0.105304)(1914,0.0948163)(1915,0.0959939)(1916,0.0883075)(1917,0.0801194)(1918,0.0857664)(1919,0.11054)(1920,0.157249)(1921,0.234468)(1922,0.229034)(1923,0.225905)(1924,0.234723)(1925,0.244909)(1926,0.254904)(1927,0.263061)(1928,0.267632)(1929,0.285313)(1930,0.267953)(1931,0.241728)(1932,0.210553)(1933,0.0895317)(1934,0.0644452)(1935,0.0411309)(1936,0.0229911)(1937,0.0113071)(1938,0.00495456)(1939,0.0019783)(1940,0.000757374)(1941,0.000302616)(1942,0.000135542)(1943,7.08931e-005)(1944,4.93211e-005)(1945,5.37008e-005)(1946,0.0991777)(1947,0.0787155)(1948,0.0580572)(1949,0.0413152)(1950,0.0298651)(1951,0.0229886)(1952,0.0193816)(1953,0.0847981)(1954,0.0842374)(1955,0.0796758)(1956,0.0733114)(1957,0.0671564)(1958,0.0621869)(1959,0.0582484)(1960,0.0543748)(1961,0.0495115)(1962,0.0525501)(1963,0.0417925)(1964,0.0317163)(1965,0.0337548)(1966,0.0360725)(1967,0.0371937)(1968,0.0353895)(1969,0.0300613)(1970,0.00764836)(1971,0.00587426)(1972,0.00428892)(1973,0.00299049)(1974,0.00207555)(1975,1.91518e-005)(1976,1.21605e-005)(1977,1.55699e-005)(1978,0.00477904)(1979,0.00644816)(1980,0.0085778)(1981,0.0102958)(1982,1.06104e-005)(1983,6.80414e-006)(1984,5.8342e-006)(1985,5.75254e-006)(1986,5.90428e-006)(1987,6.91518e-006)(1988,1.05429e-005)(1989,1.29267e-005)(1990,8.17044e-006)(1991,7.77047e-006)(1992,1.1052e-005)(1993,0.00028976)(1994,0.00011023)(1995,4.0857e-005)(1996,1.84054e-005)(1997,9.86532e-006)(1998,6.24225e-006)(1999,5.12022e-006)(2000,5.94587e-006)(2001,2.87626e-006)(2002,1.75754e-006)(2003,1.38849e-006)(2004,1.28953e-006)(2005,1.28441e-006)(2006,1.37478e-006)(2007,1.73621e-006)(2008,2.84426e-006)(2009,4.70924e-006)(2010,3.87157e-006)(2011,5.72808e-006) (2012,0)
};

\addplot [
color=tcolor3,
solid,
fill=tcolor3,
]
coordinates{
(1790,0)(1791,0.677909)(1792,0.671179)(1793,0.661592)(1794,0.650846)(1795,0.640563)(1796,0.631721)(1797,0.61611)(1798,0.614315)(1799,0.612268)(1800,0.609963)(1801,0.629502)(1802,0.626741)(1803,0.623205)(1804,0.619173)(1805,0.614901)(1806,0.610507)(1807,0.605957)(1808,0.601128)(1809,0.598676)(1810,0.592423)(1811,0.585131)(1812,0.577568)(1813,0.5705)(1814,0.564361)(1815,0.559178)(1816,0.554728)(1817,0.53068)(1818,0.528671)(1819,0.527366)(1820,0.526411)(1821,0.525476)(1822,0.524294)(1823,0.522612)(1824,0.520153)(1825,0.524915)(1826,0.51763)(1827,0.508998)(1828,0.499344)(1829,0.50126)(1830,0.491539)(1831,0.480259)(1832,0.467585)(1833,0.453865)(1834,0.439623)(1835,0.425467)(1836,0.411935)(1837,0.416769)(1838,0.405478)(1839,0.394343)(1840,0.383481)(1841,0.373278)(1842,0.363836)(1843,0.354818)(1844,0.346271)(1845,0.245832)(1846,0.238304)(1847,0.23409)(1848,0.232823)(1849,0.310015)(1850,0.323823)(1851,0.31797)(1852,0.311366)(1853,0.293757)(1854,0.285513)(1855,0.277037)(1856,0.269007)(1857,0.201398)(1858,0.19607)(1859,0.193067)(1860,0.192468)(1861,0.217241)(1862,0.209031)(1863,0.201726)(1864,0.195794)(1865,0.22534)(1866,0.217788)(1867,0.210815)(1868,0.205473)(1869,0.173414)(1870,0.163641)(1871,0.154289)(1872,0.146429)(1873,0.140942)(1874,0.138238)(1875,0.138172)(1876,0.140109)(1877,0.158244)(1878,0.15149)(1879,0.145353)(1880,0.139945)(1881,0.125189)(1882,0.117274)(1883,0.109017)(1884,0.100132)(1885,0.166413)(1886,0.126164)(1887,0.0766653)(1888,0.0332501)(1889,0.0486327)(1890,0.0350795)(1891,0.025265)(1892,0.0187685)(1893,4.8906e-005)(1894,2.19513e-005)(1895,1.62751e-005)(1896,2.5672e-005)(1897,0.0924878)(1898,0.084542)(1899,0.0771945)(1900,0.0723598)(1901,0.0960479)(1902,0.0990243)(1903,0.0988914)(1904,0.0957594)(1905,0.0909162)(1906,0.0859193)(1907,0.0822894)(1908,0.0811332)(1909,0.0403093)(1910,0.0415922)(1911,0.0441349)(1912,0.0480293)(1913,0.165148)(1914,0.173342)(1915,0.171142)(1916,0.189741)(1917,0.197012)(1918,0.168935)(1919,0.13542)(1920,0.111573)(1921,0.0556882)(1922,0.0472961)(1923,0.0637815)(1924,0.0643435)(1925,0.066578)(1926,0.0699982)(1927,0.0741259)(1928,0.0784648)(1929,0.0557691)(1930,0.0528526)(1931,0.0504396)(1932,0.048889)(1933,0.0990155)(1934,0.100925)(1935,0.0987349)(1936,0.0920461)(1937,0.0823903)(1938,0.0721545)(1939,0.0632257)(1940,0.0565)(1941,0.0520554)(1942,0.0495024)(1943,0.048209)(1944,0.0474278)(1945,0.0464554)(1946,0.00450815)(1947,0.00520801)(1948,0.00639529)(1949,0.00797132)(1950,0.00974656)(1951,0.0114938)(1952,0.0129876)(1953,0.0195759)(1954,0.0187655)(1955,0.0186718)(1956,0.0192063)(1957,0.019878)(1958,0.0198344)(1959,0.0183014)(1960,0.0152738)(1961,0.0116584)(1962,0.0201813)(1963,0.0146579)(1964,2.0868e-005)(1965,1.05644e-005)(1966,7.73268e-006)(1967,7.12274e-006)(1968,8.14537e-006)(1969,1.28656e-005)(1970,0.00412317)(1971,0.00350587)(1972,0.00287188)(1973,0.00229213)(1974,0.00185628)(1975,0.00229892)(1976,0.00256777)(1977,0.00287367)(1978,2.0862e-005)(1979,1.46442e-005)(1980,1.74548e-005)(1981,3.20283e-005)(1982,0.00151478)(1983,0.00182667)(1984,0.00225005)(1985,0.00276863)(1986,0.00335656)(1987,0.0039592)(1988,0.00446417)(1989,0.00206255)(1990,0.00244572)(1991,0.00283884)(1992,0.00309365)(1993,0.000798315)(1994,0.000748922)(1995,0.000722563)(1996,0.000690627)(1997,0.000643875)(1998,0.000587414)(1999,0.00053255)(2000,0.000491989)(2001,9.55488e-006)(2002,6.10543e-006)(2003,4.96088e-006)(2004,4.70372e-006)(2005,4.79099e-006)(2006,5.24783e-006)(2007,6.71049e-006)(2008,1.08809e-005)(2009,0.00284287)(2010,0.00310847)(2011,0.00332138) (2012,0)
};

\addplot [
color=tcolor4,
solid,
fill=tcolor4,
]
coordinates{
 (1790,8.84854e-006)(1791,6.09282e-006)(1792,4.92129e-006)(1793,4.39443e-006)(1794,4.16388e-006)(1795,4.29121e-006)(1796,5.05399e-006)(1797,4.47173e-006)(1798,3.40807e-006)(1799,3.26185e-006)(1800,3.89822e-006)(1801,3.39701e-006)(1802,2.53796e-006)(1803,2.18701e-006)(1804,2.06647e-006)(1805,2.03804e-006)(1806,2.09387e-006)(1807,2.34798e-006)(1808,3.01815e-006)(1809,5.38265e-006)(1810,4.18025e-006)(1811,3.64995e-006)(1812,3.3957e-006)(1813,3.20225e-006)(1814,3.06959e-006)(1815,3.16963e-006)(1816,3.75723e-006)(1817,2.61191e-006)(1818,2.00367e-006)(1819,1.76499e-006)(1820,1.68607e-006)(1821,1.65453e-006)(1822,1.66488e-006)(1823,1.80831e-006)(1824,2.24381e-006)(1825,2.5983e-006)(1826,2.10219e-006)(1827,2.11005e-006)(1828,2.62465e-006)(1829,1.9498e-006)(1830,1.55615e-006)(1831,1.43945e-006)(1832,1.45785e-006)(1833,1.53087e-006)(1834,1.66445e-006)(1835,1.97675e-006)(1836,2.71902e-006)(1837,3.50424e-006)(1838,3.23957e-006)(1839,3.80235e-006)(1840,5.6178e-006)(1841,6.64525e-006)(1842,7.08272e-006)(1843,9.64918e-006)(1844,1.61926e-005)(1845,0.000206564)(1846,0.000325322)(1847,0.000522799)(1848,0.000841768)(1849,0.0105187)(1850,0.00880818)(1851,0.013376)(1852,0.0188189)(1853,0.00929312)(1854,0.0128057)(1855,0.0169998)(1856,0.02175)(1857,0.0187653)(1858,0.0229337)(1859,0.0276199)(1860,0.0328547)(1861,0.0595515)(1862,0.067357)(1863,0.0744206)(1864,0.0807245)(1865,0.0579498)(1866,0.0629777)(1867,0.0683768)(1868,0.0745288)(1869,0.1293)(1870,0.143004)(1871,0.156404)(1872,0.169318)(1873,0.181728)(1874,0.193599)(1875,0.204701)(1876,0.214597)(1877,0.159544)(1878,0.17251)(1879,0.186321)(1880,0.199844)(1881,0.179237)(1882,0.184063)(1883,0.188705)(1884,0.193291)(1885,0.236882)(1886,0.245891)(1887,0.257967)(1888,0.269765)(1889,0.240658)(1890,0.23636)(1891,0.229777)(1892,0.222127)(1893,0.274939)(1894,0.261545)(1895,0.243904)(1896,0.22624)(1897,0.136641)(1898,0.150203)(1899,0.170527)(1900,0.193386)(1901,0.373275)(1902,0.381218)(1903,0.39323)(1904,0.403176)(1905,0.406765)(1906,0.40037)(1907,0.382189)(1908,0.353554)(1909,0.402909)(1910,0.405564)(1911,0.401108)(1912,0.389255)(1913,0.00157804)(1914,0.00440337)(1915,0.0183277)(1916,0.0346563)(1917,0.0612189)(1918,0.126541)(1919,0.181141)(1920,0.190039)(1921,0.152911)(1922,0.16143)(1923,0.21945)(1924,0.210679)(1925,0.200319)(1926,0.18853)(1927,0.174109)(1928,0.155617)(1929,0.148835)(1930,0.110046)(1931,0.0761815)(1932,0.0503613)(1933,0.0997269)(1934,0.059854)(1935,0.0294152)(1936,0.0118249)(1937,0.00405727)(1938,0.00124498)(1939,0.000364221)(1940,0.000115972)(1941,4.47777e-005)(1942,2.09786e-005)(1943,1.18214e-005)(1944,9.16762e-006)(1945,1.15529e-005)(1946,9.83347e-006)(1947,5.87132e-006)(1948,5.01977e-006)(1949,4.8633e-006)(1950,4.53685e-006)(1951,4.22075e-006)(1952,4.80467e-006)(1953,8.82827e-006)(1954,4.9383e-006)(1955,4.09644e-006)(1956,4.16194e-006)(1957,4.41583e-006)(1958,4.73695e-006)(1959,5.18722e-006)(1960,6.74534e-006)(1961,1.28626e-005)(1962,0.00874867)(1963,0.00550178)(1964,1.17836e-005)(1965,5.86958e-006)(1966,4.24266e-006)(1967,3.87017e-006)(1968,4.38163e-006)(1969,6.8276e-006)(1970,1.17062e-005)(1971,5.98099e-006)(1972,4.41188e-006)(1973,4.53062e-006)(1974,6.48108e-006)(1975,6.62142e-006)(1976,4.22028e-006)(1977,5.55218e-006)(1978,3.52e-006)(1979,1.83703e-006)(1980,1.64342e-006)(1981,2.65384e-006)(1982,3.84904e-006)(1983,2.47037e-006)(1984,2.15023e-006)(1985,2.17451e-006)(1986,2.29963e-006)(1987,2.77574e-006)(1988,4.35315e-006)(1989,5.2036e-006)(1990,3.30443e-006)(1991,3.19155e-006)(1992,4.65505e-006)(1993,3.6159e-006)(1994,1.96735e-006)(1995,1.49342e-006)(1996,1.32909e-006)(1997,1.20514e-006)(1998,1.12905e-006)(1999,1.24446e-006)(2000,1.80983e-006)(2001,1.55509e-006)(2002,9.73271e-007)(2003,7.87887e-007)(2004,7.50402e-007)(2005,7.67111e-007)(2006,8.43113e-007)(2007,1.09322e-006)(2008,1.83742e-006)(2009,3.09521e-006)(2010,2.57864e-006)(2011,3.87284e-006) (2012,0)
};

\addplot [
color=tcolor1,
solid,
fill=tcolor1,
]
coordinates{
 (1790,7.50563e-006)(1791,5.1439e-006)(1792,4.13909e-006)(1793,3.68514e-006)(1794,3.48346e-006)(1795,3.58198e-006)(1796,4.20877e-006)(1797,3.70728e-006)(1798,2.81117e-006)(1799,2.67763e-006)(1800,3.18546e-006)(1801,2.7582e-006)(1802,2.04731e-006)(1803,1.75255e-006)(1804,1.64499e-006)(1805,1.61179e-006)(1806,1.6455e-006)(1807,1.83394e-006)(1808,2.34323e-006)(1809,4.15788e-006)(1810,3.19713e-006)(1811,2.76322e-006)(1812,2.54474e-006)(1813,2.37613e-006)(1814,2.2559e-006)(1815,2.30727e-006)(1816,2.70807e-006)(1817,1.86563e-006)(1818,1.40938e-006)(1819,1.22123e-006)(1820,1.14634e-006)(1821,1.1042e-006)(1822,1.08961e-006)(1823,1.15947e-006)(1824,1.40797e-006)(1825,1.56817e-006)(1826,1.23012e-006)(1827,1.19593e-006)(1828,1.44007e-006)(1829,9.93819e-007)(1830,7.47905e-007)(1831,6.46883e-007)(1832,6.07366e-007)(1833,5.86985e-007)(1834,5.85262e-007)(1835,6.38868e-007)(1836,8.13854e-007)(1837,8.84641e-007)(1838,6.93627e-007)(1839,6.92204e-007)(1840,8.86471e-007)(1841,9.32412e-007)(1842,7.41286e-007)(1843,7.52783e-007)(1844,9.85723e-007)(1845,7.43446e-007)(1846,5.79776e-007)(1847,5.90483e-007)(1848,7.9494e-007)(1849,3.7636e-006)(1850,1.32555e-006)(1851,1.14338e-006)(1852,1.43223e-006)(1853,1.05395e-006)(1854,8.26241e-007)(1855,8.6688e-007)(1856,1.23819e-006)(1857,8.14066e-007)(1858,6.32881e-007)(1859,6.81199e-007)(1860,1.03274e-006)(1861,3.08828e-006)(1862,2.33673e-006)(1863,2.50146e-006)(1864,3.85932e-006)(1865,1.97115e-006)(1866,1.40507e-006)(1867,1.43628e-006)(1868,2.156e-006)(1869,2.02403e-006)(1870,1.32514e-006)(1871,1.07706e-006)(1872,9.95203e-007)(1873,9.57822e-007)(1874,9.72843e-007)(1875,1.17725e-006)(1876,1.92003e-006)(1877,2.30369e-006)(1878,1.57354e-006)(1879,1.66598e-006)(1880,2.83173e-006)(1881,2.62547e-006)(1882,1.92893e-006)(1883,2.28218e-006)(1884,4.49685e-006)(1885,2.37298e-006)(1886,1.72754e-006)(1887,2.12908e-006)(1888,3.85129e-006)(1889,5.02839e-006)(1890,3.49224e-006)(1891,4.49084e-006)(1892,1.12646e-005)(1893,7.66581e-006)(1894,5.9942e-006)(1895,7.31914e-006)(1896,1.71713e-005)(1897,0.00828208)(1898,0.0111221)(1899,0.0136395)(1900,0.0157907)(1901,0.0826449)(1902,0.0819602)(1903,0.0738931)(1904,0.0646114)(1905,0.0572669)(1906,0.0525624)(1907,0.0506381)(1908,0.0517912)(1909,0.000962343)(1910,0.00194238)(1911,0.00543798)(1912,0.0159203)(1913,0.14238)(1914,0.150136)(1915,0.152575)(1916,0.17582)(1917,0.205134)(1918,0.203968)(1919,0.180468)(1920,0.152504)(1921,0.119297)(1922,0.11314)(1923,0.0836072)(1924,0.0798095)(1925,0.0764878)(1926,0.0743216)(1927,0.07431)(1928,0.0774115)(1929,0.0945003)(1930,0.106904)(1931,0.122031)(1932,0.139417)(1933,0.223755)(1934,0.255712)(1935,0.285779)(1936,0.310618)(1937,0.330464)(1938,0.347968)(1939,0.36571)(1940,0.384839)(1941,0.404934)(1942,0.424577)(1943,0.44208)(1944,0.456173)(1945,0.466513)(1946,0.359245)(1947,0.395034)(1948,0.432495)(1949,0.464493)(1950,0.486056)(1951,0.496538)(1952,0.498953)(1953,0.39179)(1954,0.396222)(1955,0.404328)(1956,0.415386)(1957,0.426568)(1958,0.434286)(1959,0.436613)(1960,0.433928)(1961,0.427787)(1962,0.382031)(1963,0.405313)(1964,0.340373)(1965,0.322432)(1966,0.305995)(1967,0.291828)(1968,0.280379)(1969,0.271659)(1970,0.292165)(1971,0.277055)(1972,0.266836)(1973,0.261011)(1974,0.258092)(1975,0.281953)(1976,0.276967)(1977,0.27414)(1978,0.246298)(1979,0.240988)(1980,0.234171)(1981,0.228204)(1982,0.238393)(1983,0.225171)(1984,0.209786)(1985,0.193656)(1986,0.178498)(1987,0.165566)(1988,0.15529)(1989,0.179626)(1990,0.175365)(1991,0.170726)(1992,0.165287)(1993,0.0844667)(1994,0.0782641)(1995,0.0744584)(1996,0.0729693)(1997,0.0734998)(1998,0.0755748)(1999,0.0786381)(2000,0.0821722)(2001,0.132086)(2002,0.132946)(2003,0.133791)(2004,0.134253)(2005,0.134109)(2006,0.133328)(2007,0.132036)(2008,0.130455)(2009,0.0678128)(2010,0.0645736)(2011,0.0629721)(2012,0)
};

\addplot [
color=tcolor8,
solid,
fill=tcolor8,
]
coordinates{
 (1790,7.79201e-006)(1791,5.34883e-006)(1792,4.3128e-006)(1793,3.8486e-006)(1794,3.64587e-006)(1795,3.75516e-006)(1796,4.41637e-006)(1797,3.81093e-006)(1798,2.89078e-006)(1799,2.75585e-006)(1800,3.28314e-006)(1801,2.85584e-006)(1802,2.12108e-006)(1803,1.81775e-006)(1804,1.70913e-006)(1805,1.67849e-006)(1806,1.71836e-006)(1807,1.92106e-006)(1808,2.46252e-006)(1809,4.41871e-006)(1810,3.40699e-006)(1811,2.95217e-006)(1812,2.72508e-006)(1813,2.54994e-006)(1814,2.42597e-006)(1815,2.48676e-006)(1816,2.92604e-006)(1817,2.0224e-006)(1818,1.53305e-006)(1819,1.3331e-006)(1820,1.25591e-006)(1821,1.21428e-006)(1822,1.20288e-006)(1823,1.28517e-006)(1824,1.56728e-006)(1825,1.75171e-006)(1826,1.38097e-006)(1827,1.34969e-006)(1828,1.63429e-006)(1829,1.13939e-006)(1830,8.63398e-007)(1831,7.52241e-007)(1832,7.11682e-007)(1833,6.93225e-007)(1834,6.96797e-007)(1835,7.6701e-007)(1836,9.85737e-007)(1837,1.07483e-006)(1838,8.52104e-007)(1839,8.60652e-007)(1840,1.11673e-006)(1841,1.19349e-006)(1842,9.63394e-007)(1843,9.94598e-007)(1844,1.32593e-006)(1845,1.03067e-006)(1846,8.22265e-007)(1847,8.57705e-007)(1848,1.18388e-006)(1849,5.78235e-006)(1850,2.14149e-006)(1851,1.91282e-006)(1852,2.47972e-006)(1853,1.88676e-006)(1854,1.54227e-006)(1855,1.68317e-006)(1856,2.4896e-006)(1857,1.72185e-006)(1858,1.38773e-006)(1859,1.53175e-006)(1860,2.35114e-006)(1861,0.00373144)(1862,0.00440482)(1863,0.00405548)(1864,0.00301049)(1865,4.2189e-006)(1866,2.93466e-006)(1867,2.90204e-006)(1868,4.18657e-006)(1869,3.86015e-006)(1870,2.44585e-006)(1871,1.91867e-006)(1872,1.70947e-006)(1873,1.5879e-006)(1874,1.56006e-006)(1875,1.8316e-006)(1876,2.90828e-006)(1877,3.5005e-006)(1878,2.37419e-006)(1879,2.49362e-006)(1880,4.19699e-006)(1881,3.71685e-006)(1882,2.72967e-006)(1883,3.22479e-006)(1884,6.3275e-006)(1885,3.04372e-006)(1886,2.16979e-006)(1887,2.59764e-006)(1888,4.52084e-006)(1889,0.00137172)(1890,0.00225384)(1891,0.00341149)(1892,0.00455204)(1893,6.05934e-006)(1894,4.19658e-006)(1895,4.58829e-006)(1896,1.00366e-005)(1897,0.00285763)(1898,0.000875378)(1899,0.000255573)(1900,0.000189939)(1901,0.0179039)(1902,0.0180814)(1903,0.0182627)(1904,0.0184691)(1905,0.0183498)(1906,0.0174982)(1907,0.0159264)(1908,0.0140656)(1909,7.60583e-006)(1910,3.36849e-006)(1911,3.30973e-006)(1912,7.35767e-006)(1913,0.0660864)(1914,0.0671673)(1915,0.0748066)(1916,0.0627803)(1917,0.0467874)(1918,0.0410343)(1919,0.0395394)(1920,0.0373154)(1921,0.00694677)(1922,0.0074125)(1923,0.00739118)(1924,0.00447525)(1925,0.00220056)(1926,0.000915822)(1927,0.000392557)(1928,0.000257576)(1929,2.37433e-005)(1930,1.20733e-005)(1931,1.14296e-005)(1932,1.94354e-005)(1933,0.0364906)(1934,0.0546508)(1935,0.0769581)(1936,0.0997528)(1937,0.119941)(1938,0.136697)(1939,0.150073)(1940,0.159295)(1941,0.162818)(1942,0.159866)(1943,0.151693)(1944,0.141329)(1945,0.132157)(1946,0.0212833)(1947,0.0330762)(1948,0.0526758)(1949,0.0815997)(1950,0.118113)(1951,0.15696)(1952,0.192247)(1953,0.0811385)(1954,0.0871846)(1955,0.09401)(1956,0.102089)(1957,0.111154)(1958,0.120341)(1959,0.129063)(1960,0.137749)(1961,0.14773)(1962,0.302019)(1963,0.341866)(1964,0.421684)(1965,0.436106)(1966,0.443147)(1967,0.444668)(1968,0.443709)(1969,0.44289)(1970,0.477762)(1971,0.482452)(1972,0.482936)(1973,0.48058)(1974,0.476926)(1975,0.396438)(1976,0.397784)(1977,0.400804)(1978,0.354113)(1979,0.304925)(1980,0.260375)(1981,0.229834)(1982,0.521697)(1983,0.552587)(1984,0.586703)(1985,0.619628)(1986,0.64747)(1987,0.668476)(1988,0.6832)(1989,0.635603)(1990,0.656978)(1991,0.676398)(1992,0.693439)(1993,0.757145)(1994,0.769864)(1995,0.779328)(1996,0.785396)(1997,0.788059)(1998,0.787648)(1999,0.784817)(2000,0.780304)(2001,0.714849)(2002,0.713366)(2003,0.711029)(2004,0.708229)(2005,0.705348)(2006,0.702646)(2007,0.70018)(2008,0.69782)(2009,0.744131)(2010,0.746901)(2011,0.748636) (2012,0)
};

\addplot [
color=tcolor2,
solid,
fill=tcolor2,
]
coordinates{
 (1790,6.87782e-006)(1791,4.69874e-006)(1792,3.77183e-006)(1793,3.35374e-006)(1794,3.16989e-006)(1795,3.2628e-006)(1796,3.84066e-006)(1797,3.405e-006)(1798,2.58114e-006)(1799,2.45922e-006)(1800,2.92817e-006)(1801,2.54097e-006)(1802,1.88546e-006)(1803,1.61422e-006)(1804,1.51617e-006)(1805,1.48738e-006)(1806,1.52103e-006)(1807,1.69857e-006)(1808,2.1749e-006)(1809,3.88304e-006)(1810,2.98668e-006)(1811,2.58177e-006)(1812,2.37789e-006)(1813,2.22076e-006)(1814,2.10934e-006)(1815,2.15914e-006)(1816,2.53728e-006)(1817,1.75049e-006)(1818,1.32357e-006)(1819,1.14819e-006)(1820,1.07925e-006)(1821,1.04116e-006)(1822,1.02908e-006)(1823,1.09687e-006)(1824,1.33415e-006)(1825,1.49244e-006)(1826,1.17175e-006)(1827,1.14017e-006)(1828,1.37405e-006)(1829,9.50457e-007)(1830,7.15839e-007)(1831,6.1959e-007)(1832,5.82087e-007)(1833,5.62808e-007)(1834,5.61324e-007)(1835,6.12843e-007)(1836,7.80769e-007)(1837,8.45583e-007)(1838,6.63141e-007)(1839,6.61996e-007)(1840,8.48146e-007)(1841,8.87231e-007)(1842,7.05424e-007)(1843,7.16646e-007)(1844,9.39059e-007)(1845,7.12608e-007)(1846,5.55911e-007)(1847,5.66457e-007)(1848,7.63096e-007)(1849,3.61361e-006)(1850,1.27344e-006)(1851,1.09932e-006)(1852,1.37844e-006)(1853,1.01377e-006)(1854,7.95694e-007)(1855,8.36081e-007)(1856,1.19631e-006)(1857,7.87186e-007)(1858,6.13167e-007)(1859,6.61507e-007)(1860,1.00555e-006)(1861,2.95997e-006)(1862,2.25134e-006)(1863,2.42483e-006)(1864,3.76496e-006)(1865,1.94735e-006)(1866,1.39286e-006)(1867,1.42913e-006)(1868,2.15401e-006)(1869,2.03497e-006)(1870,1.33762e-006)(1871,1.09177e-006)(1872,1.01331e-006)(1873,9.7998e-007)(1874,1.00061e-006)(1875,1.21787e-006)(1876,1.99889e-006)(1877,2.40781e-006)(1878,1.65478e-006)(1879,1.76376e-006)(1880,3.01995e-006)(1881,2.84022e-006)(1882,2.10032e-006)(1883,2.49982e-006)(1884,4.95058e-006)(1885,5.53245e-006)(1886,4.92045e-006)(1887,7.61991e-006)(1888,1.78315e-005)(1889,5.01411e-006)(1890,3.36338e-006)(1891,4.17016e-006)(1892,1.01378e-005)(1893,0.00213992)(1894,0.00451756)(1895,0.00734918)(1896,0.0100298)(1897,0.000412156)(1898,0.000616456)(1899,0.00169116)(1900,0.00411391)(1901,9.53883e-006)(1902,4.84117e-006)(1903,3.785e-006)(1904,3.50622e-006)(1905,3.32564e-006)(1906,3.38459e-006)(1907,4.57966e-006)(1908,1.09425e-005)(1909,0.0207872)(1910,0.0199278)(1911,0.0170502)(1912,0.0135236)(1913,0.0222913)(1914,0.0277081)(1915,0.0240132)(1916,0.00722882)(1917,0.00132322)(1918,0.000264325)(1919,9.52067e-005)(1920,9.08844e-005)(1921,0.00825589)(1922,0.00729526)(1923,0.0135665)(1924,0.0133898)(1925,0.0130037)(1926,0.0124273)(1927,0.0116184)(1928,0.0104531)(1929,0.00399331)(1930,0.00174278)(1931,0.000694521)(1932,0.000405242)(1933,7.66035e-005)(1934,3.22873e-005)(1935,1.61319e-005)(1936,1.15452e-005)(1937,9.65854e-006)(1938,8.13671e-006)(1939,6.90667e-006)(1940,6.05328e-006)(1941,5.49537e-006)(1942,5.08901e-006)(1943,4.87928e-006)(1944,5.66128e-006)(1945,9.53051e-006)(1946,0.0118514)(1947,0.00546247)(1948,0.0018551)(1949,0.000521359)(1950,0.000149132)(1951,5.55297e-005)(1952,3.32418e-005)(1953,0.0146195)(1954,0.0202726)(1955,0.0260972)(1956,0.0302843)(1957,0.0320508)(1958,0.0325168)(1959,0.0335919)(1960,0.0365532)(1961,0.0415532)(1962,1.64824e-005)(1963,3.76713e-005)(1964,0.026069)(1965,0.0352507)(1966,0.045105)(1967,0.0547738)(1968,0.0639948)(1969,0.0730241)(1970,0.0672139)(1971,0.0807127)(1972,0.0921294)(1973,0.101528)(1974,0.109659)(1975,0.122915)(1976,0.13387)(1977,0.142054)(1978,0.157395)(1979,0.185861)(1980,0.215804)(1981,0.24543)(1982,0.109352)(1983,0.107884)(1984,0.104986)(1985,0.101487)(1986,0.0983078)(1987,0.0961)(1988,0.0950794)(1989,0.108597)(1990,0.103)(1991,0.0984074)(1992,0.0951999)(1993,0.134017)(1994,0.128827)(1995,0.124369)(1996,0.121057)(1997,0.11911)(1998,0.11846)(1999,0.118808)(2000,0.119804)(2001,0.113223)(2002,0.112966)(2003,0.113517)(2004,0.115011)(2005,0.11737)(2006,0.120326)(2007,0.123528)(2008,0.126652)(2009,0.156087)(2010,0.154987)(2011,0.15285) (2012,0)
};

\addplot [
color=tcolor5,
solid,
fill=tcolor5,
]
coordinates{
 (1790,0.318039)(1791,0.320733)(1792,0.327338)(1793,0.336872)(1794,0.347707)(1795,0.35822)(1796,0.367347)(1797,0.383833)(1798,0.385641)(1799,0.38769)(1800,0.389988)(1801,0.370441)(1802,0.373211)(1803,0.376749)(1804,0.380779)(1805,0.385049)(1806,0.389439)(1807,0.393982)(1808,0.398798)(1809,0.398222)(1810,0.403517)(1811,0.41011)(1812,0.417583)(1813,0.42526)(1814,0.432436)(1815,0.438673)(1816,0.443927)(1817,0.468888)(1818,0.470707)(1819,0.471699)(1820,0.472171)(1821,0.4724)(1822,0.4726)(1823,0.472975)(1824,0.473755)(1825,0.458572)(1826,0.459592)(1827,0.461165)(1828,0.463517)(1829,0.465666)(1830,0.467222)(1831,0.470231)(1832,0.474988)(1833,0.48144)(1834,0.489197)(1835,0.497678)(1836,0.506324)(1837,0.492598)(1838,0.495611)(1839,0.500192)(1840,0.506379)(1841,0.541526)(1842,0.546866)(1843,0.552202)(1844,0.557306)(1845,0.689133)(1846,0.694629)(1847,0.69564)(1848,0.692356)(1849,0.568383)(1850,0.521722)(1851,0.516096)(1852,0.511396)(1853,0.552437)(1854,0.55308)(1855,0.553527)(1856,0.552606)(1857,0.670755)(1858,0.669105)(1859,0.662798)(1860,0.651117)(1861,0.457311)(1862,0.441575)(1863,0.425092)(1864,0.407175)(1865,0.492294)(1866,0.472874)(1867,0.449112)(1868,0.420473)(1869,0.361115)(1870,0.334433)(1871,0.308744)(1872,0.284278)(1873,0.26015)(1874,0.234732)(1875,0.206687)(1876,0.17607)(1877,0.198705)(1878,0.171468)(1879,0.143754)(1880,0.11693)(1881,0.119079)(1882,0.104471)(1883,0.0912295)(1884,0.0797335)(1885,2.1164e-005)(1886,1.88787e-005)(1887,2.84217e-005)(1888,6.33858e-005)(1889,0.052077)(1890,0.0522067)(1891,0.0513871)(1892,0.0507237)(1893,0.0207114)(1894,0.046902)(1895,0.0782438)(1896,0.103304)(1897,0.108223)(1898,0.104739)(1899,0.0979426)(1900,0.089945)(1901,0.0145256)(1902,0.0128009)(1903,0.0105296)(1904,0.00795937)(1905,0.00579975)(1906,0.00438062)(1907,0.00362869)(1908,0.00341936)(1909,0.0311032)(1910,0.0335897)(1911,0.0370185)(1912,0.0395197)(1913,0.146351)(1914,0.134035)(1915,0.107251)(1916,0.070714)(1917,0.0269989)(1918,0.00613843)(1919,0.00136919)(1920,0.000684802)(1921,0.000183542)(1922,0.000117973)(1923,0.0297575)(1924,0.030702)(1925,0.0312281)(1926,0.0311211)(1927,0.0306933)(1928,0.0305265)(1929,0.00548583)(1930,0.00627503)(1931,0.00724489)(1932,0.00850724)(1933,0.0324999)(1934,0.0408683)(1935,0.0490591)(1936,0.0559374)(1937,0.0610842)(1938,0.0650813)(1939,0.0688831)(1940,0.0731841)(1941,0.0781494)(1942,0.0834815)(1943,0.0885846)(1944,0.0926871)(1945,0.0950394)(1946,0.0440766)(1947,0.0415269)(1948,0.038693)(1949,0.0362253)(1950,0.034765)(1951,0.0346791)(1952,0.0359666)(1953,0.0295206)(1954,0.0237485)(1955,0.0199533)(1956,0.0181857)(1957,0.0182201)(1958,0.0197136)(1959,0.022131)(1960,0.0247036)(1961,0.0267298)(1962,0.015607)(1963,0.0119101)(1964,0.0320256)(1965,0.0337983)(1966,0.0370325)(1967,0.0414259)(1968,0.0463635)(1969,0.0509803)(1970,0.0305908)(1971,0.0305135)(1972,0.0300834)(1973,0.0292665)(1974,0.0283509)(1975,0.036157)(1976,0.0372891)(1977,0.0388937)(1978,0.0201759)(1979,0.0210235)(1980,0.0224847)(1981,0.024639)(1982,0.0287761)(1983,0.0301836)(1984,0.0324012)(1985,0.0354509)(1986,0.0392041)(1987,0.0432592)(1988,0.0469342)(1989,0.0322391)(1990,0.0307651)(1991,0.0295806)(1992,0.0288001)(1993,0.0211273)(1994,0.0197222)(1995,0.0186206)(1996,0.0177241)(1997,0.0169925)(1998,0.0164752)(1999,0.0162784)(2000,0.0165062)(2001,0.0398149)(2002,0.0407052)(2003,0.0416496)(2004,0.0424949)(2005,0.0431591)(2006,0.0436853)(2007,0.0442378)(2008,0.0450428)(2009,0.0242672)(2010,0.0245343)(2011,0.0255844) (2012,0)
};

\addplot [
color=tcolor6,
solid,
fill=tcolor6,
]
coordinates{
 (1790,7.18686e-006)(1791,4.91382e-006)(1792,3.94756e-006)(1793,3.51283e-006)(1794,3.32307e-006)(1795,3.42327e-006)(1796,4.03242e-006)(1797,3.56565e-006)(1798,2.70483e-006)(1799,2.5789e-006)(1800,3.07286e-006)(1801,2.67274e-006)(1802,1.98524e-006)(1803,1.70139e-006)(1804,1.59969e-006)(1805,1.57087e-006)(1806,1.60792e-006)(1807,1.79712e-006)(1808,2.30284e-006)(1809,4.11512e-006)(1810,3.16967e-006)(1811,2.74392e-006)(1812,2.53074e-006)(1813,2.36636e-006)(1814,2.24975e-006)(1815,2.30445e-006)(1816,2.70944e-006)(1817,1.86584e-006)(1818,1.41244e-006)(1819,1.22676e-006)(1820,1.15453e-006)(1821,1.1152e-006)(1822,1.10369e-006)(1823,1.178e-006)(1824,1.43486e-006)(1825,1.60634e-006)(1826,1.26341e-006)(1827,1.23167e-006)(1828,1.48729e-006)(1829,1.02927e-006)(1830,7.76807e-007)(1831,6.73865e-007)(1832,6.3463e-007)(1833,6.15269e-007)(1834,6.1546e-007)(1835,6.74083e-007)(1836,8.61676e-007)(1837,9.40265e-007)(1838,7.39958e-007)(1839,7.41256e-007)(1840,9.53044e-007)(1841,1.00836e-006)(1842,8.05186e-007)(1843,8.21367e-007)(1844,1.08055e-006)(1845,8.20307e-007)(1846,6.43049e-007)(1847,6.58462e-007)(1848,8.91495e-007)(1849,4.24592e-006)(1850,1.50406e-006)(1851,1.30715e-006)(1852,1.65069e-006)(1853,1.22483e-006)(1854,9.69965e-007)(1855,1.02894e-006)(1856,1.48739e-006)(1857,9.89478e-007)(1858,7.80875e-007)(1859,8.5451e-007)(1860,1.31928e-006)(1861,4.05746e-006)(1862,3.16334e-006)(1863,3.49217e-006)(1864,5.55188e-006)(1865,2.92954e-006)(1866,2.16574e-006)(1867,2.29795e-006)(1868,3.58223e-006)(1869,3.7325e-006)(1870,2.64604e-006)(1871,2.35894e-006)(1872,2.42017e-006)(1873,2.60651e-006)(1874,2.95486e-006)(1875,3.92024e-006)(1876,6.80192e-006)(1877,0.00143536)(1878,0.0023663)(1879,0.00363539)(1880,0.00509184)(1881,0.00127953)(1882,0.00171143)(1883,0.00220372)(1884,0.00270996)(1885,0.00932912)(1886,0.012346)(1887,0.0152914)(1888,0.0172906)(1889,0.00408178)(1890,0.00406903)(1891,0.00399074)(1892,0.00385487)(1893,0.010703)(1894,0.00898414)(1895,0.0075701)(1896,0.00651427)(1897,1.51824e-005)(1898,6.13839e-006)(1899,5.73164e-006)(1900,1.26228e-005)(1901,0.00011447)(1902,0.000168371)(1903,0.000631092)(1904,0.00290929)(1905,0.00896901)(1906,0.0190666)(1907,0.030214)(1908,0.0384372)(1909,0.0483624)(1910,0.0591675)(1911,0.0694401)(1912,0.0792489)(1913,0.0730473)(1914,0.0735065)(1915,0.0836571)(1916,0.077888)(1917,0.071532)(1918,0.0771442)(1919,0.0928907)(1920,0.116597)(1921,0.170206)(1922,0.189655)(1923,0.19674)(1924,0.204133)(1925,0.207736)(1926,0.209303)(1927,0.211485)(1928,0.216458)(1929,0.244759)(1930,0.263748)(1931,0.278033)(1932,0.286844)(1933,0.18165)(1934,0.176888)(1935,0.168119)(1936,0.15764)(1937,0.148525)(1938,0.142329)(1939,0.138911)(1940,0.137698)(1941,0.138646)(1942,0.142228)(1943,0.148776)(1944,0.157885)(1945,0.168343)(1946,0.32505)(1947,0.305753)(1948,0.276598)(1949,0.240999)(1950,0.205007)(1951,0.174207)(1952,0.151612)(1953,0.283932)(1954,0.28488)(1955,0.279761)(1956,0.26846)(1957,0.254926)(1958,0.245161)(1959,0.243613)(1960,0.250613)(1961,0.261983)(1962,0.151053)(1963,0.125396)(1964,0.148063)(1965,0.138623)(1966,0.132623)(1967,0.130088)(1968,0.130139)(1969,0.131347)(1970,0.120449)(1971,0.119862)(1972,0.120836)(1973,0.122314)(1974,0.123014)(1975,0.160193)(1976,0.151493)(1977,0.141197)(1978,0.217052)(1979,0.240508)(1980,0.258145)(1981,0.260876)(1982,0.10024)(1983,0.0823307)(1984,0.063858)(1985,0.046994)(1986,0.033148)(1987,0.0226204)(1988,0.0150011)(1989,0.0418339)(1990,0.0314201)(1991,0.0220235)(1992,0.0141428)(1993,3.21727e-005)(1994,1.52244e-005)(1995,9.34938e-006)(1996,6.64928e-006)(1997,4.94093e-006)(1998,3.9293e-006)(1999,3.79609e-006)(2000,4.96451e-006)(2001,4.28526e-006)(2002,2.5827e-006)(2003,2.00537e-006)(2004,1.82516e-006)(2005,1.78025e-006)(2006,1.86808e-006)(2007,2.31715e-006)(2008,3.7352e-006)(2009,8.21779e-006)(2010,6.77168e-006)(2011,9.79444e-006) (2012,0)
};

\addplot [
color=tcolor7,
solid,
fill=tcolor7,
]
coordinates{
 (1790,0.00118445)(1791,0.00131261)(1792,0.00144534)(1793,0.00150265)(1794,0.00141423)(1795,0.0011837)(1796,0.00089235)(1797,1.78364e-005)(1798,1.27621e-005)(1799,1.11141e-005)(1800,1.18695e-005)(1801,8.57297e-006)(1802,5.99444e-006)(1803,4.79452e-006)(1804,4.18905e-006)(1805,3.82418e-006)(1806,3.65244e-006)(1807,3.82941e-006)(1808,4.63002e-006)(1809,7.55081e-006)(1810,5.61668e-006)(1811,4.7012e-006)(1812,4.19765e-006)(1813,3.80631e-006)(1814,3.51708e-006)(1815,3.5101e-006)(1816,4.03112e-006)(1817,2.74847e-006)(1818,2.04534e-006)(1819,1.74789e-006)(1820,1.61959e-006)(1821,1.54119e-006)(1822,1.50364e-006)(1823,1.58339e-006)(1824,1.90461e-006)(1825,2.11215e-006)(1826,1.64609e-006)(1827,1.59046e-006)(1828,1.90393e-006)(1829,1.31162e-006)(1830,9.83822e-007)(1831,8.48345e-007)(1832,7.94288e-007)(1833,7.65705e-007)(1834,7.61818e-007)(1835,8.30192e-007)(1836,1.05642e-006)(1837,1.14313e-006)(1838,8.97851e-007)(1839,8.98215e-007)(1840,1.15405e-006)(1841,1.2082e-006)(1842,9.65607e-007)(1843,9.87166e-007)(1844,1.30327e-006)(1845,1.00448e-006)(1846,7.92503e-007)(1847,8.17655e-007)(1848,1.11649e-006)(1849,5.37008e-006)(1850,1.93215e-006)(1851,1.70024e-006)(1852,2.17541e-006)(1853,1.65294e-006)(1854,1.33397e-006)(1855,1.4424e-006)(1856,2.12506e-006)(1857,1.49595e-006)(1858,1.21544e-006)(1859,1.36466e-006)(1860,2.15018e-006)(1861,6.09574e-006)(1862,4.84502e-006)(1863,5.41804e-006)(1864,8.63853e-006)(1865,0.00456641)(1866,0.00607731)(1867,0.00661955)(1868,0.00598999)(1869,5.37067e-006)(1870,3.57678e-006)(1871,2.93718e-006)(1872,2.72662e-006)(1873,2.62444e-006)(1874,2.65135e-006)(1875,3.16762e-006)(1876,5.06311e-006)(1877,7.91322e-006)(1878,6.06407e-006)(1879,7.06964e-006)(1880,1.25706e-005)(1881,8.95164e-006)(1882,7.44911e-006)(1883,9.73282e-006)(1884,1.99439e-005)(1885,0.0174401)(1886,0.0256336)(1887,0.0326924)(1888,0.0359021)(1889,0.0197117)(1890,0.0272412)(1891,0.0349166)(1892,0.0421844)(1893,0.023432)(1894,0.0257011)(1895,0.0311132)(1896,0.040216)(1897,0.0299865)(1898,0.0319582)(1899,0.0347367)(1900,0.0390814)(1901,0.180299)(1902,0.174835)(1903,0.172149)(1904,0.176573)(1905,0.190357)(1906,0.213181)(1907,0.243101)(1908,0.276866)(1909,0.037845)(1910,0.041766)(1911,0.0479779)(1912,0.0561162)(1913,0.252322)(1914,0.237922)(1915,0.214497)(1916,0.244793)(1917,0.280454)(1918,0.273565)(1919,0.250659)(1920,0.230321)(1921,0.252012)(1922,0.244597)(1923,0.159785)(1924,0.157739)(1925,0.157534)(1926,0.158475)(1927,0.160202)(1928,0.163174)(1929,0.161311)(1930,0.190461)(1931,0.223631)(1932,0.254996)(1933,0.237216)(1934,0.246609)(1935,0.25078)(1936,0.249173)(1937,0.242217)(1938,0.229559)(1939,0.210845)(1940,0.187602)(1941,0.163041)(1942,0.140181)(1943,0.120569)(1944,0.104432)(1945,0.0914134)(1946,0.134796)(1947,0.135216)(1948,0.133224)(1949,0.126869)(1950,0.116293)(1951,0.103072)(1952,0.0888117)(1953,0.0946101)(1954,0.08468)(1955,0.0774954)(1956,0.0730709)(1957,0.0700376)(1958,0.0659508)(1959,0.0584289)(1960,0.0467919)(1961,0.0330221)(1962,0.0655317)(1963,0.0503198)(1964,2.46502e-005)(1965,1.19271e-005)(1966,8.16319e-006)(1967,6.95222e-006)(1968,7.34729e-006)(1969,1.0781e-005)(1970,2.12882e-005)(1971,1.06757e-005)(1972,7.61069e-006)(1973,7.50379e-006)(1974,1.03242e-005)(1975,9.0112e-006)(1976,5.78515e-006)(1977,7.68177e-006)(1978,4.98431e-006)(1979,2.68324e-006)(1980,2.48211e-006)(1981,4.14107e-006)(1982,6.05062e-006)(1983,4.04656e-006)(1984,3.68827e-006)(1985,3.92072e-006)(1986,4.366e-006)(1987,5.54282e-006)(1988,9.10812e-006)(1989,1.35791e-005)(1990,9.50429e-006)(1991,9.9837e-006)(1992,1.54342e-005)(1993,0.0021159)(1994,0.00244396)(1995,0.00244694)(1996,0.00213465)(1997,0.00167639)(1998,0.0012413)(1999,0.000914464)(2000,0.000706393)(2001,6.09789e-006)(2002,4.06828e-006)(2003,3.49577e-006)(2004,3.51686e-006)(2005,3.77694e-006)(2006,4.31769e-006)(2007,5.71003e-006)(2008,9.51517e-006)(2009,0.00484035)(2010,0.00588011)(2011,0.00661213) (2012,0)
};

\addplot [
color=tcolor9,
solid,
fill=tcolor9,
]
coordinates{
 (1790,1.39478e-005)(1791,1.00258e-005)(1792,8.50728e-006)(1793,8.02526e-006)(1794,8.0597e-006)(1795,8.79883e-006)(1796,1.09276e-005)(1797,1.24456e-005)(1798,1.05124e-005)(1799,1.10159e-005)(1800,1.40661e-005)(1801,2.75822e-005)(1802,2.60417e-005)(1803,2.72965e-005)(1804,3.00711e-005)(1805,3.32391e-005)(1806,3.65405e-005)(1807,4.09914e-005)(1808,4.84657e-005)(1809,0.00305718)(1810,0.00402422)(1811,0.00472612)(1812,0.00481694)(1813,0.00420647)(1814,0.00316602)(1815,0.00210495)(1816,0.00128609)(1817,9.07769e-006)(1818,6.36584e-006)(1819,5.01111e-006)(1820,4.23164e-006)(1821,3.67345e-006)(1822,3.29606e-006)(1823,3.22664e-006)(1824,3.64762e-006)(1825,4.21105e-006)(1826,3.1734e-006)(1827,2.95186e-006)(1828,3.39843e-006)(1829,2.1953e-006)(1830,1.62678e-006)(1831,1.38735e-006)(1832,1.28702e-006)(1833,1.23235e-006)(1834,1.2211e-006)(1835,1.32876e-006)(1836,1.69291e-006)(1837,1.85236e-006)(1838,1.48222e-006)(1839,1.51112e-006)(1840,1.97759e-006)(1841,2.16889e-006)(1842,1.79538e-006)(1843,1.89548e-006)(1844,2.56931e-006)(1845,1.95399e-006)(1846,1.59613e-006)(1847,1.68976e-006)(1848,2.33651e-006)(1849,0.00141751)(1850,4.08524e-006)(1851,3.61032e-006)(1852,4.56911e-006)(1853,3.44919e-006)(1854,2.81676e-006)(1855,3.05556e-006)(1856,4.46319e-006)(1857,3.37411e-006)(1858,2.91352e-006)(1859,3.44504e-006)(1860,5.58621e-006)(1861,0.006321)(1862,0.0122638)(1863,0.0216944)(1864,0.033469)(1865,1.16003e-005)(1866,1.00944e-005)(1867,1.23257e-005)(1868,2.06457e-005)(1869,0.0310543)(1870,0.0449195)(1871,0.0590373)(1872,0.0710555)(1873,0.0798016)(1874,0.0855906)(1875,0.0895784)(1876,0.0929116)(1877,0.0266041)(1878,0.0333149)(1879,0.0410041)(1880,0.0495005)(1881,0.0970256)(1882,0.117488)(1883,0.137358)(1884,0.15442)(1885,0.0828717)(1886,0.0884046)(1887,0.0956503)(1888,0.104415)(1889,0.104059)(1890,0.103777)(1891,0.101759)(1892,0.0988545)(1893,0.133054)(1894,0.124923)(1895,0.111856)(1896,0.0973872)(1897,0.310772)(1898,0.307493)(1899,0.286068)(1900,0.247742)(1901,0.0295477)(1902,0.0169216)(1903,0.00685732)(1904,0.00201859)(1905,0.000470378)(1906,0.000119639)(1907,5.66021e-005)(1908,6.48934e-005)(1909,0.032322)(1910,0.0330448)(1911,0.0342667)(1912,0.0335007)(1913,0.0254906)(1914,0.0369628)(1915,0.0577367)(1916,0.0480717)(1917,0.0294202)(1918,0.016643)(1919,0.00787703)(1920,0.00362695)(1921,3.07247e-005)(1922,2.08894e-005)(1923,1.5595e-005)(1924,5.89055e-006)(1925,3.84752e-006)(1926,3.16069e-006)(1927,3.24888e-006)(1928,5.52843e-006)(1929,9.58567e-006)(1930,4.80154e-006)(1931,4.61352e-006)(1932,8.08109e-006)(1933,3.79068e-005)(1934,1.55916e-005)(1935,7.63381e-006)(1936,5.37384e-006)(1937,4.42247e-006)(1938,3.64649e-006)(1939,3.00066e-006)(1940,2.51934e-006)(1941,2.16683e-006)(1942,1.8897e-006)(1943,1.71351e-006)(1944,1.91512e-006)(1945,3.20341e-006)(1946,2.64905e-006)(1947,1.59022e-006)(1948,1.45542e-006)(1949,1.58978e-006)(1950,1.71188e-006)(1951,1.83744e-006)(1952,2.38327e-006)(1953,6.89994e-006)(1954,3.96926e-006)(1955,3.38284e-006)(1956,3.52858e-006)(1957,3.84185e-006)(1958,4.22831e-006)(1959,4.75082e-006)(1960,6.33945e-006)(1961,1.24027e-005)(1962,0.00226239)(1963,0.00320488)(1964,1.20707e-005)(1965,6.16547e-006)(1966,4.58014e-006)(1967,4.30247e-006)(1968,5.02353e-006)(1969,8.08096e-006)(1970,1.45961e-005)(1971,7.72691e-006)(1972,5.90374e-006)(1973,6.27347e-006)(1974,9.27242e-006)(1975,9.85293e-006)(1976,6.50881e-006)(1977,8.82871e-006)(1978,0.000158764)(1979,0.000226753)(1980,0.000420954)(1981,0.000682683)(1982,6.73266e-006)(1983,4.32482e-006)(1984,3.72729e-006)(1985,3.6953e-006)(1986,3.80229e-006)(1987,4.44556e-006)(1988,6.7423e-006)(1989,7.65182e-006)(1990,4.74412e-006)(1991,4.47351e-006)(1992,6.37376e-006)(1993,4.82242e-006)(1994,2.57509e-006)(1995,1.92041e-006)(1996,1.68121e-006)(1997,1.50173e-006)(1998,1.38795e-006)(1999,1.51114e-006)(2000,2.1732e-006)(2001,1.84639e-006)(2002,1.14537e-006)(2003,9.1964e-007)(2004,8.69236e-007)(2005,8.82284e-007)(2006,9.63237e-007)(2007,1.24114e-006)(2008,2.07368e-006)(2009,3.56771e-006)(2010,2.95652e-006)(2011,4.41146e-006) (2012,0)
};

\end{axis}
\end{tikzpicture}

%% file: MimnoStateOfTheUnion_stacked.tikz
%
%
\begin{tikzpicture}

\definecolor{mycolor1}{rgb}{0.5,0.5,0}
\definecolor{mycolor2}{rgb}{0,1,1}
\definecolor{mycolor3}{rgb}{1,0,1}
\definecolor{mycolor4}{rgb}{1,1,0}

\definecolor{tcolor10}{rgb}{1.0,0.0,0.0}
\definecolor{tcolor3}{rgb}{0.75,0.0,0.0}
\definecolor{tcolor4}{rgb}{0.5,0.0,0.0}
\definecolor{tcolor1}{rgb}{0.0,0.5,0.0}
\definecolor{tcolor8}{rgb}{0.0,0.75,0.0}
\definecolor{tcolor2}{rgb}{0.0,1.0,0.0}
\definecolor{tcolor5}{rgb}{0.0,0.0,0.25}
\definecolor{tcolor6}{rgb}{0.0,0.0,0.5}
\definecolor{tcolor7}{rgb}{0.0,0.0,0.75}
\definecolor{tcolor9}{rgb}{0.0,0.0,1.0}

\pgfkeys{/pgf/number format/.cd, set thousands separator={}}

\begin{axis}[%
scale only axis,
stack plots=y,
width=360pt,
height=225pt,
xmin=1790, xmax=2012,
ymin=0, ymax=1.0,
xlabel={year},
ylabel={$\langle\pi_k\g\phi\rangle$},
axis on top]

\node[text=white,text centered,text width=3.0cm] at (axis cs:1830,0.05) {\tiny{shall, made, present}};
\node[text=white,text centered,text width=3.0cm] at (axis cs:1950,0.2) {\tiny{new, war, national}};
\node[text=white,text centered,text width=4.0cm] at (axis cs:1970,0.4) {\tiny{America, peace, American}};
\node[text=white,text centered,text width=3.0cm] at (axis cs:1970,0.56) {\tiny{work, legislation, act}};
\node[text=white,text centered,text width=3.0cm] at (axis cs:1840,0.35) {\tiny{treasury, amount, public}};
\node[text=white,text centered,text width=3.0cm] at (axis cs:1970,0.7) {\tiny{people, world, nations}};
\node[text=white,text centered,text width=3.0cm] at (axis cs:1850,0.55) {\tiny{public, power, people}};
\node[text=white,text centered,text width=3.0cm] at (axis cs:1850,0.65) {\tiny{citizens, war, commerce}};
\node[text=white,text centered,text width=3.0cm] at (axis cs:1850,0.8) {\tiny{department, act, interest}};
\node[text=white,text centered,text width=3.0cm] at (axis cs:1850,0.95) {\tiny{law, secretary, service}};

\addplot [
color=tcolor1,
solid,
fill=tcolor1
]
coordinates{
(1790,0) (1790,0.133693)(1791,0.13775)(1792,0.140534)(1793,0.141284)(1794,0.139456)(1795,0.134956)(1796,0.128212)(1797,0.136703)(1798,0.127727)(1799,0.119232)(1800,0.111871)(1801,0.143986)(1802,0.139866)(1803,0.137897)(1804,0.13806)(1805,0.140173)(1806,0.143886)(1807,0.148687)(1808,0.15394)(1809,0.128025)(1810,0.132325)(1811,0.135688)(1812,0.138031)(1813,0.139437)(1814,0.140065)(1815,0.140101)(1816,0.139768)(1817,0.118492)(1818,0.118703)(1819,0.11941)(1820,0.120612)(1821,0.122117)(1822,0.123604)(1823,0.12474)(1824,0.125284)(1825,0.138697)(1826,0.138256)(1827,0.137305)(1828,0.135958)(1829,0.130098)(1830,0.12872)(1831,0.127243)(1832,0.125748)(1833,0.124361)(1834,0.123248)(1835,0.122586)(1836,0.12251)(1837,0.117695)(1838,0.118948)(1839,0.120417)(1840,0.121749)(1841,0.132631)(1842,0.132984)(1843,0.132734)(1844,0.132153)(1845,0.120146)(1846,0.120872)(1847,0.122455)(1848,0.124951)(1849,0.137755)(1850,0.122892)(1851,0.126479)(1852,0.129202)(1853,0.134208)(1854,0.135301)(1855,0.135522)(1856,0.135571)(1857,0.123868)(1858,0.125798)(1859,0.128958)(1860,0.133034)(1861,0.108575)(1862,0.11159)(1863,0.113068)(1864,0.112359)(1865,0.136748)(1866,0.132568)(1867,0.126966)(1868,0.121143)(1869,0.123276)(1870,0.12061)(1871,0.119628)(1872,0.120013)(1873,0.12111)(1874,0.122152)(1875,0.122483)(1876,0.121732)(1877,0.114916)(1878,0.112639)(1879,0.109965)(1880,0.107278)(1881,0.116145)(1882,0.114276)(1883,0.112737)(1884,0.111399)(1885,0.131274)(1886,0.130557)(1887,0.129905)(1888,0.129426)(1889,0.124001)(1890,0.123995)(1891,0.124093)(1892,0.124121)(1893,0.129132)(1894,0.128476)(1895,0.127276)(1896,0.125575)(1897,0.1373)(1898,0.135424)(1899,0.133673)(1900,0.132222)(1901,0.157841)(1902,0.157456)(1903,0.15724)(1904,0.156911)(1905,0.156271)(1906,0.155285)(1907,0.154175)(1908,0.153457)(1909,0.126849)(1910,0.132137)(1911,0.139306)(1912,0.148143)(1913,0.154784)(1914,0.162123)(1915,0.167284)(1916,0.168662)(1917,0.165502)(1918,0.158232)(1919,0.148287)(1920,0.137519)(1921,0.132971)(1922,0.125153)(1923,0.135674)(1924,0.132093)(1925,0.129986)(1926,0.12827)(1927,0.12585)(1928,0.121944)(1929,0.123641)(1930,0.117286)(1931,0.110122)(1932,0.103109)(1933,0.129789)(1934,0.123495)(1935,0.118973)(1936,0.116013)(1937,0.114224)(1938,0.113205)(1939,0.11267)(1940,0.112515)(1941,0.112805)(1942,0.113715)(1943,0.115423)(1944,0.118016)(1945,0.121413)(1946,0.0908225)(1947,0.0962112)(1948,0.101331)(1949,0.105759)(1950,0.109297)(1951,0.112031)(1952,0.114242)(1953,0.105164)(1954,0.108527)(1955,0.111346)(1956,0.113476)(1957,0.114804)(1958,0.115342)(1959,0.115258)(1960,0.114843)(1961,0.11448)(1962,0.106257)(1963,0.107909)(1964,0.0947709)(1965,0.0984161)(1966,0.10184)(1967,0.104083)(1968,0.104129)(1969,0.101366)(1970,0.125134)(1971,0.117835)(1972,0.109403)(1973,0.101363)(1974,0.0948577)(1975,0.0772789)(1976,0.075297)(1977,0.0746412)(1978,0.0754648)(1979,0.0757748)(1980,0.075512)(1981,0.0744099)(1982,0.0849787)(1983,0.0830873)(1984,0.0816206)(1985,0.0812975)(1986,0.0826393)(1987,0.0858689)(1988,0.0908339)(1989,0.0790291)(1990,0.0843847)(1991,0.0889003)(1992,0.0917328)(1993,0.0740607)(1994,0.0733851)(1995,0.0719023)(1996,0.0704215)(1997,0.0697067)(1998,0.0703008)(1999,0.0724199)(2000,0.0758762)(2001,0.0790464)(2002,0.0829222)(2003,0.0852613)(2004,0.0850419)(2005,0.0818515)(2006,0.0760992)(2007,0.0688755)(2008,0.0615258)(2009,0.0602155)(2010,0.0553468)(2011,0.0526938) (2012,0)
};

\addplot [
color=tcolor2,
solid,
fill=tcolor2
]
coordinates{
(1790,0) (1790,0.0549105)(1791,0.0548718)(1792,0.0559268)(1793,0.0575982)(1794,0.0592812)(1795,0.0603614)(1796,0.0603927)(1797,0.0797272)(1798,0.0774154)(1799,0.0744959)(1800,0.0716484)(1801,0.0567904)(1802,0.0565303)(1803,0.0573293)(1804,0.0592691)(1805,0.0622552)(1806,0.0659975)(1807,0.0700089)(1808,0.0736679)(1809,0.0658424)(1810,0.0674024)(1811,0.0676391)(1812,0.0667378)(1813,0.0650783)(1814,0.0630853)(1815,0.0611278)(1816,0.0594876)(1817,0.0497957)(1818,0.0495318)(1819,0.0498626)(1820,0.0506697)(1821,0.0517104)(1822,0.0526475)(1823,0.0531224)(1824,0.0528534)(1825,0.067951)(1826,0.0656422)(1827,0.0626546)(1828,0.0594208)(1829,0.0707348)(1830,0.0678813)(1831,0.0660789)(1832,0.0654692)(1833,0.066021)(1834,0.0675301)(1835,0.0696267)(1836,0.0718177)(1837,0.0629695)(1838,0.0639479)(1839,0.0641803)(1840,0.0639127)(1841,0.0533515)(1842,0.0537525)(1843,0.0552833)(1844,0.0583858)(1845,0.0442655)(1846,0.0492404)(1847,0.0551889)(1848,0.0613104)(1849,0.0547875)(1850,0.0606096)(1851,0.0610544)(1852,0.0592073)(1853,0.0632675)(1854,0.0594708)(1855,0.0563082)(1856,0.0545265)(1857,0.0517464)(1858,0.0535355)(1859,0.0567512)(1860,0.0607357)(1861,0.0668922)(1862,0.0694677)(1863,0.0695383)(1864,0.0667663)(1865,0.0826887)(1866,0.0753799)(1867,0.067848)(1868,0.0613075)(1869,0.0651331)(1870,0.0621846)(1871,0.0611411)(1872,0.0615648)(1873,0.0628431)(1874,0.0643131)(1875,0.0654224)(1876,0.065884)(1877,0.0582847)(1878,0.0580601)(1879,0.0579658)(1880,0.0583268)(1881,0.0668386)(1882,0.0688241)(1883,0.0711107)(1884,0.0731722)(1885,0.0580617)(1886,0.0584837)(1887,0.057967)(1888,0.0567615)(1889,0.0606248)(1890,0.0592152)(1891,0.0583842)(1892,0.0583278)(1893,0.0538555)(1894,0.0549525)(1895,0.0560703)(1896,0.05668)(1897,0.0705206)(1898,0.0687525)(1899,0.0655366)(1900,0.0613923)(1901,0.129926)(1902,0.121772)(1903,0.116111)(1904,0.113723)(1905,0.114981)(1906,0.119886)(1907,0.128065)(1908,0.13873)(1909,0.0635819)(1910,0.0703686)(1911,0.0765361)(1912,0.0813291)(1913,0.142102)(1914,0.142747)(1915,0.141373)(1916,0.138947)(1917,0.136378)(1918,0.134283)(1919,0.132919)(1920,0.132244)(1921,0.140737)(1922,0.141445)(1923,0.123507)(1924,0.124507)(1925,0.12547)(1926,0.126511)(1927,0.127782)(1928,0.129399)(1929,0.135817)(1930,0.139371)(1931,0.14298)(1932,0.146496)(1933,0.153272)(1934,0.156617)(1935,0.159785)(1936,0.163007)(1937,0.166564)(1938,0.170732)(1939,0.175707)(1940,0.18153)(1941,0.188042)(1942,0.194887)(1943,0.201571)(1944,0.207545)(1945,0.212292)(1946,0.164988)(1947,0.170266)(1948,0.173712)(1949,0.175084)(1950,0.174441)(1951,0.172151)(1952,0.168787)(1953,0.162511)(1954,0.161342)(1955,0.160024)(1956,0.159115)(1957,0.1591)(1958,0.160275)(1959,0.162589)(1960,0.165563)(1961,0.168386)(1962,0.157653)(1963,0.158946)(1964,0.161442)(1965,0.161114)(1966,0.159616)(1967,0.15776)(1968,0.156331)(1969,0.155976)(1970,0.151399)(1971,0.156819)(1972,0.163599)(1973,0.171198)(1974,0.178904)(1975,0.171672)(1976,0.176771)(1977,0.179739)(1978,0.179708)(1979,0.178744)(1980,0.176087)(1981,0.172316)(1982,0.183078)(1983,0.179006)(1984,0.174673)(1985,0.170195)(1986,0.165674)(1987,0.161234)(1988,0.157017)(1989,0.154867)(1990,0.151912)(1991,0.149508)(1992,0.147683)(1993,0.164025)(1994,0.163167)(1995,0.16237)(1996,0.161407)(1997,0.160224)(1998,0.158946)(1999,0.157786)(2000,0.156931)(2001,0.147745)(2002,0.148036)(2003,0.148899)(2004,0.150311)(2005,0.152078)(2006,0.153832)(2007,0.155156)(2008,0.155784)(2009,0.181027)(2010,0.180841)(2011,0.180402) (2012,0)
};

\addplot [
color=tcolor3,
solid,
fill=tcolor3
]
coordinates{
(1790,0) (1790,0.0151667)(1791,0.0159844)(1792,0.0181966)(1793,0.0219466)(1794,0.0273006)(1795,0.0340195)(1796,0.0413673)(1797,0.0562701)(1798,0.0625264)(1799,0.0659374)(1800,0.0664462)(1801,0.0709764)(1802,0.0684655)(1803,0.0654079)(1804,0.0624467)(1805,0.0598789)(1806,0.0577378)(1807,0.055912)(1808,0.054261)(1809,0.0525876)(1810,0.051478)(1811,0.0505504)(1812,0.0499774)(1813,0.0499174)(1814,0.0504529)(1815,0.0515637)(1816,0.0531325)(1817,0.027436)(1818,0.0284575)(1819,0.0293952)(1820,0.0301072)(1821,0.0304742)(1822,0.0304385)(1823,0.0300318)(1824,0.0293703)(1825,0.0419502)(1826,0.0410681)(1827,0.0404715)(1828,0.0402373)(1829,0.0278722)(1830,0.028245)(1831,0.0287474)(1832,0.0292901)(1833,0.0297987)(1834,0.0302324)(1835,0.030592)(1836,0.0309155)(1837,0.0211)(1838,0.0214371)(1839,0.0218654)(1840,0.0224087)(1841,0.0329884)(1842,0.0341846)(1843,0.0355959)(1844,0.0371855)(1845,0.0313917)(1846,0.0328784)(1847,0.0341519)(1848,0.0349653)(1849,0.0301173)(1850,0.0326572)(1851,0.0313103)(1852,0.0293033)(1853,0.0431077)(1854,0.0395053)(1855,0.0362953)(1856,0.0338287)(1857,0.0351659)(1858,0.0346366)(1859,0.03506)(1860,0.0361554)(1861,0.036444)(1862,0.037374)(1863,0.03719)(1864,0.0354233)(1865,0.0379688)(1866,0.0332872)(1867,0.0281786)(1868,0.0235008)(1869,0.0313841)(1870,0.0275605)(1871,0.0254672)(1872,0.0249651)(1873,0.0259066)(1874,0.0281507)(1875,0.0315062)(1876,0.0356553)(1877,0.0242369)(1878,0.0268679)(1879,0.0290875)(1880,0.0307448)(1881,0.0286207)(1882,0.0293387)(1883,0.0297904)(1884,0.0300401)(1885,0.0217996)(1886,0.0217345)(1887,0.0213819)(1888,0.0206965)(1889,0.0189062)(1890,0.0177079)(1891,0.0164006)(1892,0.0151204)(1893,0.0145655)(1894,0.013594)(1895,0.0128566)(1896,0.0123381)(1897,0.0198245)(1898,0.0195578)(1899,0.0194312)(1900,0.0193858)(1901,0.0361271)(1902,0.0363648)(1903,0.0367359)(1904,0.0372743)(1905,0.0380001)(1906,0.0389017)(1907,0.0399464)(1908,0.0411172)(1909,0.0204341)(1910,0.0217673)(1911,0.0233051)(1912,0.0250726)(1913,0.0759143)(1914,0.0812339)(1915,0.0867667)(1916,0.0919235)(1917,0.0959629)(1918,0.098148)(1919,0.0979827)(1920,0.0954289)(1921,0.0731819)(1922,0.0690672)(1923,0.0693386)(1924,0.0655389)(1925,0.0627451)(1926,0.0612671)(1927,0.0612851)(1928,0.0629038)(1929,0.0695271)(1930,0.075478)(1931,0.0833999)(1932,0.0934126)(1933,0.0865027)(1934,0.0982701)(1935,0.111456)(1936,0.125565)(1937,0.139864)(1938,0.153512)(1939,0.165772)(1940,0.17624)(1941,0.184979)(1942,0.19252)(1943,0.199713)(1944,0.207469)(1945,0.216488)(1946,0.1578)(1947,0.170201)(1948,0.182696)(1949,0.193773)(1950,0.201869)(1951,0.205911)(1952,0.205742)(1953,0.181002)(1954,0.178806)(1955,0.175565)(1956,0.172939)(1957,0.172306)(1958,0.174551)(1959,0.179894)(1960,0.187784)(1961,0.196929)(1962,0.21467)(1963,0.222781)(1964,0.210245)(1965,0.21198)(1966,0.209392)(1967,0.203507)(1968,0.195731)(1969,0.187481)(1970,0.246891)(1971,0.242588)(1972,0.239979)(1973,0.238853)(1974,0.238807)(1975,0.232515)(1976,0.233169)(1977,0.233445)(1978,0.233366)(1979,0.234387)(1980,0.23636)(1981,0.239916)(1982,0.257441)(1983,0.266189)(1984,0.275453)(1985,0.283706)(1986,0.28931)(1987,0.291071)(1988,0.288738)(1989,0.281721)(1990,0.275544)(1991,0.269923)(1992,0.266712)(1993,0.261651)(1994,0.266176)(1995,0.27374)(1996,0.282915)(1997,0.291959)(1998,0.299346)(1999,0.30426)(2000,0.306878)(2001,0.264811)(2002,0.267209)(2003,0.272275)(2004,0.28156)(2005,0.295711)(2006,0.314045)(2007,0.334376)(2008,0.353295)(2009,0.355115)(2010,0.360918)(2011,0.356724) (2012,0)
};

\addplot [
color=tcolor4,
solid,
fill=tcolor4
]
coordinates{
(1790,0) (1790,0.0212649)(1791,0.0226579)(1792,0.0250128)(1793,0.028058)(1794,0.0312485)(1795,0.0337852)(1796,0.0348519)(1797,0.0398826)(1798,0.0370425)(1799,0.0328942)(1800,0.0283696)(1801,0.0509798)(1802,0.0444006)(1803,0.0396118)(1804,0.0365841)(1805,0.0351074)(1806,0.0348993)(1807,0.0356371)(1808,0.0369585)(1809,0.0369403)(1810,0.0384762)(1811,0.0395655)(1812,0.0400723)(1813,0.0400081)(1814,0.0394874)(1815,0.0386775)(1816,0.0377627)(1817,0.0233694)(1818,0.0230556)(1819,0.0229708)(1820,0.0231248)(1821,0.0234756)(1822,0.0239374)(1823,0.0244001)(1824,0.0247553)(1825,0.0419075)(1826,0.0419661)(1827,0.0417535)(1828,0.0414185)(1829,0.0371297)(1830,0.0373309)(1831,0.0380293)(1832,0.039337)(1833,0.0412497)(1834,0.0436018)(1835,0.0460378)(1836,0.0480418)(1837,0.0449854)(1838,0.0447767)(1839,0.0433071)(1840,0.0409343)(1841,0.0315438)(1842,0.0296432)(1843,0.0284521)(1844,0.0282574)(1845,0.0272028)(1846,0.0294812)(1847,0.0330312)(1848,0.0377527)(1849,0.0378889)(1850,0.0409147)(1851,0.0457957)(1852,0.0499203)(1853,0.0441003)(1854,0.0459345)(1855,0.0469281)(1856,0.0472851)(1857,0.0358966)(1858,0.0356996)(1859,0.0353374)(1860,0.0348333)(1861,0.0576837)(1862,0.0567559)(1863,0.0558849)(1864,0.0551815)(1865,0.0466953)(1866,0.0471898)(1867,0.0478517)(1868,0.0485988)(1869,0.0551005)(1870,0.0563558)(1871,0.0574781)(1872,0.0584034)(1873,0.0591054)(1874,0.059605)(1875,0.0599431)(1876,0.0601286)(1877,0.070762)(1878,0.0705047)(1879,0.0696621)(1880,0.0681122)(1881,0.0705454)(1882,0.0678934)(1883,0.0650449)(1884,0.0623958)(1885,0.0854577)(1886,0.0840622)(1887,0.0838816)(1888,0.0848437)(1889,0.0960534)(1890,0.0986552)(1891,0.101435)(1892,0.10422)(1893,0.0967347)(1894,0.0998898)(1895,0.10405)(1896,0.109734)(1897,0.0773474)(1898,0.0839194)(1899,0.0915738)(1900,0.0997763)(1901,0.128975)(1902,0.137955)(1903,0.145277)(1904,0.150711)(1905,0.154434)(1906,0.156818)(1907,0.158208)(1908,0.158834)(1909,0.121421)(1910,0.124176)(1911,0.126237)(1912,0.127363)(1913,0.116711)(1914,0.114945)(1915,0.112973)(1916,0.111093)(1917,0.109549)(1918,0.108442)(1919,0.107708)(1920,0.107165)(1921,0.130534)(1922,0.130295)(1923,0.126482)(1924,0.126423)(1925,0.12659)(1926,0.127335)(1927,0.128991)(1928,0.131749)(1929,0.152202)(1930,0.158663)(1931,0.165153)(1932,0.170726)(1933,0.122936)(1934,0.123828)(1935,0.122722)(1936,0.119983)(1937,0.116315)(1938,0.112533)(1939,0.109309)(1940,0.106988)(1941,0.105502)(1942,0.10438)(1943,0.102855)(1944,0.100069)(1945,0.0953648)(1946,0.178315)(1947,0.165463)(1948,0.149678)(1949,0.132838)(1950,0.116766)(1951,0.102762)(1952,0.091451)(1953,0.138741)(1954,0.130587)(1955,0.125192)(1956,0.122116)(1957,0.120954)(1958,0.121354)(1959,0.122966)(1960,0.125403)(1961,0.128287)(1962,0.119953)(1963,0.123752)(1964,0.122848)(1965,0.128199)(1966,0.133619)(1967,0.138814)(1968,0.143206)(1969,0.146079)(1970,0.117237)(1971,0.118028)(1972,0.116968)(1973,0.114617)(1974,0.111934)(1975,0.13412)(1976,0.133628)(1977,0.135541)(1978,0.125069)(1979,0.131503)(1980,0.1392)(1981,0.146949)(1982,0.125247)(1983,0.129022)(1984,0.13011)(1985,0.128738)(1986,0.12577)(1987,0.12235)(1988,0.119534)(1989,0.125832)(1990,0.1265)(1991,0.129095)(1992,0.1334)(1993,0.11554)(1994,0.120647)(1995,0.125258)(1996,0.128518)(1997,0.129771)(1998,0.128718)(1999,0.125441)(2000,0.12035)(2001,0.131473)(2002,0.124001)(2003,0.117012)(2004,0.111246)(2005,0.10715)(2006,0.104882)(2007,0.104413)(2008,0.105649)(2009,0.0887692)(2010,0.0925924)(2011,0.0975656) (2012,0)
};

\addplot [
color=tcolor5,
solid,
fill=tcolor5
]
coordinates{
(1790,0) (1790,0.162844)(1791,0.164432)(1792,0.161935)(1793,0.155984)(1794,0.147745)(1795,0.138589)(1796,0.129747)(1797,0.132398)(1798,0.126721)(1799,0.123018)(1800,0.121063)(1801,0.134185)(1802,0.135871)(1803,0.137774)(1804,0.139359)(1805,0.140173)(1806,0.139922)(1807,0.13853)(1808,0.136185)(1809,0.152718)(1810,0.150484)(1811,0.148876)(1812,0.148407)(1813,0.149242)(1814,0.151066)(1815,0.153108)(1816,0.154349)(1817,0.205702)(1818,0.202785)(1819,0.197275)(1820,0.189921)(1821,0.181847)(1822,0.174221)(1823,0.168019)(1824,0.163861)(1825,0.15794)(1826,0.158565)(1827,0.160717)(1828,0.163548)(1829,0.15279)(1830,0.154753)(1831,0.155503)(1832,0.155108)(1833,0.154045)(1834,0.153036)(1835,0.15284)(1836,0.154068)(1837,0.144283)(1838,0.148855)(1839,0.154433)(1840,0.160024)(1841,0.147173)(1842,0.14873)(1843,0.147058)(1844,0.141954)(1845,0.184995)(1846,0.172959)(1847,0.160818)(1848,0.150316)(1849,0.135664)(1850,0.160546)(1851,0.160298)(1852,0.163198)(1853,0.140338)(1854,0.145115)(1855,0.148215)(1856,0.148426)(1857,0.153338)(1858,0.148295)(1859,0.142621)(1860,0.138189)(1861,0.138217)(1862,0.14106)(1863,0.1485)(1864,0.160372)(1865,0.120676)(1866,0.133884)(1867,0.14616)(1868,0.155324)(1869,0.162874)(1870,0.164461)(1871,0.162488)(1872,0.158518)(1873,0.154061)(1874,0.150203)(1875,0.147472)(1876,0.14588)(1877,0.162377)(1878,0.162139)(1879,0.161606)(1880,0.160442)(1881,0.16294)(1882,0.161282)(1883,0.159585)(1884,0.158242)(1885,0.146745)(1886,0.147715)(1887,0.149511)(1888,0.152086)(1889,0.150368)(1890,0.153825)(1891,0.157339)(1892,0.160579)(1893,0.168571)(1894,0.17013)(1895,0.170005)(1896,0.167749)(1897,0.158213)(1898,0.152095)(1899,0.14417)(1900,0.135148)(1901,0.07457)(1902,0.0696178)(1903,0.0652462)(1904,0.0615841)(1905,0.0586019)(1906,0.0561371)(1907,0.0539588)(1908,0.051854)(1909,0.108211)(1910,0.106144)(1911,0.103696)(1912,0.100987)(1913,0.0557084)(1914,0.053823)(1915,0.0524014)(1916,0.0515143)(1917,0.0511851)(1918,0.0513848)(1919,0.0520479)(1920,0.0531077)(1921,0.070609)(1922,0.0733532)(1923,0.0708568)(1924,0.0752215)(1925,0.0807772)(1926,0.0877272)(1927,0.0960893)(1928,0.105504)(1929,0.0754852)(1930,0.0814851)(1931,0.0851715)(1932,0.0852625)(1933,0.054525)(1934,0.0492537)(1935,0.0423277)(1936,0.0349503)(1937,0.0281379)(1938,0.0224743)(1939,0.0181174)(1940,0.0149483)(1941,0.0127288)(1942,0.0112019)(1943,0.0101332)(1944,0.0093206)(1945,0.00859771)(1946,0.0281812)(1947,0.0257977)(1948,0.0228909)(1949,0.0196575)(1950,0.0164212)(1951,0.0134871)(1952,0.0110429)(1953,0.0379128)(1954,0.0326184)(1955,0.0288381)(1956,0.026337)(1957,0.0248902)(1958,0.0243098)(1959,0.0244383)(1960,0.0251338)(1961,0.0262668)(1962,0.0201299)(1963,0.0215134)(1964,0.0166353)(1965,0.0179142)(1966,0.0190915)(1967,0.0200044)(1968,0.020464)(1969,0.0203226)(1970,0.00962831)(1971,0.00915475)(1972,0.00850907)(1973,0.00782632)(1974,0.00723722)(1975,0.0101282)(1976,0.0099163)(1977,0.0101252)(1978,0.010296)(1979,0.0114243)(1980,0.0130052)(1981,0.0149683)(1982,0.00758743)(1983,0.00859208)(1984,0.00946571)(1985,0.0101007)(1986,0.0104427)(1987,0.0104981)(1988,0.0103152)(1989,0.00802356)(1990,0.00766205)(1991,0.00723695)(1992,0.00677393)(1993,0.0072854)(1994,0.00674111)(1995,0.00621719)(1996,0.0057355)(1997,0.00531535)(1998,0.00496779)(1999,0.00469112)(2000,0.0044702)(2001,0.00915555)(2002,0.00877454)(2003,0.00836822)(2004,0.00792858)(2005,0.007482)(2006,0.00708532)(2007,0.00681809)(2008,0.00677747)(2009,0.001775)(2010,0.00198351)(2011,0.00238166) (2012,0)
};

\addplot [
color=tcolor6,
solid,
fill=tcolor6
]
coordinates{
(1790,0) (1790,0.0343138)(1791,0.0346383)(1792,0.0368852)(1793,0.0410496)(1794,0.0470714)(1795,0.0547261)(1796,0.0635482)(1797,0.0628637)(1798,0.0710837)(1799,0.0786078)(1800,0.0849625)(1801,0.0479578)(1802,0.0502106)(1803,0.0514673)(1804,0.051636)(1805,0.0506593)(1806,0.0485725)(1807,0.0455606)(1808,0.0419685)(1809,0.0549058)(1810,0.0503224)(1811,0.0466057)(1812,0.0440969)(1813,0.0429512)(1814,0.0431633)(1815,0.0445936)(1816,0.0469796)(1817,0.0303675)(1818,0.0323452)(1819,0.0341487)(1820,0.0355211)(1821,0.0363023)(1822,0.036469)(1823,0.0361275)(1824,0.0354586)(1825,0.0446937)(1826,0.0437517)(1827,0.0428649)(1828,0.0420087)(1829,0.0465834)(1830,0.0456924)(1831,0.044815)(1832,0.0440814)(1833,0.0436808)(1834,0.0438129)(1835,0.0446352)(1836,0.0462118)(1837,0.0411357)(1838,0.0434953)(1839,0.045835)(1840,0.0476785)(1841,0.0582193)(1842,0.0580253)(1843,0.0566318)(1844,0.0544772)(1845,0.0368172)(1846,0.035614)(1847,0.0349166)(1848,0.0348233)(1849,0.0426265)(1850,0.0389035)(1851,0.0400661)(1852,0.0408892)(1853,0.0469358)(1854,0.0462723)(1855,0.044552)(1856,0.0420628)(1857,0.0451125)(1858,0.0420096)(1859,0.0392528)(1860,0.0369488)(1861,0.0576293)(1862,0.0551194)(1863,0.0527221)(1864,0.0501439)(1865,0.0459521)(1866,0.0434305)(1867,0.0407009)(1868,0.0380399)(1869,0.0550209)(1870,0.0527765)(1871,0.0514485)(1872,0.0509598)(1873,0.0511012)(1874,0.0516198)(1875,0.0523138)(1876,0.0531144)(1877,0.0556842)(1878,0.0574012)(1879,0.0600742)(1880,0.0641645)(1881,0.0500009)(1882,0.0558053)(1883,0.0628136)(1884,0.0703069)(1885,0.0502534)(1886,0.0535101)(1887,0.0546257)(1888,0.053471)(1889,0.0533329)(1890,0.0495064)(1891,0.0458977)(1892,0.043255)(1893,0.0398549)(1894,0.0400934)(1895,0.0416723)(1896,0.044305)(1897,0.0389696)(1898,0.0416245)(1899,0.0436098)(1900,0.0445339)(1901,0.0713074)(1902,0.0700245)(1903,0.0682784)(1904,0.0668281)(1905,0.0662032)(1906,0.0666142)(1907,0.0679748)(1908,0.0699883)(1909,0.0485717)(1910,0.0513852)(1911,0.0539929)(1912,0.0562706)(1913,0.100413)(1914,0.102708)(1915,0.105402)(1916,0.108719)(1917,0.112604)(1918,0.116622)(1919,0.119993)(1920,0.121807)(1921,0.142314)(1922,0.139718)(1923,0.11192)(1924,0.107469)(1925,0.103286)(1926,0.100404)(1927,0.0996032)(1928,0.101366)(1929,0.0817758)(1930,0.0881332)(1931,0.0963495)(1932,0.105921)(1933,0.157378)(1934,0.170747)(1935,0.182077)(1936,0.190149)(1937,0.194148)(1938,0.193797)(1939,0.18937)(1940,0.181604)(1941,0.171573)(1942,0.160539)(1943,0.14979)(1944,0.140475)(1945,0.133489)(1946,0.17179)(1947,0.175107)(1948,0.182512)(1949,0.193565)(1950,0.207254)(1951,0.221933)(1952,0.235457)(1953,0.167304)(1954,0.173432)(1955,0.175153)(1956,0.173226)(1957,0.169295)(1958,0.165228)(1959,0.16247)(1960,0.161696)(1961,0.162787)(1962,0.157766)(1963,0.160937)(1964,0.15012)(1965,0.150888)(1966,0.148421)(1967,0.14283)(1968,0.134935)(1969,0.126011)(1970,0.17144)(1971,0.163707)(1972,0.158461)(1973,0.156176)(1974,0.156963)(1975,0.166858)(1976,0.172849)(1977,0.17979)(1978,0.188976)(1979,0.194295)(1980,0.196605)(1981,0.19536)(1982,0.170576)(1983,0.165115)(1984,0.158316)(1985,0.151353)(1986,0.145215)(1987,0.140524)(1988,0.137457)(1989,0.164758)(1990,0.164139)(1991,0.163485)(1992,0.161836)(1993,0.185126)(1994,0.179331)(1995,0.171782)(1996,0.163539)(1997,0.155948)(1998,0.150254)(1999,0.147294)(2000,0.147382)(2001,0.157154)(2002,0.162754)(2003,0.169658)(2004,0.176537)(2005,0.181964)(2006,0.184768)(2007,0.184482)(2008,0.181613)(2009,0.215381)(2010,0.21131)(2011,0.209319) (2012,0)
};

\addplot [
color=tcolor7,
solid,
fill=tcolor7
]
coordinates{
(1790,0) (1790,0.176381)(1791,0.160443)(1792,0.146694)(1793,0.136273)(1794,0.129843)(1795,0.127701)(1796,0.12988)(1797,0.105991)(1798,0.114316)(1799,0.124753)(1800,0.136042)(1801,0.112905)(1802,0.121012)(1803,0.127322)(1804,0.132079)(1805,0.13603)(1806,0.14004)(1807,0.144725)(1808,0.15021)(1809,0.146515)(1810,0.152501)(1811,0.156953)(1812,0.159038)(1813,0.158438)(1814,0.155457)(1815,0.150916)(1816,0.145903)(1817,0.119115)(1818,0.116937)(1819,0.116335)(1820,0.117166)(1821,0.118913)(1822,0.120829)(1823,0.122164)(1824,0.122396)(1825,0.143642)(1826,0.141634)(1827,0.138938)(1828,0.136035)(1829,0.141135)(1830,0.139033)(1831,0.137313)(1832,0.135829)(1833,0.134396)(1834,0.132874)(1835,0.131214)(1836,0.129474)(1837,0.136872)(1838,0.135593)(1839,0.13471)(1840,0.134383)(1841,0.129356)(1842,0.130652)(1843,0.132888)(1844,0.136028)(1845,0.12335)(1846,0.127866)(1847,0.132529)(1848,0.136951)(1849,0.120257)(1850,0.118954)(1851,0.122065)(1852,0.125183)(1853,0.133904)(1854,0.138806)(1855,0.144146)(1856,0.149499)(1857,0.140112)(1858,0.143084)(1859,0.143709)(1860,0.141521)(1861,0.141236)(1862,0.134709)(1863,0.127293)(1864,0.12)(1865,0.151876)(1866,0.147043)(1867,0.143764)(1868,0.141823)(1869,0.113933)(1870,0.114549)(1871,0.115149)(1872,0.115335)(1873,0.114842)(1874,0.113622)(1875,0.111821)(1876,0.109677)(1877,0.113905)(1878,0.111828)(1879,0.109875)(1880,0.1081)(1881,0.101159)(1882,0.100362)(1883,0.0998834)(1884,0.0997239)(1885,0.105029)(1886,0.105757)(1887,0.106242)(1888,0.106158)(1889,0.0990777)(1890,0.0973477)(1891,0.0949299)(1892,0.092238)(1893,0.0954113)(1894,0.0935495)(1895,0.0926517)(1896,0.0926941)(1897,0.1154)(1898,0.116865)(1899,0.117864)(1900,0.117943)(1901,0.113531)(1902,0.112632)(1903,0.112009)(1904,0.112412)(1905,0.114378)(1906,0.118011)(1907,0.12287)(1908,0.127992)(1909,0.0962346)(1910,0.100205)(1911,0.101826)(1912,0.100814)(1913,0.107807)(1914,0.101995)(1915,0.0963871)(1916,0.0918482)(1917,0.0888198)(1918,0.0873063)(1919,0.0869379)(1920,0.0870745)(1921,0.077413)(1922,0.0768041)(1923,0.113231)(1924,0.108592)(1925,0.102124)(1926,0.0944932)(1927,0.086477)(1928,0.0787458)(1929,0.0960485)(1930,0.088793)(1931,0.0827968)(1932,0.0781216)(1933,0.0803906)(1934,0.0781823)(1935,0.0771903)(1936,0.0773388)(1937,0.0785325)(1938,0.0806528)(1939,0.0835391)(1940,0.0869606)(1941,0.0905978)(1942,0.094053)(1943,0.0969024)(1944,0.0987854)(1945,0.0995109)(1946,0.0712493)(1947,0.0723074)(1948,0.0731102)(1949,0.074157)(1950,0.0760843)(1951,0.0795812)(1952,0.085286)(1953,0.055628)(1954,0.0631143)(1955,0.0716383)(1956,0.0800445)(1957,0.0865614)(1958,0.0892916)(1959,0.0870419)(1960,0.0800216)(1961,0.0698589)(1962,0.0901736)(1963,0.0754902)(1964,0.118372)(1965,0.104982)(1966,0.0979636)(1967,0.096911)(1968,0.101335)(1969,0.110674)(1970,0.0601508)(1971,0.068922)(1972,0.0776793)(1973,0.0846867)(1974,0.0883942)(1975,0.085103)(1976,0.0811279)(1977,0.0745586)(1978,0.0872284)(1979,0.077746)(1980,0.0696488)(1981,0.0635078)(1982,0.0662688)(1983,0.0642285)(1984,0.0639285)(1985,0.0649349)(1986,0.0667303)(1987,0.0687646)(1988,0.0705504)(1989,0.0733529)(1990,0.0742304)(1991,0.0747356)(1992,0.0752623)(1993,0.075631)(1994,0.0773514)(1995,0.0797496)(1996,0.0825969)(1997,0.0854594)(1998,0.087769)(1999,0.0889308)(2000,0.0884704)(2001,0.0934655)(2002,0.0893762)(2003,0.0841186)(2004,0.0784259)(2005,0.0729695)(2006,0.0682513)(2007,0.0646231)(2008,0.0623595)(2009,0.0478118)(2010,0.0488717)(2011,0.0515526) (2012,0)
};

\addplot [
color=tcolor8,
solid,
fill=tcolor8
]
coordinates{
(1790,0) (1790,0.161605)(1791,0.16219)(1792,0.1617)(1793,0.160758)(1794,0.159969)(1795,0.159787)(1796,0.160388)(1797,0.123066)(1798,0.124866)(1799,0.126288)(1800,0.126657)(1801,0.141884)(1802,0.140378)(1803,0.137481)(1804,0.134146)(1805,0.131428)(1806,0.130203)(1807,0.130978)(1808,0.133818)(1809,0.130463)(1810,0.136296)(1811,0.141627)(1812,0.145096)(1813,0.145553)(1814,0.142442)(1815,0.136053)(1816,0.127479)(1817,0.151104)(1818,0.140775)(1819,0.132858)(1820,0.127917)(1821,0.125908)(1822,0.12633)(1823,0.128382)(1824,0.131102)(1825,0.114748)(1826,0.116329)(1827,0.116778)(1828,0.116079)(1829,0.140515)(1830,0.138366)(1831,0.136003)(1832,0.133641)(1833,0.131311)(1834,0.128927)(1835,0.126404)(1836,0.123759)(1837,0.147658)(1838,0.145168)(1839,0.143396)(1840,0.142614)(1841,0.128446)(1842,0.129638)(1843,0.131289)(1844,0.132769)(1845,0.156835)(1846,0.156814)(1847,0.155003)(1848,0.151638)(1849,0.156824)(1850,0.140106)(1851,0.13778)(1852,0.137097)(1853,0.129509)(1854,0.133611)(1855,0.139582)(1856,0.14701)(1857,0.148557)(1858,0.156729)(1859,0.163922)(1860,0.169182)(1861,0.143135)(1862,0.143381)(1863,0.141412)(1864,0.1377)(1865,0.147012)(1866,0.143587)(1867,0.140614)(1868,0.138794)(1869,0.112197)(1870,0.114593)(1871,0.118375)(1872,0.123011)(1873,0.127682)(1874,0.131501)(1875,0.133775)(1876,0.134221)(1877,0.130101)(1878,0.12819)(1879,0.125897)(1880,0.123863)(1881,0.114283)(1882,0.114314)(1883,0.115093)(1884,0.116323)(1885,0.105816)(1886,0.107103)(1887,0.107554)(1888,0.10688)(1889,0.106241)(1890,0.103102)(1891,0.0990377)(1892,0.094471)(1893,0.0888281)(1894,0.0846841)(1895,0.0811944)(1896,0.0785015)(1897,0.100178)(1898,0.0989541)(1899,0.0983375)(1900,0.097988)(1901,0.0846532)(1902,0.0842909)(1903,0.0834688)(1904,0.082098)(1905,0.0801818)(1906,0.0777818)(1907,0.0750094)(1908,0.0720351)(1909,0.0849839)(1910,0.0838279)(1911,0.0828902)(1912,0.0821501)(1913,0.0959987)(1914,0.0943204)(1915,0.092658)(1916,0.0908193)(1917,0.0886941)(1918,0.0862626)(1919,0.0835969)(1920,0.0808548)(1921,0.0780315)(1922,0.0761756)(1923,0.0770736)(1924,0.0766599)(1925,0.0769349)(1926,0.0778101)(1927,0.0791006)(1928,0.0805291)(1929,0.064075)(1930,0.0651513)(1931,0.0654701)(1932,0.064859)(1933,0.0901997)(1934,0.0866993)(1935,0.082099)(1936,0.0767976)(1937,0.0712456)(1938,0.0658811)(1939,0.061069)(1940,0.0570595)(1941,0.0539786)(1942,0.051842)(1943,0.050576)(1944,0.0500348)(1945,0.0500137)(1946,0.0391442)(1947,0.0403828)(1948,0.0413873)(1949,0.0419667)(1950,0.042045)(1951,0.0416726)(1952,0.0409856)(1953,0.0669715)(1954,0.0664857)(1955,0.0656898)(1956,0.0646466)(1957,0.0633819)(1958,0.0618973)(1959,0.0601695)(1960,0.0581666)(1961,0.0559004)(1962,0.0596894)(1963,0.0574247)(1964,0.046468)(1965,0.0455634)(1966,0.0450133)(1967,0.0448288)(1968,0.0449221)(1969,0.0451413)(1970,0.0465166)(1971,0.0472976)(1972,0.0476647)(1973,0.0475085)(1974,0.0468286)(1975,0.0491936)(1976,0.0476323)(1977,0.0458073)(1978,0.037644)(1979,0.0361595)(1980,0.0348674)(1981,0.0339043)(1982,0.0368558)(1983,0.0370502)(1984,0.0378307)(1985,0.0391861)(1986,0.0410313)(1987,0.0431861)(1988,0.0453665)(1989,0.0384919)(1990,0.0395473)(1991,0.0398332)(1992,0.0393051)(1993,0.0423778)(1994,0.0406257)(1995,0.0387284)(1996,0.0369935)(1997,0.0356485)(1998,0.0348016)(1999,0.0344159)(2000,0.0342992)(2001,0.0406201)(2002,0.0399594)(2003,0.038431)(2004,0.0359145)(2005,0.0326008)(2006,0.0289345)(2007,0.0254581)(2008,0.0226492)(2009,0.0127879)(2010,0.01246)(2011,0.0130058) (2012,0)
};

\addplot [
color=tcolor9,
solid,
fill=tcolor9
]
coordinates{
(1790,0) (1790,0.108185)(1791,0.111892)(1792,0.115111)(1793,0.117037)(1794,0.117008)(1795,0.114786)(1796,0.11069)(1797,0.118267)(1798,0.112998)(1799,0.108531)(1800,0.105432)(1801,0.106753)(1802,0.107832)(1803,0.109903)(1804,0.112356)(1805,0.114458)(1806,0.115542)(1807,0.115232)(1808,0.113589)(1809,0.104353)(1810,0.102605)(1811,0.101429)(1812,0.101416)(1813,0.102877)(1814,0.105776)(1815,0.109758)(1816,0.114278)(1817,0.117948)(1818,0.122466)(1819,0.126704)(1820,0.130768)(1821,0.134852)(1822,0.139125)(1823,0.143646)(1824,0.148322)(1825,0.113376)(1826,0.116763)(1827,0.119554)(1828,0.121472)(1829,0.130275)(1830,0.130737)(1831,0.130551)(1832,0.130039)(1833,0.129496)(1834,0.129116)(1835,0.128957)(1836,0.128967)(1837,0.133669)(1838,0.133949)(1839,0.134085)(1840,0.134117)(1841,0.129395)(1842,0.129847)(1843,0.130782)(1844,0.132215)(1845,0.124091)(1846,0.126266)(1847,0.127781)(1848,0.127914)(1849,0.141192)(1850,0.129875)(1851,0.123987)(1852,0.116425)(1853,0.14667)(1854,0.13612)(1855,0.126409)(1856,0.118193)(1857,0.118662)(1858,0.114053)(1859,0.111128)(1860,0.109477)(1861,0.114259)(1862,0.114059)(1863,0.113742)(1864,0.112896)(1865,0.111735)(1866,0.11072)(1867,0.108711)(1868,0.105863)(1869,0.136985)(1870,0.133397)(1871,0.129903)(1872,0.12681)(1873,0.124395)(1874,0.122953)(1875,0.122795)(1876,0.124168)(1877,0.107964)(1878,0.11212)(1879,0.117323)(1880,0.123091)(1881,0.108209)(1882,0.112967)(1883,0.116945)(1884,0.120075)(1885,0.119)(1886,0.121713)(1887,0.124307)(1888,0.126775)(1889,0.127026)(1890,0.127981)(1891,0.127326)(1892,0.124815)(1893,0.122473)(1894,0.117616)(1895,0.113102)(1896,0.110033)(1897,0.11908)(1898,0.121783)(1899,0.12771)(1900,0.136442)(1901,0.100199)(1902,0.107675)(1903,0.113662)(1904,0.116736)(1905,0.116101)(1906,0.111861)(1907,0.104934)(1908,0.0966558)(1909,0.166391)(1910,0.156469)(1911,0.148326)(1912,0.14219)(1913,0.0735065)(1914,0.0714739)(1915,0.0704176)(1916,0.0702487)(1917,0.0709569)(1918,0.0725587)(1919,0.0750327)(1920,0.0782523)(1921,0.0795118)(1922,0.0834363)(1923,0.0927138)(1924,0.0949712)(1925,0.0950224)(1926,0.0925757)(1927,0.0878855)(1928,0.0816607)(1929,0.109383)(1930,0.100519)(1931,0.0925367)(1932,0.085911)(1933,0.0586199)(1934,0.0558594)(1935,0.053792)(1936,0.0522)(1937,0.0509045)(1938,0.0498056)(1939,0.0488816)(1940,0.0481508)(1941,0.0476175)(1942,0.0472232)(1943,0.0468203)(1944,0.0461793)(1945,0.0450397)(1946,0.0427994)(1947,0.0412783)(1948,0.0389202)(1949,0.0359822)(1950,0.0328621)(1951,0.0299613)(1952,0.027574)(1953,0.0488347)(1954,0.0475437)(1955,0.0471085)(1956,0.0472945)(1957,0.047772)(1958,0.0481749)(1959,0.0481714)(1960,0.0475536)(1961,0.0463272)(1962,0.0456361)(1963,0.0443782)(1964,0.0542876)(1965,0.0549088)(1966,0.0568479)(1967,0.0602659)(1968,0.065075)(1969,0.0709005)(1970,0.046421)(1971,0.0506839)(1972,0.0540678)(1973,0.0560354)(1974,0.0563552)(1975,0.0487751)(1976,0.046616)(1977,0.0437093)(1978,0.0460784)(1979,0.0424425)(1980,0.0390356)(1981,0.0360766)(1982,0.0541546)(1983,0.051771)(1984,0.0505581)(1985,0.0505932)(1986,0.051904)(1987,0.0544374)(1988,0.0580089)(1989,0.0473597)(1990,0.0508264)(1991,0.0538924)(1992,0.0560982)(1993,0.0582886)(1994,0.0583554)(1995,0.0575789)(1996,0.0564187)(1997,0.0553601)(1998,0.054766)(1999,0.0547695)(2000,0.0552203)(2001,0.0624256)(2002,0.0624184)(2003,0.0611587)(2004,0.0582925)(2005,0.0539265)(2006,0.0486261)(2007,0.0432212)(2008,0.0385319)(2009,0.0339267)(2010,0.032421)(2011,0.0328099) (2012,0)
};

\addplot [
color=tcolor10,
solid,
fill=tcolor10
]
coordinates{
(1790,0) (1790,0.131636)(1791,0.135141)(1792,0.138003)(1793,0.140012)(1794,0.141077)(1795,0.14129)(1796,0.140923)(1797,0.14483)(1798,0.145303)(1799,0.146242)(1800,0.14751)(1801,0.133583)(1802,0.135434)(1803,0.135806)(1804,0.134064)(1805,0.129837)(1806,0.1232)(1807,0.11473)(1808,0.105403)(1809,0.12765)(1810,0.118109)(1811,0.111067)(1812,0.107128)(1813,0.106498)(1814,0.109005)(1815,0.114101)(1816,0.12086)(1817,0.15667)(1818,0.164943)(1819,0.17104)(1820,0.174192)(1821,0.174401)(1822,0.172397)(1823,0.169368)(1824,0.166597)(1825,0.135094)(1826,0.136023)(1827,0.138963)(1828,0.143823)(1829,0.122865)(1830,0.129241)(1831,0.135717)(1832,0.141458)(1833,0.145641)(1834,0.147623)(1835,0.147107)(1836,0.144235)(1837,0.149633)(1838,0.143829)(1839,0.137771)(1840,0.132179)(1841,0.156895)(1842,0.152544)(1843,0.149287)(1844,0.146576)(1845,0.150907)(1846,0.14801)(1847,0.144126)(1848,0.139378)(1849,0.142889)(1850,0.154542)(1851,0.151165)(1852,0.149574)(1853,0.117959)(1854,0.119865)(1855,0.122042)(1856,0.123598)(1857,0.147541)(1858,0.146159)(1859,0.143261)(1860,0.139923)(1861,0.13593)(1862,0.136483)(1863,0.140649)(1864,0.149156)(1865,0.118649)(1866,0.13291)(1867,0.149206)(1868,0.165606)(1869,0.144096)(1870,0.153513)(1871,0.158922)(1872,0.160421)(1873,0.158954)(1874,0.155881)(1875,0.152469)(1876,0.14954)(1877,0.161769)(1878,0.160249)(1879,0.158543)(1880,0.155878)(1881,0.181258)(1882,0.174937)(1883,0.166997)(1884,0.158322)(1885,0.176564)(1886,0.169364)(1887,0.164624)(1888,0.162903)(1889,0.16437)(1890,0.168665)(1891,0.175157)(1892,0.182854)(1893,0.190573)(1894,0.197015)(1895,0.201122)(1896,0.202391)(1897,0.163167)(1898,0.161026)(1899,0.158094)(1900,0.155168)(1901,0.10287)(1902,0.102212)(1903,0.101972)(1904,0.101723)(1905,0.100848)(1906,0.0987038)(1907,0.0948592)(1908,0.0893367)(1909,0.163323)(1910,0.15352)(1911,0.143886)(1912,0.135681)(1913,0.0770548)(1914,0.0746308)(1915,0.0743369)(1916,0.0762257)(1917,0.080349)(1918,0.0867598)(1919,0.0954956)(1920,0.106547)(1921,0.0746969)(1922,0.0845527)(1923,0.0792029)(1924,0.0885242)(1925,0.0970628)(1926,0.103606)(1927,0.106936)(1928,0.106199)(1929,0.0920454)(1930,0.0851206)(1931,0.0760202)(1932,0.0661803)(1933,0.0663875)(1934,0.0570478)(1935,0.0495784)(1936,0.0439964)(1937,0.0400646)(1938,0.0374084)(1939,0.035565)(1940,0.0340038)(1941,0.0321766)(1942,0.0296386)(1943,0.0262162)(1944,0.0221057)(1945,0.0177916)(1946,0.0549113)(1947,0.0429843)(1948,0.0337616)(1949,0.0272179)(1950,0.0229602)(1951,0.0205091)(1952,0.0194335)(1953,0.0359316)(1954,0.0375439)(1955,0.0394456)(1956,0.0408051)(1957,0.0409354)(1958,0.039578)(1959,0.0370022)(1960,0.033836)(1961,0.0307772)(1962,0.0280722)(1963,0.0268681)(1964,0.0248108)(1965,0.0260354)(1966,0.0281953)(1967,0.0309961)(1968,0.033871)(1969,0.0360485)(1970,0.0251821)(1971,0.0249653)(1972,0.0236694)(1973,0.0217358)(1974,0.019719)(1975,0.024356)(1976,0.0229938)(1977,0.0226428)(1978,0.0161688)(1979,0.0175231)(1980,0.0196791)(1981,0.0225915)(1982,0.0138127)(1983,0.0159389)(1984,0.0180447)(1985,0.0198956)(1986,0.0212835)(1987,0.022066)(1988,0.0221783)(1989,0.0265649)(1990,0.0252544)(1991,0.0233902)(1992,0.0211972)(1993,0.0160145)(1994,0.0142209)(1995,0.0126721)(1996,0.0114545)(1997,0.0106073)(1998,0.0101308)(1999,0.00999186)(2000,0.0101226)(2001,0.0141038)(2002,0.0145499)(2003,0.0148184)(2004,0.0147431)(2005,0.0142675)(2006,0.0134778)(2007,0.0125767)(2008,0.0118149)(2009,0.00319153)(2010,0.00325606)(2011,0.00354614) (2012,0)
};

\end{axis}
\end{tikzpicture}

%% file: perplexity.tikz
%
%
\begin{tikzpicture}

\begin{axis}[%
scale only axis,
width=0.8\columnwidth,
height=0.4\columnwidth,
xmin=0, xmax=100,
ymin=1900, ymax=2400,
xlabel={$$\# iterations$$},
ylabel={$$perplexity$$},
axis on top,
legend entries={$$Linear Model$$,$$Kernel Model$$},
legend style={nodes=right}]
\addplot [
color=red,
solid
]
coordinates{
 (1.88594,2347.23)(2,2339.35)(3,2291.72)(4,2264.23)(5,2248.79)(6,2239.33)(7,2232.94)(8,2228.43)(9,2225.21)(10,2222.73)(11,2220.73)(12,2219.11)(13,2217.73)(14,2216.53)(15,2215.47)(16,2214.54)(17,2213.72)(18,2212.97)(19,2212.26)(20,2211.6)(21,2210.99)(22,2210.43)(23,2209.92)(24,2209.46)(25,2209.01)(26,2208.61)(27,2208.22)(28,2207.86)(29,2207.51)(30,2207.18)(31,2206.86)(32,2206.56)(33,2206.27)(34,2205.98)(35,2205.7)(36,2205.43)(37,2205.19)(38,2204.96)(39,2204.74)(40,2204.53)(41,2204.33)(42,2204.14)(43,2203.97)(44,2203.8)(45,2203.64)(46,2203.49)(47,2203.34)(48,2203.19)(49,2203.05)(50,2202.92)(51,2202.79)(52,2202.66)(53,2202.51)(54,2202.38)(55,2202.25)(56,2202.13)(57,2202.01)(58,2201.89)(59,2201.78)(60,2201.69)(61,2201.6)(62,2201.52)(63,2201.43)(64,2201.35)(65,2201.27)(66,2201.19)(67,2201.11)(68,2201.03)(69,2200.96)(70,2200.87)(71,2200.76)(72,2200.63)(73,2200.54)(74,2200.47)(75,2200.41)(76,2200.35)(77,2200.3)(78,2200.25)(79,2200.2)(80,2200.15)(81,2200.1)(82,2200.06)(83,2200.01)(84,2199.97)(85,2199.93)(86,2199.88)(87,2199.84)(88,2199.8)(89,2199.76)(90,2199.72)(91,2199.69)(92,2199.65)(93,2199.61)(94,2199.57)(95,2199.54)(96,2199.5)(97,2199.47)(98,2199.44)(99,2199.4)(100,2199.37) 
};

\addplot [
color=blue,
solid
]
coordinates{
 (0.0655831,4817.55)(1,2218.05)(2,2106.3)(3,2041.35)(4,2015.41)(5,2002.73)(6,1994.12)(7,1987.06)(8,1983.48)(9,1981.31)(10,1979.89)(11,1977.9)(12,1975.77)(13,1973.81)(14,1972.2)(15,1972.27)(16,1970.99)(17,1969.89)(18,1969.36)(19,1968.3)(20,1968.05)(21,1967.18)(22,1966.68)(23,1966.03)(24,1965.59)(25,1965.15)(26,1965.03)(27,1965.33)(28,1964.98)(29,1964.58)(30,1964.37)(31,1964.33)(32,1964.22)(33,1964.02)(34,1963.9)(35,1963.75)(36,1963.76)(37,1963.48)(38,1963.55)(39,1963.88)(40,1963.85)(41,1963.68)(42,1963.65)(43,1963.85)(44,1963.9)(45,1963.76)(46,1963.92)(47,1964.38)(48,1964.18)(49,1964.15)(50,1963.82)(51,1963.67)(52,1963.77)(53,1963.69)(54,1963.55)(55,1963.69)(56,1963.32)(57,1963.31)(58,1963.57)(59,1963.95)(60,1963.63)(61,1963.49)(62,1963.27)(63,1963.34)(64,1963.51)(65,1963.13)(66,1962.88)(67,1963.3)(68,1963.25)(69,1963.23)(70,1963.06)(71,1963.05)(72,1963.02)(73,1962.97)(74,1962.96)(75,1962.91)(76,1962.89)(77,1962.9)(78,1962.92)(79,1962.91)(80,1962.93)(81,1962.91)(82,1962.92)(83,1962.91)(84,1962.93)(85,1962.93)(86,1962.96)(87,1963.02)(88,1962.99)(89,1963.02)(90,1963.06)(91,1963.06)(92,1963.48)(93,1963.09)(94,1963.21)(95,1963.1)(96,1963.08)(97,1963.08)(98,1963.1)(99,1963.11)(100,1963.08) 
};

\end{axis}
\end{tikzpicture}

%% file: wiki_perplexity.tikz
%
%
\begin{tikzpicture}

\begin{axis}[%
scale only axis,
width=0.8\columnwidth,
height=0.4\columnwidth,
xmin=0, xmax=100,
ymin=800, ymax=1150,
xlabel={$$\# iterations$$},
ylabel={$$perplexity$$},
axis on top,
legend entries={$$Constant Model$$,$$Kernel Model$$},
legend style={nodes=right}]
\addplot [
color=red,
solid
]
coordinates{
 (1,1097.17)(2,1022.35)(3,991.605)(4,971.343)(5,957.652)(6,947.647)(7,940.251)(8,934.599)(9,929.77)(10,925.499)(11,922.017)(12,918.86)(13,915.554)(14,912.251)(15,909.179)(16,906.432)(17,904.36)(18,902.364)(19,900.586)(20,899.046)(21,897.564)(22,895.896)(23,894.672)(24,893.667)(25,892.8)(26,891.932)(27,891.123)(28,890.39)(29,889.7)(30,889.123)(31,888.551)(32,887.984)(33,887.36)(34,886.758)(35,886.2)(36,885.628)(37,885.184)(38,884.71)(39,884.317)(40,883.941)(41,883.554)(42,883.113)(43,882.596)(44,882.064)(45,881.561)(46,881.157)(47,880.794)(48,880.468)(49,880.15)(50,879.873)(51,879.624)(52,879.4)(53,879.186)(54,878.907)(55,878.677)(56,878.483)(57,878.314)(58,878.142)(59,877.946)(60,877.714)(61,877.479)(62,877.26)(63,877.033)(64,876.795)(65,876.548)(66,876.267)(67,876.097)(68,875.95)(69,875.791)(70,875.646)(71,875.518)(72,875.4)(73,875.284)(74,875.17)(75,875.059)(76,874.932)(77,874.763)(78,874.53)(79,874.244)(80,873.977)(81,873.777)(82,873.628)(83,873.495)(84,873.362)(85,873.236)(86,873.126)(87,873.034)(88,872.943)(89,872.84)(90,872.721)(91,872.592)(92,872.474)(93,872.36)(94,872.231)(95,872.059)(96,871.884)(97,871.762)(98,871.659)(99,871.568)(100,871.488) 
};

\addplot [
color=blue,
solid
]
coordinates{
 (1,1116.5)(2,1042.31)(3,1006.62)(4,982.701)(5,966.786)(6,956.284)(7,947.779)(8,940.114)(9,933.549)(10,927.716)(11,922.599)(12,1006.21)(13,900.48)(14,898.504)(15,885.134)(16,882.783)(17,879.442)(18,877.886)(19,876.192)(20,874.29)(21,873.707)(22,866.267)(23,863.786)(24,861.879)(25,860.644)(26,859.53)(27,858.073)(28,856.652)(29,854.654)(30,853.687)(31,852.842)(32,851.268)(33,850.661)(34,850.201)(35,849.787)(36,849.395)(37,848.955)(38,848.059)(39,847.617)(40,847.217)(41,846.793)(42,846.103)(43,845.688)(44,845.495)(45,845.223)(46,844.818)(47,844.373)(48,843.938)(49,843.543)(50,843.178)(51,842.926)(52,842.575)(53,842.306)(54,842.131)(55,841.989)(56,841.88)(57,841.76)(58,841.67)(59,841.554)(60,841.39)(61,841.103)(62,840.933)(63,840.797)(64,840.684)(65,840.583)(66,840.456)(67,840.263)(68,840.125)(69,840.034)(70,839.977)(71,839.889)(72,839.823)(73,839.749)(74,839.693)(75,839.636)(76,839.54)(77,839.454)(78,839.209)(79,839.111)(80,839.065)(81,838.338)(82,837.856)(83,837.702)(84,837.614)(85,837.539)(86,837.49)(87,837.434)(88,837.393)(89,837.336)(90,837.293)(91,837.234)(92,837.229)(93,837.167)(94,837.122)(95,837.045)(96,836.974)(97,836.883)(98,836.844)(99,836.803)(100,836.777) 
};

\end{axis}
\end{tikzpicture}

%% file: NIPS_perplexity.tikz
%
%
\begin{tikzpicture}

\begin{axis}[%
scale only axis,
width=0.8\columnwidth,
height=0.4\columnwidth,
xmin=0, xmax=100,
ymin=2500, ymax=3200,
xlabel={$$\# iterations$$},
ylabel={$$perplexity$$},
axis on top,
legend entries={$$Linear Model$$,$$Kernel Model$$},
legend style={nodes=right}]
\addplot [
color=red,
solid
]
coordinates{
 (1,3141.69)(2,3109.59)(3,3070.65)(4,3036.11)(5,3006.28)(6,2982.3)(7,2962.7)(8,2946.53)(9,2933.13)(10,2922.16)(11,2913.03)(12,2905.46)(13,2899.03)(14,2893.5)(15,2888.47)(16,2884.33)(17,2880.81)(18,2877.85)(19,2875.34)(20,2872.97)(21,2870.57)(22,2868.37)(23,2866.37)(24,2864.61)(25,2862.91)(26,2861.06)(27,2859.2)(28,2857.42)(29,2855.95)(30,2854.7)(31,2853.62)(32,2852.65)(33,2851.79)(34,2851.01)(35,2850.19)(36,2849.23)(37,2848.32)(38,2847.58)(39,2846.95)(40,2846.36)(41,2845.79)(42,2845.3)(43,2844.94)(44,2844.61)(45,2844.32)(46,2844.03)(47,2843.7)(48,2843.33)(49,2842.95)(50,2842.62)(51,2842.31)(52,2841.97)(53,2841.52)(54,2841.1)(55,2840.75)(56,2840.39)(57,2840.02)(58,2839.66)(59,2839.3)(60,2839.01)(61,2838.78)(62,2838.58)(63,2838.39)(64,2838.16)(65,2837.89)(66,2837.63)(67,2837.42)(68,2837.25)(69,2837.06)(70,2836.86)(71,2836.66)(72,2836.51)(73,2836.4)(74,2836.29)(75,2836.2)(76,2836.13)(77,2836.06)(78,2835.99)(79,2835.9)(80,2835.83)(81,2835.76)(82,2835.7)(83,2835.65)(84,2835.57)(85,2835.48)(86,2835.38)(87,2835.28)(88,2835.18)(89,2835.1)(90,2835.03)(91,2834.98)(92,2834.95)(93,2834.91)(94,2834.87)(95,2834.83)(96,2834.79)(97,2834.74)(98,2834.69)(99,2834.63)(100,2834.57) 
};

\addplot [
color=blue,
solid
]
coordinates{
 (1,2942.9)(2,2847.1)(3,2822.6)(4,2813.06)(5,2807.85)(6,2804.36)(7,2802)(8,2800.23)(9,2798.7)(10,2797.47)(11,2796.32)(12,2697.77)(13,2607.99)(14,2566.57)(15,2544.77)(16,2532.43)(17,2524.8)(18,2519.78)(19,2516.49)(20,2514.21)(21,2512.57)(22,2510.78)(23,2509.75)(24,2509.07)(25,2508.63)(26,2508.3)(27,2508.09)(28,2507.98)(29,2507.98)(30,2508.04)(31,2508.11)(32,2506.49)(33,2506.05)(34,2505.89)(35,2505.85)(36,2505.88)(37,2505.96)(38,2506.05)(39,2506.16)(40,2506.29)(41,2506.44)(42,2506.02)(43,2505.98)(44,2506.04)(45,2506.12)(46,2506.22)(47,2506.31)(48,2506.4)(49,2506.5)(50,2506.62)(51,2506.74)(52,2506.47)(53,2506.45)(54,2506.49)(55,2506.54)(56,2506.6)(57,2506.7)(58,2506.81)(59,2506.92)(60,2507.02)(61,2507.1)(62,2506.86)(63,2506.85)(64,2506.91)(65,2506.99)(66,2507.07)(67,2507.15)(68,2507.22)(69,2507.29)(70,2507.36)(71,2507.44)(72,2507.27)(73,2507.27)(74,2507.3)(75,2507.34)(76,2507.39)(77,2507.43)(78,2507.48)(79,2507.53)(80,2507.58)(81,2507.64)(82,2507.44)(83,2507.42)(84,2507.44)(85,2507.49)(86,2507.54)(87,2507.59)(88,2507.64)(89,2507.69)(90,2507.75)(91,2507.82)(92,2507.64)(93,2507.62)(94,2507.65)(95,2507.69)(96,2507.75)(97,2507.8)(98,2507.86)(99,2507.92)(100,2507.98) 
};

\end{axis}
\end{tikzpicture}